\documentclass[sigconf, nonacm]{acmart}

\usepackage{placeins}
\usepackage{amsmath}
\usepackage{xspace}
\usepackage{enumitem}
\usepackage{booktabs}
\usepackage{multirow}
\usepackage{caption}
\usepackage{subcaption}
\usepackage{algorithm}
\usepackage{algorithmic}
\usepackage[linguistics]{forest}
\usepackage{xcolor}

\newcommand{\sysname}{\texttt{SHiFT}\xspace}
\newcommand{\queryname}{\texttt{SHiFT-QL}\xspace}

\usepackage{enumitem}
\setitemize{noitemsep,topsep=0pt,parsep=0pt,partopsep=0pt}

\def \cS {\mathcal S}

\def \cD {\mathcal D}

\def \cM {\mathcal M}

\begin{document}
\title{\sysname: An Efficient, Flexible Search Engine for Transfer Learning}

\author{Cedric Renggli$^{\dagger}$,~~Xiaozhe Yao$^{\dagger}$,~~Luka Kolar$^{\dagger}$,~~Luka Rimanic$^{\dagger}$,~~Ana Klimovic$^\dagger$,~~Ce Zhang$^\dagger$}
\affiliation{%
    \institution{$^{\dagger}$ETH Zurich\\
        $^\dagger$\{cedric.renggli, xiaozhe.yao, luka.kolar, luka.rimanic, aklimovic, ce.zhang\}@inf.ethz.ch}
}

\begin{abstract}
Transfer learning can be seen as a data- and compute-efficient alternative to training models from scratch. The emergence of rich model repositories, such as TensorFlow Hub, enables practitioners and researchers to unleash the potential of these models across a wide range of downstream tasks. As these repositories keep growing exponentially, efficiently selecting a good model for the task at hand becomes paramount. However, a single generic search strategy (e.g., taking the model with the highest linear classifier accuracy) does not lead to optimal model selection for diverse downstream tasks. In fact, using hybrid or mixed strategies can often be beneficial. Therefore, we propose \sysname, the first downstream task-aware, flexible, and efficient model search engine for transfer learning.  Users interface with \sysname using the \queryname query language, which gives users the flexibility to customize their search criteria. We optimize \queryname queries using a cost-based decision maker and evaluate them on a wide rang of tasks. Motivated by the iterative nature of machine learning development, we further support efficient incremental executions of our queries, which requires a special implementation when jointly used with our optimizations.
\end{abstract}

\maketitle

\section{Introduction}

Transfer learning~\cite{weiss2016survey,mensink2021factors}
is an emerging paradigm of building machine learning (ML)
applications, which can have 
a profound impact on the architecture of 
today's machine learning
systems and platforms.
In a nutshell, transfer learning aims at training 
ML models with high quality without having to collect enormous datasets or spend a fortune on training from scratch. 
Instead, models are first \textit{pre-trained} on typically large and possibly private \textit{upstream} datasets, and are then 
made available via model repositories
such as TF-Hub, PyTorch Hub, and HuggingFace.
Given a new \textit{downstream} dataset, representative for the ML task, a user picks \textit{some}
of these pre-trained models
to \textit{fine-tune}, which typically requires adding randomly initialized layers to parts of the pre-trained deep network, and tuning all the parameters using the limited amount of downstream data. Transfer learning has been successfully applied to many domains and tasks~\citep{azizpour2015factors, long2015fully, ruder2019transfer, houlsby2019parameter}.

This process, despite being cheaper compared to training from scratch (i.e., with a fully randomly initialized network), still requires all parameters to be updated several times, which can be computationally demanding. With the growing number of pre-trained models available in online platforms like TF-Hub, PyTorch Hub, and HuggingFace, it is computationally infeasible to fine-tune \textit{all} models to find the one that performs best for a downstream task. As a result, a key defining component of a transfer learning application is a \textit{model search strategy}, which provides cheaper ways to identify promising models
to use. One challenge is that 
{\em different tasks might require very
different search strategies}~\cite{zhai2019visual, mensink2021factors}
and Table~\ref{tbl:queries} illustrates
several popular ones. 
Today, a user of these model repositories conducting transfer learning needs to
manage models and
implement search strategies 
all by themselves. Over the years,
in the context of building usable machine learning system (e.g., Ease.ML~\cite{karlavs2018ease, li2018ease, aguilar2021ease}),
we have observed several challenges
that our users face:

\begin{table}[t]
    \vspace{1em}
    \caption{Example search queries supported by \sysname, returning an ordering of registered models by their...}
    \vspace{-1.2em}
    \label{tbl:queries}
    \begin{tabular}{@{}l@{}}
        \toprule
        Q1: best reported upstream accuracy~\citep{kornblith2019better}                                         \\
        Q2: best nearest neighbor classifier accuracy~\citep{meiseles2020source, puigcerver2020experts}        \\
        Q3: best linear classifier accuracy~\citep{bao2019information, nguyen2020leep, tran2019transferability} \\
        Q4: Q1 and Q3 (on models excluding the result of Q1)~\citep{renggli2020model}                           \\
        Q5: highest (fine-tune) accuracy on the most similar task~\citep{achille2019task2vec}                       \\ \bottomrule
    \end{tabular}
    \vspace{-1em}
\end{table}

\begin{figure*}[ht!]
    \begin{center}
        \includegraphics[width=0.85\linewidth]{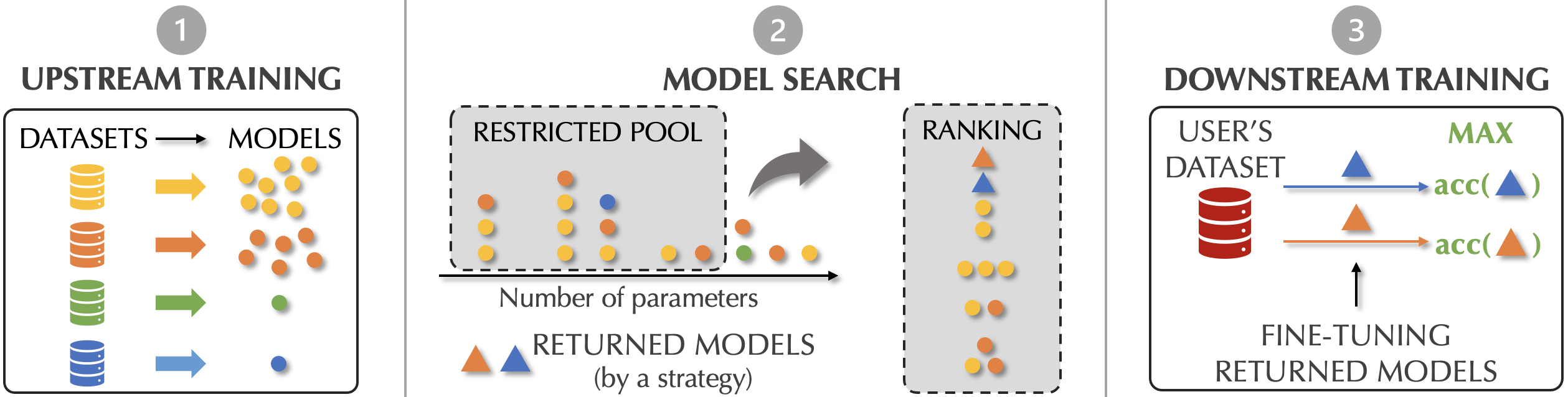}
    \end{center}
    \vspace{-1em}
    \caption{Three stages of a transfer learning setup: (1) Models are trained \textit{upstream} for various architecture and datasets, (2) the models are restricted into a \textit{pool}, and ranked by a search strategy, and (3) a subset of the models based on the ranking are fine-tuned on user's dataset in a downstream process. \sysname focuses on supporting the middle stage (2) efficiently and in a flexible, generic way.}
    \vspace{-1em}
    \label{fig:tranfer_learning}
\end{figure*}

{\bf Challenge 1. }
Our first observation is that the amount of
data and computation a user needs to perform quickly exceeds what most users, even 
competent software engineers, can handle. 
Some challenges are more on the engineering side ---
today's pre-trained models are scattered in different repositories, 
including HuggingFace, TF-Hub, and PyTorch Hub,
lacking a unified abstraction. 
However, many others are technical --- HuggingFace, TF-Hub, and PyTorch Hub
contain
34K, 1K, and 48 models, respectively, and sum up to more than
100GB in size.
Simply downloading all these models 
can take hours, not to mention running 
inference of all these models over 
a given dataset and implementing 
state-of-the-art search strategies. As a 
consequence, many users in our experience 
simply resort to only using the latest 
model --- \textit{this ignores the vast diversity 
of available models and, as we show in our previous 
systematic benchmark~\cite{renggli2020model},
can lead to a significant quality gap.}

{\bf Challenge 2. } 
The second challenge we observe is that
today's data-centric development pattern
of ML applications provides unique 
opportunities to speed up the 
model search process, which, if left to
the users, are quite hard to capture.
In many cases, users will conduct iterations
on the datasets---data cleaning, acquisition
of both labels and features---and execute similar  
model search queries for each of these datasets over time.
Since these datasets are similar to each other,
it is possible to save a significant amount of
computation if we carefully 
design incremental maintenance 
strategies for different search queries ---
\textit{all these opportunities to speed up 
model search are not captured by today's model repositories.}

{\bf Challenge 3. } 
The third challenge we observe is that 
model search is an active research area where 
new strategies are coming out quite frequently, yet 
easily implementing them in a system and comparing them 
to existing search strategies is a painful task.
Just in the last three years, researchers proposed 
at least eight new strategies~\citep{kornblith2019better, achille2019task2vec, bao2019information, nguyen2020leep, meiseles2020source, puigcerver2020experts, deshpande2021linearized, renggli2020model}.
Having users to catch up with these 
latest developments can be tedious and 
potentially a huge waste of 
resources. \textit{We are in a dire need 
for a unified framework that can be 
extended in a flexible way to 
support, benchmark, and automatically optimize 
newly proposed search strategies.}

\vspace{0.5em}
{\bf \sysname: Towards Data Management
for Transfer Learning.}
All three challenges, in our opinion,
lead to the same hypothesis:

\vspace{0.3em}
{\em 
We should shift the responsibility of
querying and conducting model search
over today's model repositories
from individual users to a
data management system, which 
can (1) provide a unified abstraction to 
connect all major model repositories,
(2) provide a flexible, extendable 
way of specifying search strategies, (3) automatically optimize its execution, and (4)
support efficient incremental execution. 
}
\vspace{0.3em}

In this paper, we present \sysname,
one of the first data management systems 
for transfer learning --- \sysname is definitely \textit{not} the first ``model management system'', an area that has attracted a lot of interest from 
the data management community~\cite{vartak2016modeldb,zaharia2018accelerating,orr2021managing, kumar2016model, kumar2017data, schelter2018challenges};
however, to our best knowledge,
it is the first one that 
focuses on enabling an efficient and flexible model
search functionality for transfer learning. This is achieved by supporting
diverse model search queries to find (near-)optimal pre-trained models for fine-tuning them on a downstream task.
Our technical contributions are as follows:

\vspace{0.1em}
\begin{itemize}[noitemsep,topsep=0pt,parsep=0pt,partopsep=0pt,topsep=0pt, leftmargin=*]
\item \textbf{C1 System Design.} We propose a unified abstraction in \sysname that can model all existing model search strategies that we are
aware of, and is extendable with respect to new models, model repositories, as well as potential new
search strategies.  We then present \queryname, a novel query language used to interract with \sysname. To compare different search strategies, we furthermore design an easy-to-use benchmark module.
\item \textbf{C2 System Optimization.} 
We carefully studied several system 
optimization opportunities, both for
a single query run and for incremental 
maintenance. \textbf{(C2.1)} For a single query run,
we propose several optimizations
to speed up the model search process, including
a novel strategy based on successive halving~\citep{jamieson2016non} to use resources more efficiently. 
\textbf{(C2.2)}
Furthermore, we develop a cost model to \textit{automatically} decide whether to apply different system optimizations, given a
search query, as well as systems and model properties.
\textbf{(C2.3)}
\sysname also efficiently supports incremental query execution, as required by the iterative nature of the model development process. 
This functionality is non-trivial, and we carefully design an incremental 
maintenance strategy for successive-halving.
\item \textbf{C3 Evaluation.} We evaluate \sysname over a diverse set of queries across three computer vision and two NLP datasets and more than 150 diverse models. We show that
\sysname outperforms
a baseline implementation
of the same search strategy by up to 1.57$\times$,
and in the incremental scenario up to 4.0$\times$ for 10\% feature changes for reasonably large test sets.
Compared with fine-tuning
all models without search strategy,
\sysname 
can be up to 22.6$\times$-45.0$\times$
faster on vision tasks. We furthermore provide an  extensive study using our benchmark module.
\item \textbf{C4 Open Source.} We make \sysname publicly available by open-sourcing it under \url{https://github.com/DS3Lab/shift}. We provide a clean documentation on how to implement new search strategies as well as guidelines and best-practices for choosing the right \queryname query under: \url{http://docs.shift.ml}
\end{itemize}

\section{Background and Motivation}
\label{sec:background}

\subsection{Transfer Learning}

\begin{figure}[t!]
    \begin{center}
        \includegraphics[width=\linewidth]{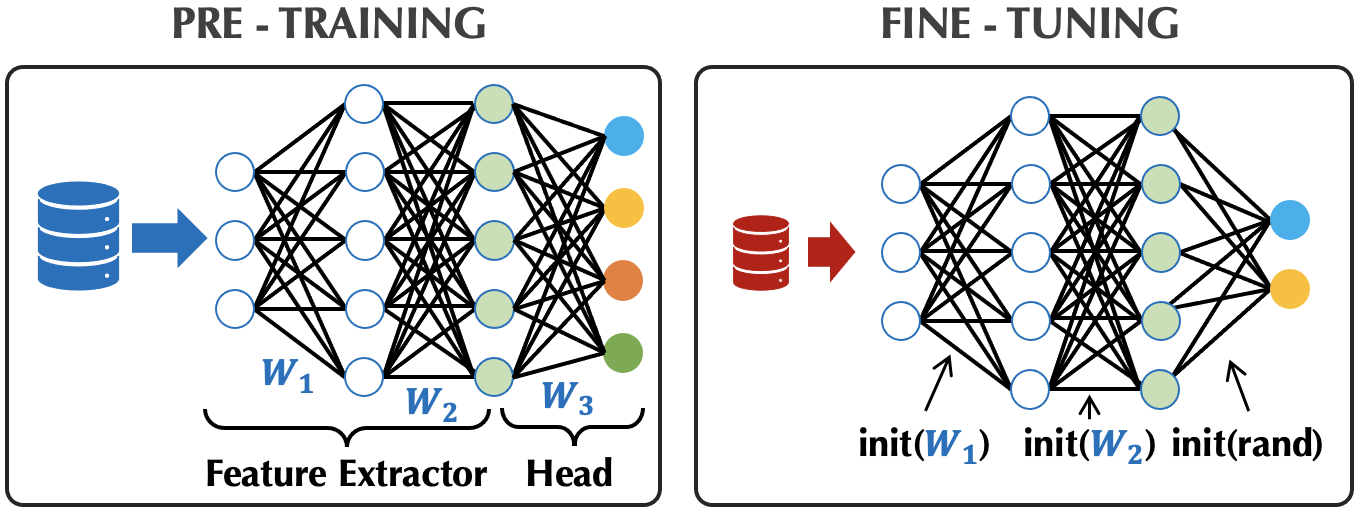}
    \end{center}
    \vspace{-1em}
    \caption{The difference between pre-training (left) and fine-tuning (right). The features stem from the last layer.}
    \label{fig:pre_training_vs_fine_tuning}
    \vspace{-2em}
\end{figure}

Transfer learning has been applied very successfully on popular deep learning modalities such as computer vision (CV) and natural language process (NLP) over the last few years~\cite{pan2009survey, tan2018survey, wang2018theoretical, weiss2016survey}. Transfer learning is typically divided into three steps as illustrated in Figure~\ref{fig:tranfer_learning}: (1) An upstream, or pre-training part, where a machine learning model is trained using a well-established approach (e.g., randomly initialized weights and using mini-batch stochastic gradient descent as an optimizer) and an upstream dataset. (2) Having access to multiple such models trained on various datasets or model architectures, users pick a single or a subset of the available models, through some model search strategy, for the subsequent part. Note that users typically do not have access to the upstream datasets for this model search part. (3) The chosen models are fine-tuned in a downstream part on users' datasets. There are multiple strategies on how to fine-tune a pre-trained model, the most prominent being illustrated in Figure~\ref{fig:pre_training_vs_fine_tuning}. Transfer learning and searching for models is mainly studied for classification tasks. We focus on this setting, hoping that our findings and proposed system will enable progress beyond classification. The pre-trained model is split into two parts: a \textit{feature extractor,} typically the entire network until the last layer, and the classification \textit{head} consisting of the final linear classification layer. The new model is then a copy of the feature extractor (i.e., weights and architecture), and a new randomly initialized head. Replacing the pre-trained linear head by a new one is usually unavoidable, as the number of classes often changes from the upstream task to the downstream one. All the weights of the network are then refined for multiple iterations using an iterative optimization algorithm like stochastic gradient descent (SGD).

The benefits of this three-step transfer learning process over training from scratch (i.e., training a randomly initialized network using the downstream dataset only) are twofold: (A) Using transfer learning, one can fine-tune a very large and complex network (i.e., many parameters forming a highly non-convex space to optimize in) even for limited amount of downstream data. (B) The fine-tuning process requires a much smaller number of steps in the iterative optimization process. Both (A) and (B) ensure a good initial condition to start the fine-tune optimization process. They are enabled by the fact that much larger, sometimes private upstream datasets are used to train for many iterations.
The choice of which pre-trained model to fine-tune has a high impact on the final accuracy one can expect (e.g., a sub-optimal pick in our experiments can lead to a 43\% downstream accuracy gap). If we are not concerned with optimization compute and power, we could fine-tune all models and pick the best one afterwards. This brute-force approach is usually infeasible in practice given the amount of currently available pre-trained models, along with the fact that this will not scale to more models in the future.
Therefore, we require a more efficient search strategy, which limits the number of models we need to fine-tune.

\subsection{Existing Pre-Trained Model Hubs}

There exist multiple prominent online pre-trained model repositories, most notably Tensorflow Hub, PyTorch Hub and HuggingFace.\footnote{\url{https://tfhub.dev}, \url{https://pytorch.org/hub}, and \url{https://huggingface.co/models}} Each repository has its own interface for accessing the fully trained model, using the corresponding deep learning framework, with optional direct access to the (last-layer) feature extractor in the case of TF-Hub. HuggingFace has an interface for both Tensorflow and PyTorch. These existing online repositories enable easy access to pre-trained models along with their weights and some additional meta-data such as the domain and tasks this model is designed for (e.g., vision and classification), number of parameters, name of the dataset used to pre-train, or performance on this upstream dataset.

\vspace{-0.5em}
\paragraph{Limitations}
Whilst accessing the models via existing online repositories is simple, they all share a common limitation when searching for the right model. That is, all platforms offer only plain search fields or filters\footnote{HuggingFace released an evaluation library (\url{https://huggingface.co/docs/evaluate}) after the submission of this paper. The functionality corresponds to simple task-aware search queries, yet not more complex hybrid or nested queries.}, which retrieve models by their name or some meta-data properties. As we will see in the next section, this allows to restrict the model pool and run (some) task-agnostic search queries. The implementation and support of any other search strategy is currently left to the user.

\subsection{Model Search Strategies}

A search query is a function $m(\cM, \cD, B)$ with a budget $B$, a set of models $\cM$, and a downstream task represented by a dataset $\cD$ as its input. The function outputs a set $\cS_m \subset \cM$ with $\vert \cS_m \vert \leq B$.
The input set $\cM$ either represents all models registered in \sysname, or a restricted subset, which we call a \textit{model pool}.
The quality of a query is not measured by the metric that the query itself uses (e.g., proxy accuracy or task similarity), but rather by the maximal accuracy attained when fine-tuning the resulting models, those in $\cS_m$. The difference between this accuracy and the maximal achievable accuracy if one would fine-tune all available models to the user is called \textit{regret}, formally:
\vspace{-0.5em}
\begin{equation}\label{eq:regret}
    \max_{m_i \in \cM} \mathbf{E}[t(m_i, \cD)] - \mathbf{E}\left[\max_{s_i \in \cS_m} t(s_i, \cD)\right],
\end{equation}
where $t(m, \cD)$ is the test accuracy achieved when fine-tuning model $m$ on dataset $\cD$.
Clearly, the budget $B$, which restricts the size of the subset $\cS_m$, has a direct impact on the regret. A large and diverse set $\cS_m$, with any diversity measure, is much more likely to result in small regret, while having a budget of $B=1$ makes the task of finding exactly the best model challenging.

\vspace{-0.5em}
\paragraph{Pool Restriction}
The downstream task itself defines the input modality (e.g., image or text) that needs to be supported by the models. Users then usually have use-case specific constraints, such as the framework (e.g., TensorFlow or Pytorch) one is restricted to, or the total number of parameters.

\begin{figure}[t!]
    \begin{center}
        \includegraphics[width=\linewidth]{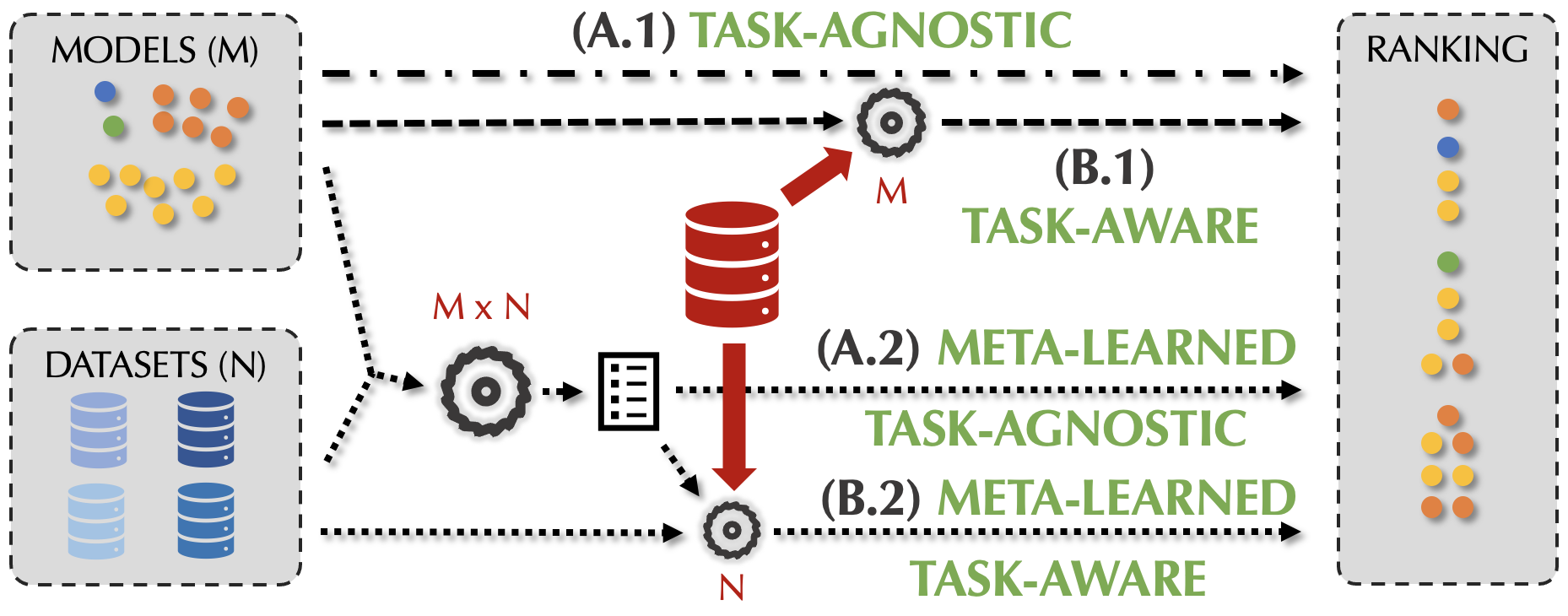}
    \end{center}
    \vspace{-1em}
    \caption{Categorization of model search strategies with their computational complexity. The $N$ benchmark datasets are different from the ones used to train upstream, and the computation of the cross-product fine-tune table ($M \times N$) is decoupled from the search complexity.}
    \vspace{-1em}
    \label{fig:search_strategies}
\end{figure}

We do not consider the pool restriction itself as a search strategy but rather as an integral part of any other search strategy that we outline next. Note that in the illustration in Figure~\ref{fig:search_strategies}, the search strategies do not return a subset of the model pool $\cM$, but rather rank models in $\cM$. We can map this ranking to our formal description of a search query $m(\cM, \cD, B)$, by selecting the \textit{top-B} models according to the ranking, randomly breaking ties.

\vspace{-0.5em}
\paragraph{Strategies}
Figure~\ref{fig:search_strategies} illustrates the different \textit{model search strategies} along with their computational requirements. We remark that model search strategies were extensively studied in our work \cite{renggli2020model}, whereas here we present an overview of these methods and facts that are important from the perspective of \sysname. As in~\cite{renggli2020model}, we divide model search strategies into two main categories: (A) task-agnostic strategies, which are those that ignore the downstream dataset, and (B) task-aware strategies, those that do take the downstream dataset into consideration.

\vspace{-0.5em}
\paragraph{(A.1) Task-Agnostic Search.} The first category ranks the models in the pool by completely ignoring the downstream dataset. This can range from naively ordering the models by their name, size, or date of appearance, to more prominent strategies suggested by related work: (a) ranking models trained on ImageNet based on the upstream accuracy~\citep{kornblith2019better} (for images), or the average GLUE~\citep{wang2018glue} performance (for text), and (b) favoring more robust models~\citep{deng2021adversarial}.

\vspace{-0.5em}
\paragraph{(A.2) Meta-Learned Task-Agnostic Search.} Instead of restricting to meta-data reported by the model publishers, one could fine-tune every model registered in the system on a fixed set of $N$ benchmark datasets (e.g., the 19 from VTAB~\citep{zhai2019visual}), or on a subset (e.g., only natural datasets in VTAB) and the aggregation metric used to rank the models (e.g., maximum over this subset) chosen by the user. Supporting such meta-learned search strategies requires \sysname to run some computation upon registering a new model by fine-tuning it using all suitable benchmark datasets, whereas the retrieval phase remains independent of the downstream task.

\vspace{-0.5em}
\paragraph{(B.1) Task-Aware Search.} 
Using linear classifier accuracies as a proxy to rank the models is often regarded as the standard when incorporating the downstream task into the search process~\citep{kornblith2019better, deshpande2021linearized}.
Instead of fine-tuning the weights of the pre-trained network as described previously, one \textit{freezes} them and only learns the weights of a newly initialized linear head. 
In a large empirical study we have shown that such a linear proxy can suffer from a relatively high regret when trusting this search strategy over exhaustively fine-tuning all models and then picking the best one~\citep{renggli2020model}. Nevertheless, this approach still represents one of the most powerful known search strategy. To improve computation time, 
researchers have proposed faster proxy methods by approximating the linear classifier accuracy~\citep{bao2019information,nguyen2020leep,tran2019transferability}, or relying on a cheaper classifier like the k-nearest neighbor~\citep{puigcerver2020experts,renggli2020model} and its approximations~\citep{meiseles2020source}.

\vspace{-0.5em}
\paragraph{(B.2) Meta-Learned Task-Aware Search.}
The goal is to favor models that perform well on benchmark datasets \textit{similar} to the downstream one. A prominent way of determining the similarity between datasets representing an ML task was introduced via learned task representations by~\citet{achille2019task2vec}. To return a ranking of the models, one has to compute a Task2Vec representation for the new dataset and then find the nearest task (i.e., via the distance metric introduced by \citet{achille2019task2vec}). Registering a new model can be computationally demanding as it requires fine-tune accuracies of this new model on all benchmark dataset, but is decoupled from the model search performed by the user.

\vspace{-0.5em}
\paragraph{Hybrid Search.} Empirically, with currently publicly available pre-trained models, for every single method there exists a case in which it suffers from a relatively high regret (i.e., returning a suboptimal model to fine-tune)~\citep{renggli2020model}. In our prior work, we proposed to extend the returned set to two or more models, where one can start mixing strategies (e.g., best task-agnostic and best task-aware model)~\cite{renggli2020model}. In particular, we showed that a simple hybrid search strategy that suggests fine-tuning the top-$1$ task-agnostic and top-($B-1$) task-aware model, often leads to superior results compared to fine-tuning the best $B$ models based on a single strategy. 

\subsection{Need For \sysname}

The hybrid strategy outlined above represents the most robust choice for searching models \textit{independently} from the model pool or possible relations between downstream task and benchmark datasets. Users may nonetheless want to break this independence assumption by incorporating specific knowledge about the filtered models or the relation to benchmark datasets. As an example, if users restrict the pool to models trained on the same upstream dataset, ranking the models based on upstream accuracy often represents a cheap and near-optimal search strategy. Hence, we require a system that gives users the \textit{flexibility} to express a model search query that reflects their most important criteria. The system should also \textit{efficiently} execute search queries, automatically applying query execution optimizations under the hood.
In order to support inexperienced users, we provide detailed guidelines for using the best queries in our system and mimicking best-practices under \url{https://docs.shift.ml/guidelines}.
Finally, bearing in mind that model development is typically an iterative process, if a user is not happy with the returned models or fine-tuning results, she may iterate by changing the data or the query, and then use the newly returned models. Thus, we need a system that can support various search strategies with efficient initial and iterative executions. %

\vspace{-0.5em}
\paragraph{Benchmarking} Comparing new search strategies is a demanding task. Firstly, as explained before, existing search strategies are not supported by the existing model repositories and therefore have to be re-implemented by every researcher. Secondly, when evaluating strategies, users need to have access to a large set of (diverse) models, each of them being fine-tuned on a large set of benchmark datasets. Researchers need support in both aspects to be able to derive new theoretical understanding for existing approaches, or to test new search strategies against existing ones.

\begin{figure}[t]
    \begin{center}
        \includegraphics[width=0.85\linewidth]{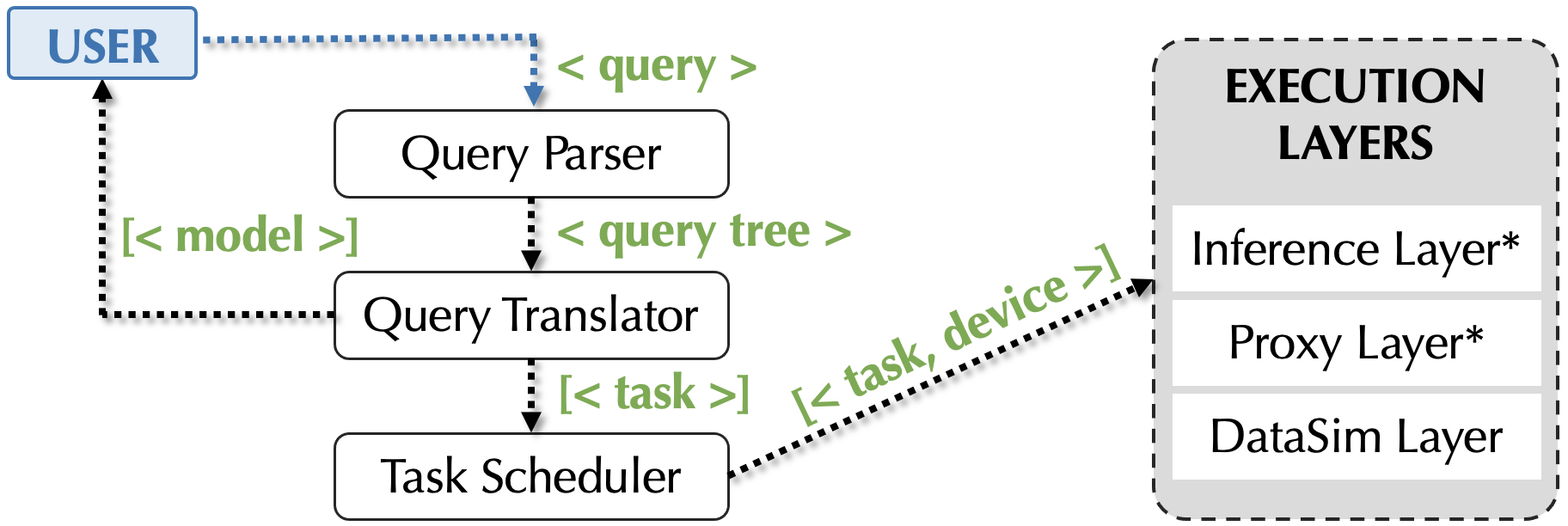}
    \end{center}
    \vspace{-1em}
    \caption{The logical components and different types that connect them that compose \sysname, where \textbf{(*)} represents components that we optimize in Section~\ref{sec:optimization} and~\ref{sec:incremental}.}
    \label{fig:logical_components}
    \vspace{-1em}
\end{figure}

\begin{figure*}[t!]
\centering
\includegraphics[width=0.85\textwidth]{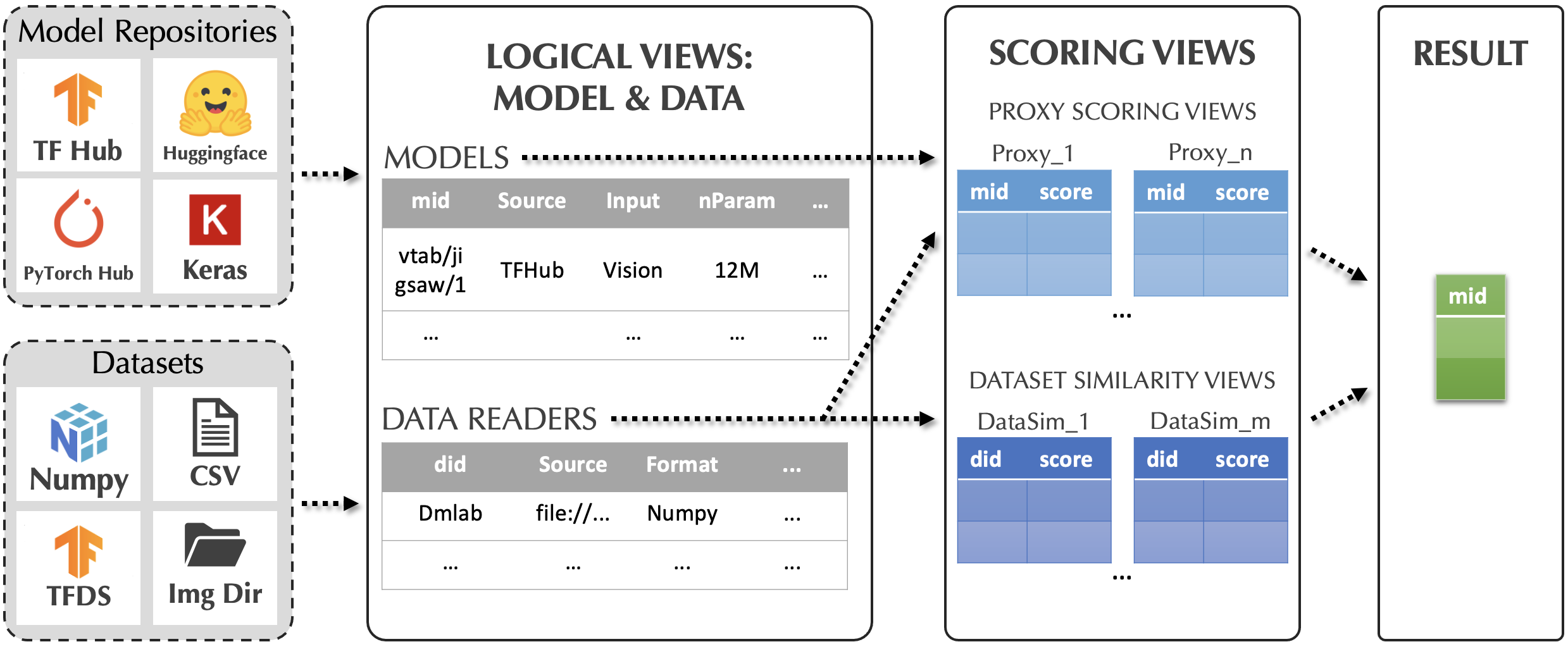}
\vspace{-1em}
\caption{\sysname's Logical Model of Transfer Learning}
\vspace{-1em}
\label{fig:logical_model}
\end{figure*}

\section{Systems Architecture}

We now present the architecture of \sysname, allowing us to efficiently and flexibly execute the search queries outlined in the previous section. We abstract our system into multiple logical components, visualized in Figure~\ref{fig:logical_components}. The labels on the arrows indicate \textit{types} of the input for each component. We start by describing our novel logical model for transfer learning in Section~\ref{sec:data_readers}. We then present our query language and its parser in Section~\ref{sec:query_parser}, and use Section~\ref{sec:scheduler_execution} to describe the scheduler and execution layer.
Figure~\ref{fig:arch} in Appendix~\ref{app:System}\footnote{Accessible via \url{https://github.com/DS3Lab/shift}} provides a complete overview of \sysname. \sysname is designed as a server-client architecture connected by standard HTTP requests. From the client perspective, the input to the system is a \queryname query, and the output is the corresponding result, i.e., a list of models, together with the information whether the query was executed successfully or not. On the server side, \sysname takes the parsed \queryname query tree as the input, and determines how to provide the results. \sysname internally caches intermediate results within and across queries to optimize execution time. We elaborate on caching and other system optimizations in Section~\ref{sec:optimization}.

\subsection{Logical Model of Transfer Learning}
\label{sec:data_readers}

One of the most important challenges 
in building \sysname is to provide a 
clean abstraction for the search 
process of transfer learning, which needs 
to be flexible enough to model most
popular search processes that users are
using in practice, but in the meantime, 
needs to be high-level enough for us
to capture the opportunities of 
system optimizations and 
incremental maintenance. 
In the following, we describe
the \sysname's logical model,
which is based not only on our own 
experience in model search~\cite{renggli2020model}
but also a comprehensive 
survey of existing
popular search strategies~\cite{kornblith2019better, achille2019task2vec, bao2019information, nguyen2020leep, meiseles2020source, puigcerver2020experts, deshpande2021linearized, mensink2021factors}.

\vspace{-0.5em}
\paragraph*{Unified Logical Views
for Model Repositories and Datasets}
During the model search process,
there are two key players: (1) a 
diverse collection of model 
repositories (e.g., 
TF-Hub, PyTorch Hub, and HuggingFace)
and (2) a collection of 
datasets stored in a diverse 
set of formats (e.g., numpy,
CSV, TFDS, etc.).
The starting point of \sysname
is to provide a unified 
view for both models and datasets.
The `\texttt{Models}' view is a relational 
view containing information 
of models across various sources.
As illustrated in Figure~\ref{fig:logical_model},
each model corresponds to a single 
row in the `\texttt{Models}' view, which 
contains ``meta-data'' about its
source, modality, number of parameters, etc. Each model ID also is associated
with various functions to deal 
with tasks such as inference
and fine-tuning, all of which are virtualized in
dockerized environments to 
unify the API differences of
different model repositories (see Section~\ref{sec:scheduler_execution}). 
In practice, we observe that 
having a relational view for 
`\texttt{Models}' is 
particularly useful as 
users often conduct specific 
filtering queries 
over all models (e.g., to only keep
models with \# parameters smaller
than a given constant to ensure 
inference latency). These
can be done via standard SQL queries
over the  `\texttt{Models}' view.
The `\texttt{DataReaders}' view
is a relational view containing 
information of datasets
stored in different formats. 
Each dataset corresponds to a single
row in the `\texttt{DataReaders}'
view, which contains its meta-data.
Each unique \texttt{DataReaders} is 
also associated with 
an \textit{iterator} that 
enumerates $(x, y)$ pairs where
$x$ is a Numpy array 
for a single feature vector and $y$ 
is a Numpy array for a single 
label.

\vspace{-0.5em}
\paragraph*{Benchmark Results View}
In addition to the two views outlined before, \sysname gives access to a view called `\texttt{BenchmarkResults}', acting as a many-to-many join between models and datasets. The view is populated by the administrator of \sysname with the accuracy reported when fine-tuning a fixed model on a fixed benchmark dataset. The system does not distinguish between benchmark and non-benchmark datasets, and treats both as tuples in the `\texttt{DataReaders}' view.
The `\texttt{BenchmarkResults}' is useful for two distinct use-cases. Firstly, when designing meta-learned search queries, users can join and use the results between models and datasets via this view. Secondly, in order to easily compare different search strategies, given a set of $M$ models and $N$ benchmark datasets, for which the post fine-tune accuracies are present in the view, the benchmark module of \sysname can run a search strategy for any of those $N$ benchmark datasets by simulating a state of the system where the corresponding fine-tune results are not present in the `\texttt{BenchmarkResults}' view. The accuracy of the returned model can then be compared to the best model accuracy out of the $M$ ignored fine-tune results.

\vspace{-0.5em}
\paragraph*{\queryname Query}
A \queryname query defines
a unique search strategy for 
transfer learning. In our design,
a \queryname query consists of
two components: (1) 
a collection of ``scoring views''
and (2) 
a generic SQL query over these
scoring views.
Note that all scoring views
are \textit{lazily evaluated
up to various fidelity and precision},
which renders the 
query execution and optimization 
non-trivial and unique for \sysname.

\vspace{-0.5em}
\paragraph*{Lazily Materialized 
top-K Scoring Views}
Given the `\texttt{Models}' view
and the `\texttt{DataReaders}' view,
a user can define scoring views
of two types. The first type,
which we call \textit{proxy scoring 
views} aims to capture 
proxy tasks that are used 
to assess a model's transferability.
A proxy scoring view extends the
SQL syntax and can be defined as

\begin{verbatim}
    CREATE PROXY SCORING VIEW name 
    SQLQUERY         -- e.g., SELECT * FROM Models
                     --       WHERE nParam < 12M
    <SHIFT>          
      ORDER BY ScoringAlgorithm [DESC | ASC] LIMIT K
      [TESTED ON DataReader1]
      [TRAINED ON DataReader2]
      [WITH DataReader3 ...]
    </SHIFT>
\end{verbatim}

\noindent
where \texttt{SQLQUERY} is a standard 
SQL query whose output
has the same schema as the \texttt{Models}
view. In this way, a user can use arbitrary SQL queries
to conduct different filtering
strategies or join with auxiliary information
about models to select models. 
Given all models that \texttt{SQLQUERY}
returns, a \texttt{ScoringAlgorithm}
is associated with a function 
(see Section~\ref{sec:scheduler_execution})
that maps a single \texttt{Model}, and a series of 
\texttt{DataReader} into a real-valued score:
\[
\texttt{ScoringAlgorithm}: \texttt{Model}  [\times \texttt{DataReader} \times ... \times \texttt{DataReader}] \rightarrow \mathbb{R}
\]
By default, a proxy scoring view will 
only return the top-\texttt{K} models
according to the output of
the \texttt{ScoringAlgorithm}. This is 
often the case in most search strategies 
that we see in practice, and as we will see
later, will open up novel opportunities 
for system optimizations.

The second type of scoring views are 
what we call \textit{dataset similarity views},
which are used to compute similarities between
different datasets, an important 
signal in many model search strategies --- a model that
performs well on a \textit{similar} dataset is likely
to perform well on the target dataset, if we are able to 
compute datasets similarities reliably.
A dataset similarity view also extends the
SQL syntax and can be defined as
\begin{verbatim}
    CREATE DATASET SIMILARITY VIEW name 
    SQLQUERY         -- e.g., SELECT * FROM DataReaders
                     --       WHERE Modality = Image
    <SHIFT>          
      ORDER BY DataSimMetric [DESC | ASC] LIMIT K
      TESTED AGAINST TargetDataReader
    </SHIFT>
\end{verbatim}

\noindent
where \texttt{SQLQUERY} is a standard 
SQL query whose output
has the same schema as the \texttt{DataReaders}
view. In this way, a user can use arbitrary SQL queries
to conduct different filtering
strategies or join with auxiliary information
about datasets similar to the selected dataset. 
\texttt{DataSimMetric} is associated with 
a function that computes the similarity between 
a pair of datasets:
\[
\texttt{DataSimMetric}: \texttt{DataSet} \times \texttt{DataSet} \mapsto \mathbb{R}
\]
\noindent
Given all datasets returned by \texttt{SQLQUERY},
\sysname computes its similarity with 
the target \texttt{TargetDataReader}. Similar to 
a proxy scoring view, a dataset similarity view also 
by default keeps the top-\texttt{K} datasets
according to its similarity with the \texttt{TargetDataReader}.

\vspace{-0.5em}
\paragraph*{\queryname Query: Putting Things Together}
Given a collection of scoring views, 
a \queryname query is a generic SQL statement 
querying these views. This allows flexible 
aggregation and voting strategies, and 
are crucial for many search strategies~\cite{renggli2020model}.
We provide several syntax sugars to make 
the query more succinct. When there is 
no ambiguity, we often ignore the 
\texttt{<SHIFT>} and \texttt{</SHIFT>} tags.
Moreover, we also support creating 
scoring views implicitly if such a query is
nested in another SQL query.
As an example, to specify one hybrid search strategy 
developed in~\cite{renggli2020model}:
{\em Return the vision model with fewer than 10M parameters  and
the best
upstream accuracy and another vision model
that has the
best linear classifier accuracy,}
a user can write the following \queryname query:
\begin{verbatim}
   (SELECT ModelId FROM Models
   WHERE Input = 'Vision' AND nParam <= 10M
   ORDER BY UpstreamAccuracy DESC LIMIT 1) Q1
            UNION
   (SELECT ModelId FROM Models
   WHERE Input = 'Vision' AND ModelId NOT IN Q1
   ORDER BY Linear(lr=0.1) ASC LIMIT 1
   TESTED ON TestReader TRAINED ON TrainReader) Q2
\end{verbatim}

We provide the \queryname for the popular search queries Q1-Q5 from Table~\ref{tbl:queries}, as well as a more complex nested query in Appendix~\ref{app:example_queries}.

\vspace{-0.5em}
\paragraph{Tracking Data Changes via Change- and Add-Readers}
To support  efficient incremental executions over data changes, we extend this simple concept of a data reader to a \textit{mutable} reader, allowing a reader to be a composite of an initial data reader and a list of \textit{change-} or \textit{add-readers}. Every change-reader is accompanied by a list of indices of the same length as the initial reader, indicating which samples to replace. The change- and add-readers are then processed in a linear order to build the final mutable data reader.
The two advantages of representing our data as such are (a) extensibility to other data sources and (b) the ability to cache inferred features on a per-reader (initial, change, or add) level. On the flip side, removing samples from a data reader requires users of \sysname to define a new reader, resulting in a new execution from scratch. We plan to support 
deletions in the future. It is also important to note that our current 
data provenance system tracks the changes in a rather naive way. In the future, we should provide 
the support of modern data provenance systems for ML (e.g.,~\cite{grafberger2021mlinspect} and ~\cite{schelterscreening}).

\vspace{-0.5em}
\paragraph*{Supporting New Search Strategies}
One design goal of \sysname and \queryname is to make it easier 
for researchers to provide new search strategies in 
the future. In our current system, this can be done
by registering a new \texttt{ScoringAlgorithm}
or a new \texttt{DataSimMetric}. We are optimistic that
this will support many new search strategies
in the near future (e.g., the ``more robust 
model is more transferable'' strategy that 
just comes up months ago~\cite{deng2021adversarial}, which requires registration of robustness metrics as additional attributes in the dataset). We provide simple tutorials on how to implement new simple strategies online, and evaluate additional search strategies beyond the ones present in the paper (e.g., random sampling or LEEP~\citep{nguyen2020leep}) in Appendix~\ref{app:benchmark_module_results} using our benchmark module.

\subsection{Query Parser and Translator}
\label{sec:query_parser}

Our query parser takes a possibly complex \queryname query as an input, and generates a parsed query tree, where every node is either a proxy scoring or dataset similarity view. The tree is then traversed in a bottom-up approach, by evaluating the leaf nodes until completion before evaluating the parent nodes. 
We leave query tree optimizations such as push-down operations or balancing compute across different nodes for future work.
Every \texttt{SQLQUERY} is evaluated directly against our database. Task-agnostic search strategies represent \queryname queries using neither proxy scoring nor dataset similarity views. Meta-learned task-agnostic queries can be defined by the user by using dataset similarity views and filtering models using populated benchmark fine-tune results in it. For queries relying on proxy scoring views, the query translator will check if the result for the specified method name is known in the corresponding view (i.e., a tuple for each model and the specified readers exists in the database). If so, the system will directly return the results, or pass them to a parent node in the parsed query tree. Otherwise, the system will dispatch a list of \sysname tasks for missing values to the task scheduler. Task-aware search queries are split into two inference tasks per model, one for the test and another for the train data source, and an additional single proxy task per model. The outputs of the inference tasks are used as input to the corresponding proxy task. This allows us to possibly reuse the cached feature representations per pre-trained model. To support better load balancing, we split both inference tasks for partitions of the datasets (c.f., Section~\ref{sec:optimization}). Meta-learned task-aware search queries rely on dataset similarities. If the embedding for the downstream task reader is not in the database, the system will dispatch a DataSim task to compute it. The embeddings are then used to compute the distance between tasks and rank the readers.

\begin{figure}
    \begin{center}
        \includegraphics[width=\linewidth]{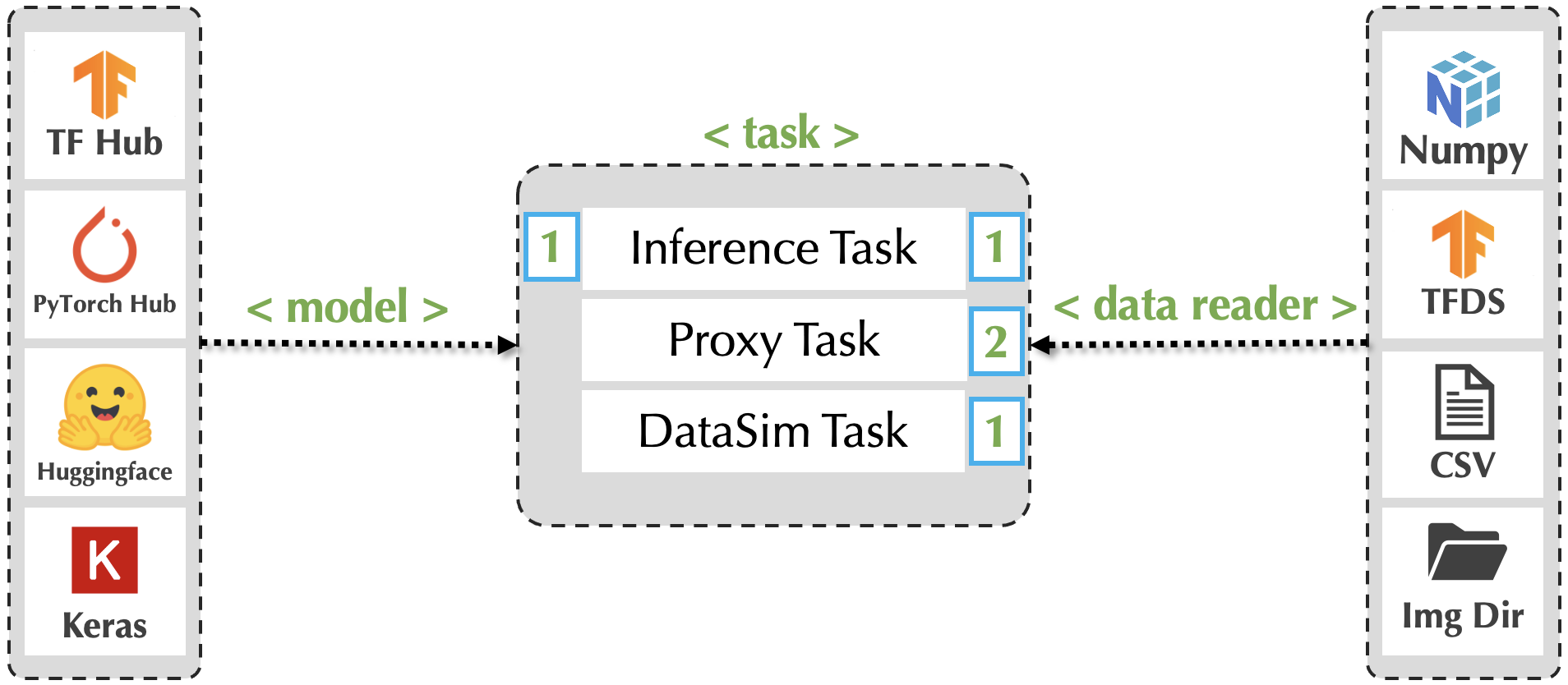}
    \end{center}
    \vspace{-1em}
    \caption{\sysname tasks and their dependencies to model or data sources. The numbers specify on how many objects a specific task depends on (e.g. Proxy tasks depends on 2 data readers).}
    \label{fig:tasks}
    \vspace{-2em}
\end{figure}

\subsection{Task Scheduler and Execution}
\label{sec:scheduler_execution}

A \sysname task represents the smallest computational unit of our system. The scheduler assigns every task to a single hardware device.
We define three different tasks \sysname supports: (1) inference tasks, (2) proxy tasks, and (3) DataSim tasks. The number and types of tasks executed depend on the query provided by the user, then parsed and translated by \sysname (e.g., Q2 will only create inference and proxy tasks). Each of them has dependencies on models or data readers, or both (c.f., Figure~\ref{fig:tasks}).
The scheduler of \sysname is fairly simple. Every task gets assigned, in order of entering the queue, to the next free device, as soon as its dependency tasks are successfully terminated. Every GPU on a single machine represents a device, and a subset of the CPU cores forms another one. Costly inference tasks are assigned to GPU devices, if any are available. Proxy and DataSim tasks are also handled by the CPU.

\vspace{-0.5em}
\paragraph{Inference Task}
To support various pre-trained model sources and frameworks, we define a minimal interface used by inference tasks. All supported model sources require a simple forward function for a batch of samples originating from every data source visualized in Figure~\ref{fig:tasks}. This function returns a 2D Numpy array with each sample (out of $n$) in the data reader representing a row. The feature dimension is determined by the pre-trained model. The resulting extracted features are stored on the disk, and references to the corresponding reader and model combination, using a hash of the earlier, and the model name as a unique identifier of the latter are saved. Formally, the following interface needs to be specified for each combination of models and data sources.

    {
        \small
        \begin{verbatim}
    def extract(pre_trained_model: model,
                source: data_reader) -> 
                np.array(shape=[n,dim])
\end{verbatim}
    }

Most of these 16 possible combinations are natively supported by the frameworks and API (e.g., using the Keras \verb!fit! function for parts of a Keras model, or \verb!KerasLayer! with TF-Hub models), or by casting the data sources into a supported format (e.g, using tensorflow data sources for both Keras and TF Hub). PyTorch Hub models typically only store the models with their PyTorch code and no standardize interface. Every model registered in \sysname therefore needs to specify a hook function to extract the (last-layer) features, which requires us to know the internal structure of the models (i.e., the layer names). Custom trained or fine-tuned models can be exported as Keras models to disk, and then used for subsequent search queries upon registration into the database.

\vspace{-0.5em}
\paragraph{Proxy Task}
By splitting task-aware search queries into inference and proxy tasks, we bypass the requirement of implementing the proxy computation for every combination of model and data source. The proxies are defined over Numpy arrays, where the extracted $n$ train and $m$ test features (ending with \verb!_X!) stem from a model and data source combination after having performed an inference task, and the labels (ending with \verb!_y!) are independent of the models.

    {
        \small
        \begin{verbatim}
    def compute_proxy(train_X: np.array(shape=[n,dim]),
                      test_X: np.array(shape=[m,dim]),
                      train_y: np.array(shape=[n,]),
                      test_y: np.array(shape=[m,])) ->
                      proxy_value: float
\end{verbatim}
    }

We implement two different proxy estimators: (1) the nearest neighbor (NN) accuracy for two different distance functions (cosine dissimilarity, and Euclidean, L2 distance), and (2) a linear classifier accuracy trained with stochastic gradient descent (SGD) and arbitrary hyper-parameters, such as learning rate, L2 regularizer, mini-batch size and number of epochs.

\vspace{-0.5em}
\paragraph{DataSim Task}
A dataset similarity (DataSim) task computed embeddings of a data reader representing a machine learning task (e.g., Task2Vec~\citep{achille2019task2vec} to compute 8512 dimensional vectors). The embedding is then stored along with the meta-data of the reader in the database. Using any distance function (e.g., the non-symmetric one suggested by \citet{achille2019task2vec}), a subset of registered data readers in the database, for which the embeddings are pre-computed, can be ordered and limited in a straightforward fashion. The final meta-learned task-aware search query is then no different from the meta-learned task-agnostic.
We use the code provided by \citet{achille2019task2vec} for running the DataSim tasks.

\section{System Optimizations}
\label{sec:optimization}

\subsection{Successive-Halving (SH)}

In \sysname, all proxy scoring views consist of a top-\texttt{K} 
query over a list of scores; furthermore, each score
is computed as a function over \texttt{DataReader}s
which consist of a set of data examples. 
This structure opens up unique opportunities
for system optimizations --- since many of these
scoring functions are relatively stable with respect
to sub-sampled datasets, we can \textit{approximate} this
top-\texttt{K} view with a scoring function evaluated 
over only a \textit{subset} of data examples.
One key optimization is to estimate the proxy value only for a small subset of the (training) data on most models, and a large fraction of the data only on a small subset of the models. There can be various ways for this. Currently, \sysname uses successive-halving (SH)~\cite{jamieson2016non}, which is invoked as a subroutine inside the popular Hyperband algorithm~\cite{li2017hyperband}.
Algorithm~\ref{alg:SH} in Appendix~\ref{app:sh_algorithm} outlines the algorithm, noting that an \textit{arm} in our context represents a model, and \textit{pulling an arm} corresponds to running inference on more data and estimating the proxy using the extracted features for \textit{all} the data seen by the model so far. In a nutshell, we can summarize the idea of SH as follows: Start by uniformly allocating a fixed initial budget ($B/\log_2(M)$) to all $M$ models and then evaluating their performance. Keep only the better half of the models and repeat this until a single model remains. The algorithm has two different hyper-parameters: a chunk size (how many samples represent an arm pull) and the overall budget $B$. Both can be specified by the user and have an impact on the accuracy of the results and compute time. We propose a chunk size guaranteeing that the last model has processed the entire dataset as a default for \sysname, and use the minimal budget required to return a fixed number of models.

\vspace{-0.5em}
\paragraph{SH Minimal Budget and Chunk Size}
In order to preserve the semantic of the queries whilst performing successive halving (i.e., not sub-sampling the data), we need to guarantee $r_k > 0, \forall k \Longleftrightarrow \frac{B}{L \lceil \log_2(|M|) \rceil} >1$. 
Let us assume a top-q queries with $q=1$ (the derivation can simply be extended to arbitrary values of $q$).
We need $B\geq L\lceil \log_2(|M|) \rceil, \forall L$. The largest $L$ is reached at the first step where $L=|M|$. Hence, we need to have $B\geq |M| \lceil \log_2(|M|) \rceil$.
Conversely, at $k$th step, each remaining model is given $r_k\times C$ additional training samples, in total each remaining model has processed $\sum_{j=0}^{k}r_j\times C$ training samples. When $k=\lceil \log_2(|M|) \rceil-1$, the remaining models have processed $C\times \sum_{j=0}^{\lceil \log_2(|M|) \rceil-1} r_j$ training samples. Hence, the minimal chunk size $C_{min}$ such that the remaining models have processed all training samples is given by $$C_{min}=\frac{N}{\sum_{j=0}^{\lceil \log_2(|M|) \rceil-1} r_j}$$ where $r_j=\lfloor \frac{B}{\lceil M/2^j \rceil \lfloor \log_2(|M|) \rfloor} \rfloor$.

\subsection{Cost Model for Successive Halving}

Successive-halving, while always being able to decrease the amount of 
examples processed, does not always outperform
the baseline strategies in wall clock time. Moreover, as we
show in the experiments, it can sometimes even be slower.
This might seem counter-intuitive, but the main reason lies in hardware accelerators, such as GPUs, which offer the ability to massively parallelize tasks up to a fixed number of samples. For instance, running inference for one sample through a deep neural network requires roughly the same time as a mini-batch of multiple samples. The maximum mini-batch size is often limited by the device memory. Therefore, one should not split very small readers into multiple chunks to speed up a task, rendering SH inappropriate for small datasets.
Additionally, SH introduces sequential dependencies between tasks, which can render the algorithm inefficient or unable to scale to multiple GPUs. One such a cause lies in the repeated model loading, or access to the extracted test features, noticing that we always use the entire test set to estimate a proxy value after an arm pull.

We therefore introduce another key component into \sysname: a cost-based 
decision making process that automatically decides 
whether to use successive halving. To this end,
we derive a cost model for \sysname with and without SH. Our cost model requires a few variables, either pre-computed or available based on the query. Let $N$ be the size of our training dataset, $O$ the size of the test dataset, and $M$ the number of models. Furthermore, let $P$ represent the number of equal devices (e.g., GPUs). The time required to load a model $i$ onto a device is given by $L_i$. Furthermore, the time to load the (training) dataset is represented by $T_N$, whereas the time required to load the inferred test representations for the model $i$ is given by $T_i$. We simplify this by assuming a global $T_O$, since the representations only differ in their dimensions. We neglect the time to load the raw test dataset as this is equal regardless of the optimization. The time to run inference for $k$ samples and model $i$ on the device is given by $I_i(k)$, which we assume to be linear, hence $I_i(k) = I_i k$, for some constant $I_i$. The time to compute a proxy $E_{Proxy}^i(k)$ follows the same principle, although we assume that it is independent of the model (neglecting the dimension of the representations), hence $E_{Proxy}^i(k) = E_{Proxy} k$.
The cost for running a top-1 query without SH on multiple GPUs is computed with

\vspace{-1em}
\begin{align*}
    T_{\text{w/o}} :=\frac{1}{P}\sum_{i=1}^{M} & \Big( \underbrace{T_N + T_O}_{\text{Load data}} + \underbrace{2L_i}_{\text{Load model}} + \underbrace{I_i O}_{\text{Test inference}} \\ & + \underbrace{I_i N}_{\text{Train inference}} + \underbrace{E_{Proxy} N}_{\text{Proxy estimation}} \Big),
\end{align*}

Notice that we need to load the train dataset again every time when there is a new model (request), as every task is executed independently. The counterpart, running a top-1 query with SH on multiple GPUs, where we assume perfect parallelization which is harder to achieve for heterogeneous models and small chunks $C$, is given by

\vspace{-1em}
\begin{align*}
    T_{\text{w/}} := & \underbrace{\frac{1}{P}\sum_{i=1}^{M} \left(L_i + I_i O \right)}_{\text{Test inference}} + \underbrace{\sum_{k=1}^{\lceil \log_2(M) \rceil} \frac{1}{\min\left(P, \vert S_k \vert\right)}\ \times }_{\text{SH iterations}} \\ &  \Big( \underbrace{\sum_{j \in S_k} \left(L_j + T_N + I_j C r_k \right)}_{\text{Train inference}} + \underbrace{\sum_{j \in S_k} \Big( T_O + E_{Proxy} \sum_{l=1}^{k} \left( C r_l \right) \Big)}_{\text{Test load and proxy estimation}} \Big),
\end{align*}

where $C$ is specified by the user or taken as $C_{min}$ (c.f., Section~\ref{sec:optimization}), $S_k$ and $r_k$ are taken from Algorithm~\ref{alg:SH}. Clearly, the sets $S_k$ of models \textit{surviving} during the SH algorithm have an impact on the runtime. Following the trend of larger and slower models surviving the longest, we define $S_k$ to be the set of $k$ models with the largest inference time $I_j$ for all $j \in [M]$.
\sysname will use this cost model (i.e., the minimum of $T_{\text{w/o}}$ and $T_{\text{w/}}$)  to automatically decide, based on system's specifications (e.g., hardware devices and model inference times), number of samples, and models in the restricted pool, whether to use the SH algorithm.
\vspace{-0.5em}
\paragraph{SH for Other Queries}
We only use the SH optimization for task-aware queries for which a fraction of the samples can be used for \textit{ranking} models with high confidence. Classifier accuracies (e.g., NN or linear classifier) are known to satisfy this property~\citep{rimanic2020convergence}. Furthermore, the overhead of running multiple sequential tasks when using the SH algorithm is kept small for task-aware queries, thanks to the two-stage approach (i.e., the inference and proxy estimation phase), which is not the case for the other search queries. Inference tasks are typically much more time-consuming compared to the proxy estimation. Nonetheless, when \textit{pulling} an arm an additional time, the data from the previous arm pulls are not required to be run through the network again as they can be fetched from the disk for the subsequent proxy estimation tasks.

\subsection{Other Optimizations}

\paragraph{Caching}

Caching is crucial for rapid incremental query executions. 
\sysname internally caches dataset similarities, feature vectors, and proxy values in order to reuse intermediate results within and across queries as much as possible.
When dispatching task requests, \sysname ensures that only necessary requests are executed. For example, if a user sends a request for a proxy estimator (e.g., the nearest neighbor accuracy) and a single model, \sysname will create two inference requests, one for the test and train reader, and a proxy estimator request. The former requests turn the input data into feature vectors and store them on the disk. If, at a later stage, the user requests another proxy estimator for the same model (e.g., the linear classifier accuracy), \sysname will notice that it can reuse the cached feature vectors. Therefore, \sysname will only dispatch a proxy estimation request, which uses the feature vectors to calculate the proxy value. The dispatched requests are then handled by the task scheduler asynchronously. Once a certain request is done, the results will be written into the database and ultimately returned to the user upon running the same query.

\vspace{-0.5em}
\paragraph{Load-Balancing}
To automatically load-balance heterogeneous workloads, mainly stemming from large discrepancies in inference times between models, to multiple hardware devices, we automatically split readers for every inference task. The number of partitions is equal to the number of GPU devices, unless a reader is smaller than a fixed threshold. This guarantees fast executions especially when running inference on small readers (i.e., change-readers).

\section{Incremental Executions}\label{sec:incremental}

If a user is not satisfied with the results of the query, either in terms of the proxy values or other properties of the models returned (e.g., diversity or downstream performance), she typically iterates by incrementally executing another, often similar query, or re-run the same query performing one of the following changes:
\begin{enumerate}[noitemsep]
    \item add a model to the database,
    \item change the data (features or labels) to test or train readers by using change-readers,
    \item add data to the test or train reader using an add-reader.
\end{enumerate}

\vspace{-0.5em}
\paragraph{Incremental Scenario 1: Changing Data or Queries, or Adding Models}
Changing data (features or labels), a query, or adding a model to the database will naturally benefit for the caching mechanisms we introduced in the previous section. For example, a task-aware search query will only require \sysname to dispatch two inference tasks and a single proxy task per additional model, retrieving the other results directly from the database. When using the SH optimization, the same idea of reusing the caches for intermediate results (features and proxy estimation values) applies. Data manipulations are slightly more involved. Based on the defined mutable data reader concept in Section~\ref{sec:data_readers}, costly inference operations only need to be executed for the changed features. The cheaper proxy task is then evaluated on the final mutable reader (i.e., after having iterated over all change- and add-readers) only once, or for every subsequent iteration in the SH algorithm.

\begin{figure}[t!]
    \centering
    \includegraphics[width=0.48\textwidth]{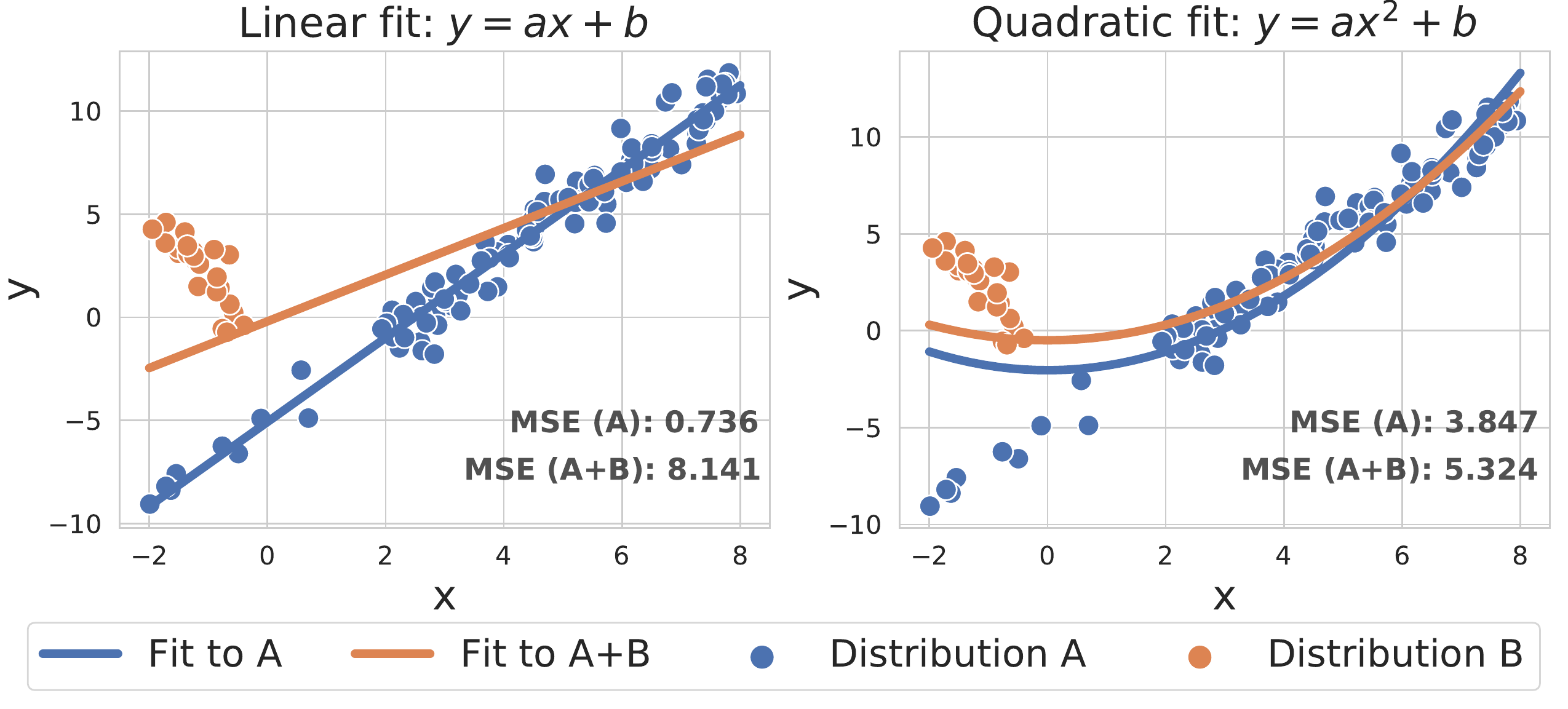}
    \vspace{-2em}
    \caption{Two models (linear and quadratic), for which the order (i.e., based on the minimal MSE) changes if they have access to data only from the first distribution (A), compared to both distributions (A+B).}
    \label{fig:example_add_data}
    \vspace{-1em}
\end{figure}

\vspace{-0.5em}
\paragraph{Incremental Scenario 2: Adding Data}
Adding data to the test reader requires us, similarly to changing data, to run inference on all models of the query using these additional data samples, and then rerun the proxy estimation value for every intermediate step in the SH algorithm. However, blindly appending \textit{training} data at the end of the data reader can result in an undesired behavior. Concretely, all models that did not process the entire dataset (i.e., were eliminated in the SH algorithm based on a subset the head of the data reader) will not benefit from the appended samples. This can be problematic if the added data stems from a different distribution, thus possibly \textit{eliminating} better arms (i.e., models) while running the SH algorithm. 
We construct and illustrate one such synthetic case in Figure~\ref{fig:example_add_data}, where we assume to have two models (a linear one on the left and a quadratic one on the right), and data coming from two distributions (A in blue and B in orange). If the models were ranked on the basis of the errors of distribution A alone, the quadratic model would be eliminated. If then data from distribution B is appended to the reader (via an add-reader), the winning linear model from before would actually be inferior to the quadratic one. We show another real-world example in Section~\ref{sec:experiments:incremental}.

To address this issue, ideally, one would randomly reshuffle the full data reader with appended data. Despite being favorable from a statistical point of view, managing the cache and preventing a complete re-execution is far from trivial.

In \sysname we \textit{uniformly distribute} the new samples as an
alternative strategy. This enables high performance and results in a small difference compared to the fully shuffle approach from a statistical point of view. As a justification 
of its statistical property,
let us assume that we have an initial data reader of size $N$ together with an add-reader that contains $\alpha N$ samples, with $\alpha \geq 0$. Moreover, assume that the chunk sizes used by the SH algorithm on the initial data reader is of size $\beta N$, with $0 \leq \beta \leq 1$. 
We want to compare the two strategies: (1) randomly inserting the new samples anywhere between the existing ones and re-partitioning the samples into buckets afterwards, and (2) uniformly distributing the new samples amongst all existing buckets. Notice that the number of buckets remains constant after handling the new samples, thus increasing the size of the buckets to $\beta N \left(1+\alpha\right)$. Furthermore, it is obvious that from an implementation perspective, the second strategy is superior to the first, whereas the first strategy introduces less bias into the sampling process. However, both approaches yield the same number of samples from both distributions in expectation. We define a random variable $C$, which represents the number of initial samples in the first bucket when the first strategy is run. Coincidentally, $C$ follows a hyper-geometric distribution with an expectation of
$\mathbb{E}\left[C\right] = \beta N,$
which is exactly the number of samples in the same first bucket we get when applying the second strategy.

\section{Evaluation}
\label{sec:experiments}

\subsection{Experimental Setup}

We conduct our experimental study next on computer vision and NLP classification tasks, representing the most prominent applications of transfer learning~\citep{zhai2019visual}. Nonetheless, \sysname is flexible and the code-base supports workload beyond these modalities.

\vspace{-0.5em}
\paragraph{Models} We compile a diverse list of 100 computer vision and 60 NLP models, whose details, including additional configuration such as inference time for the GPU type needed in the next paragraph, are given in Appendix~\ref{sec:app:model_details}. To restrict the search space for fine-tuning, we follow \citet{zhai2019visual} and train the models for 20 epochs, using a mini-batch size of 16, momentum of 0.9, and learning rate of 0.01, with the Adam optimizer.

\vspace{-0.5em}
\paragraph{Datasets} We conduct our experiments with 3 vision datasets representing different downstream tasks: (1) Oxford Flowers 102~\citep{nilsback2008automated} (Flowers) with 1K training and 6K test samples, (2) CIFAR-100~\citep{krizhevsky2009learning} (CIFAR), with 50K training and 10K test samples, and (3) Dmlab~\citep{zhai2019visual} with 65K training and 23K test samples. We furthermore use two NLP datasets from the glue benchmark~\citep{wang2018glue}: (1) cola, with 8.5K training and 1K test samples, and (2) sst2, with 67K training and 1.8K test samples.

\vspace{-0.5em}
\paragraph{Hardware} We use a GPU cluster (single machine) with 8 NVIDIA TITAN Xp for \sysname. The system is configured to either use a single or all eight GPUs. For fine-tuning models, we use a different cluster with slightly more performant NVIDIA GeForce RTX 2080 Ti GPUs.

\vspace{-0.5em}
\paragraph{Queries} We evaluate the performance of the 3 queries Q2-Q4 from Table~\ref{tbl:queries}, with a focus on computational efficiency. Q1 is included in Q4 and directly evaluated against the database and therefore omitted in the experiments. Q5 uses the Task2Vec code to find the nearest benchmark task. To successfully apply Q5, or any meta-learned query, we require a large set of benchmark datasets, all fine-tuned on all modules. We provide results for both meta-learned queries Q5 and Q7 using our benchmark module in Appendix~\ref{app:benchmark_module_results}. Whenever the SH algorithm is used, depending on the cost model and specified in the experiments, we set the budget and chunk size to be minimal according to Section~\ref{sec:optimization}. This ensures that the semantics of the queries are kept intact, i.e. the data is not sub-sampled.

\subsection{End-to-end Performance}

\begin{table}[t!]
\caption{Execution time for fine-tuning (FT) all the models via enumeration compared to running Q2-Q4 using \sysname with and without automatic optimization (AO).}
\vspace{-1em}
\label{tbl:finetune_vs_shift_runtime}
\small
\begin{tabular}{rrrllll}
\hline
                         &                     &                         & \multicolumn{2}{c}{1 GPU}                                                                                                                                      & \multicolumn{2}{c}{8 GPU}                                                                                                                                      \\
                         &                     &                         & \multicolumn{1}{r}{\begin{tabular}[c]{@{}r@{}}Runtime\\ (Hours)\end{tabular}} & \multicolumn{1}{r}{\begin{tabular}[c]{@{}r@{}}Speedup\\ (vs. FT)\end{tabular}} & \multicolumn{1}{r}{\begin{tabular}[c]{@{}r@{}}Runtime\\ (Hours)\end{tabular}} & \multicolumn{1}{r}{\begin{tabular}[c]{@{}r@{}}Speedup\\ (vs. FT)\end{tabular}} \\ \hline
\multirow{7}{*}{CIFAR}   & FT                  &                 & 251.8                                                                         &                                                                                & 31.5                                                                          &                                                                                \\
                         & \multirow{2}{*}{Q2} & w/o AO                & 8.7                                                                           & 28.8x                                                                          & 1.6                                                                           & 19.8x                                                                          \\
                         &                     & w/ AO & 5.9                                                                           & \textbf{42.9x }                                                                         & 1.3                                                                           & \textbf{24.1x }                                                                         \\
                         & \multirow{2}{*}{Q3} & w/o AO                & 9.1                                                                           & 27.6x                                                                          & 1.6                                                                           & 19.2x                                                                          \\
                         &                     & w/ AO & 5.8                                                                           & \textbf{43.4x }                                                                         & 1.3                                                                           & \textbf{24.3x  }                                                                        \\
                         & \multirow{2}{*}{Q4} & w/o AO                & 7.9                                                                           & 31.7x                                                                          & 1.0                                                                           & \textbf{30.0x}                                                                          \\
                         
                         &                     & w/ AO & 5.6                                                                           & \textbf{45.0x  }                                                                        & 1.2                                                                           & 25.6x                                                                          \\
                         
                         \hline
\multirow{7}{*}{Dmlab}   & FT                  &                 & 314.8                                                                         &                                                                                & 39.4                                                                          &                                                                                \\
                         & \multirow{2}{*}{Q2} & w/o AO                & 14.3                                                                          & 22.0x                                                                          & 2.4                                                                           & 16.2x                                                                          \\
                         &                     & w/ AO & 9.6                                                                           & \textbf{32.7x  }                                                                        & 2.0                                                                           & \textbf{19.7x  }                                                                        \\
                         & \multirow{2}{*}{Q3} & w/o AO                & 14.6                                                                          & 21.6x                                                                          & 2.5                                                                           & 16.0x                                                                          \\
                         &                     & w/ AO & 9.9                                                                           & \textbf{31.8x }                                                                         & 1.9                                                                           & \textbf{21.2x  }                                                                        \\
                         & \multirow{2}{*}{Q4} & w/o AO                & 13.3                                                                          & 23.7x                                                                          & 1.7                                                                           & \textbf{22.6x }                                                                         \\
                         &                     & w/ AO & 9.3                                                                           & \textbf{33.9x   }                                                                       & 2.0                                                                           & 20.0x                                                                          \\ \hline
\multirow{7}{*}{Flowers} & FT                  &                 & 9.6                                                                           &                                                                                & 1.2                                                                           &                                                                                \\
                         & \multirow{2}{*}{Q2} & w/o AO                & 2.9                                                                           & \textbf{3.3x}                                                                           & 0.4                                                                           & \textbf{2.8x }                                                                          \\
                         &                     & w/ AO & 2.9                                                                           & \textbf{3.3x }                                                                          & 0.4                                                                           & \textbf{2.8x  }                                                                         \\
                         & \multirow{2}{*}{Q3} & w/o AO                & 3.1                                                                           & \textbf{3.1x}                                                                           & 0.4                                                                           & \textbf{2.7x }                                                                          \\
                         &                     & w/ AO & 3.1                                                                           & \textbf{3.1x }                                                                          & 0.4                                                                           & \textbf{2.7x   }                                                                        \\
                         & \multirow{2}{*}{Q4} & w/o AO                & 2.8                                                                           & \textbf{3.4x}                                                                           & 0.4                                                                           & \textbf{3.1x}                                                                           \\
                         &                     & w/ AO & 2.8                                                                           & \textbf{3.4x }                                                                          & 0.4                                                                           & \textbf{3.1x  }                                                                         \\ \hline
\end{tabular}
\vspace{-1em}
\end{table}

\begin{figure}[t!]
    \centering
    \includegraphics[width=0.48\textwidth]{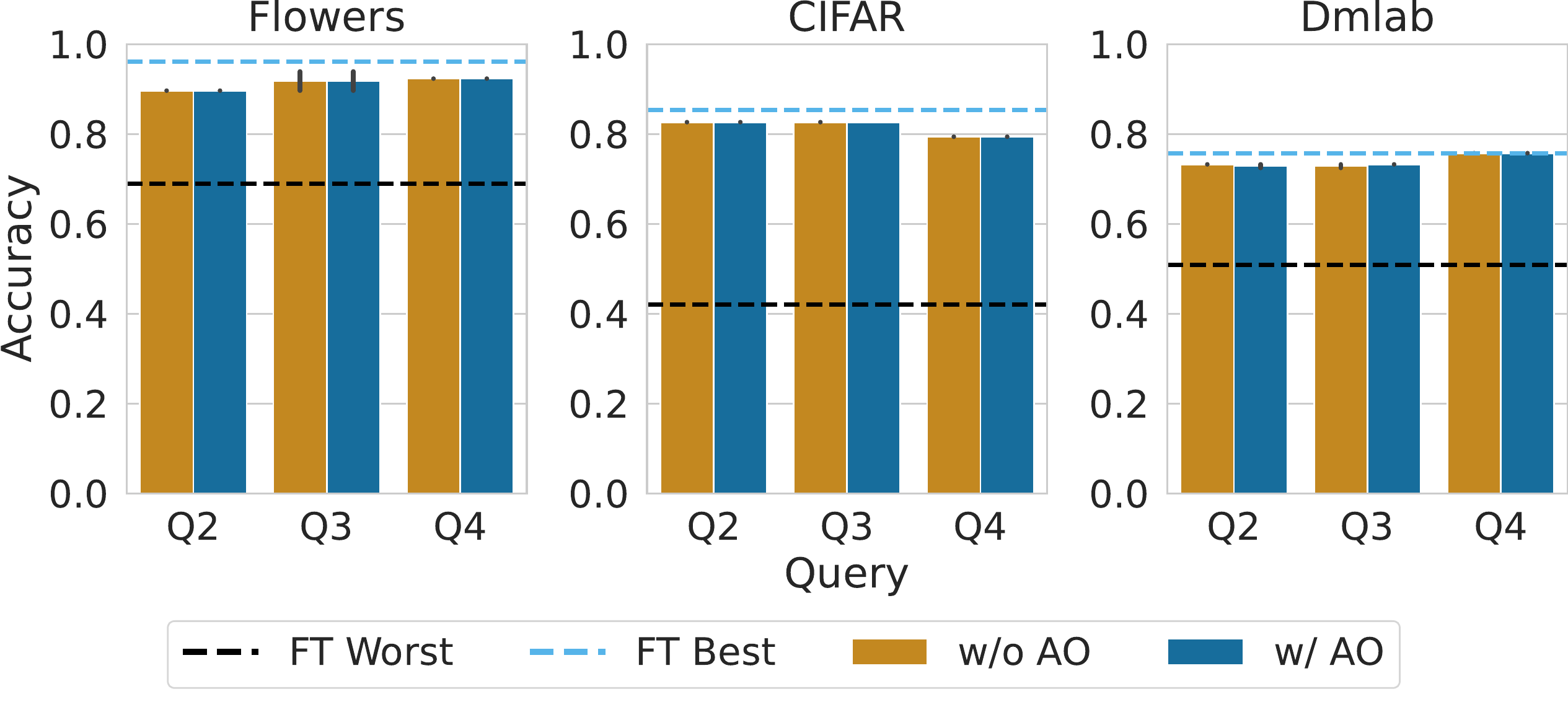}
    \vspace{-2em}
    \caption{Fine-tune (FT) accuracy of returned model for queries with and without automatic optimization (AO). The variance illustrates the min and max over 4 independent runs, showing some fluctuations for the small Flowers dataset and the linear proxy, which is sensitive to its hyper-parameters.}
    \label{fig:finetune_vs_shift_accuracy}
    \vspace{-1em}
\end{figure}

\begin{figure}[t!]
        \centering
        \includegraphics[width=0.48\textwidth]{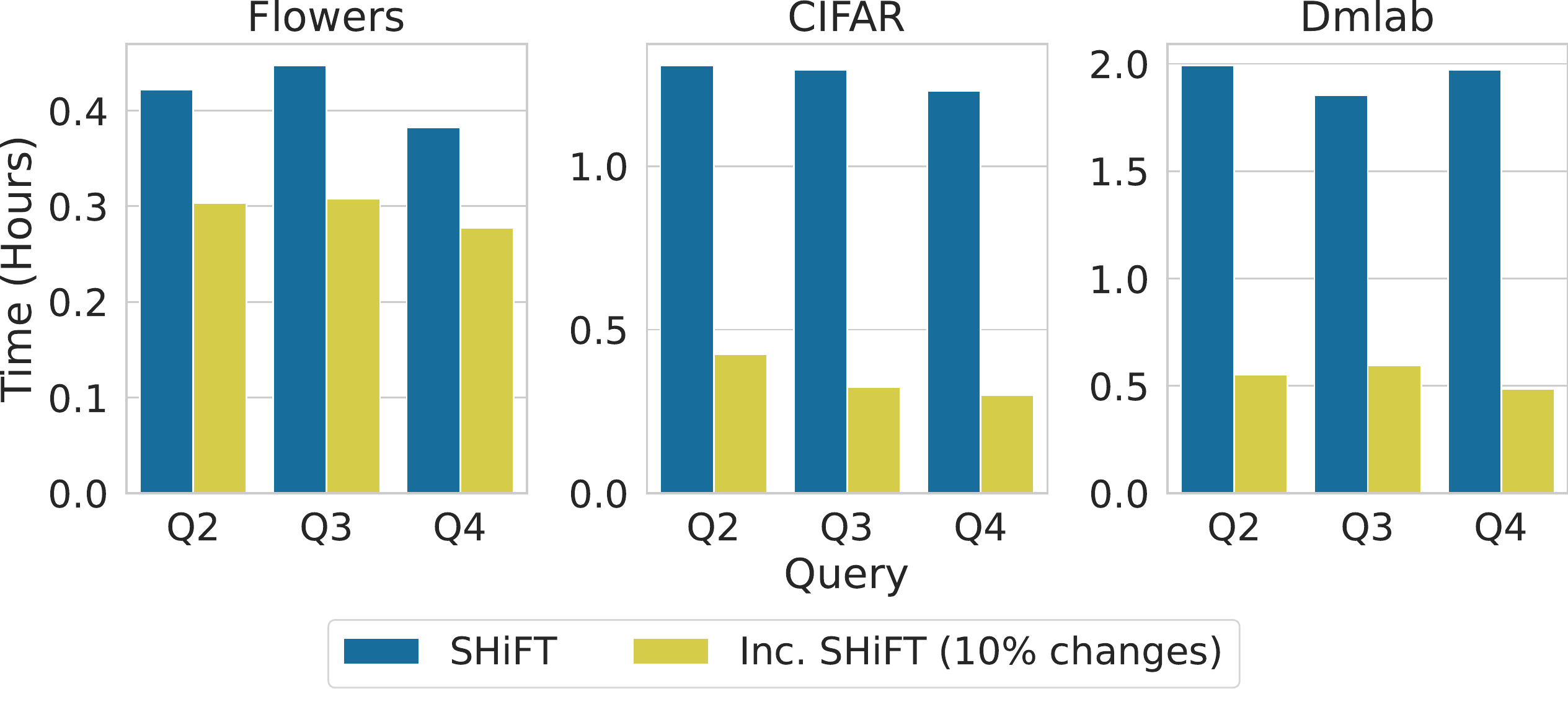}
    \vspace{-2em}
    \caption{Incremental execution using \sysname on 8 GPUs. 10\% of the samples are randomly replaced.}
    \label{fig:incremental_10_percent}
    \vspace{-1em}
\end{figure}

We start by validating the end-to-end performance of \sysname on computer vision task. Due to the space limitation we present the corresponding NLP results to Appendix~\ref{sec:app:nlp_results}. Table~\ref{tbl:finetune_vs_shift_runtime} compares the runtimes of fine-tuning all the models, the method which we call \textit{enumerate}, to running and using the output of queries Q2-Q4 on \sysname, with and without automatic optimization (AO). Our cost model, which we validate later in this section, suggests not to use SH for the small Flowers dataset, which is why the SH optimization is enabled only for CIFAR and Dmlab.  When it comes to accuracy, Albeit being up to 1.5 orders of magnitude faster (c.f., Table~\ref{tbl:finetune_vs_shift_runtime}), Figure~\ref{fig:finetune_vs_shift_accuracy} shows that the queries manage to retrieve near-optimal models for all datasets (i.e., suffering from very small regret). Furthermore, the SH optimizations for CIFAR and Dmlab retain the query semantics, not affecting the accuracy over the baseline for each query.
Remember that the main focus of this \sysname is to support a large set of possibly complex queries independent of the actions on the returned model (e.g., fine-tuning) as efficiently as possible.
For a complete empirical study that compares different search queries, we refer to our companion work~\citep{renggli2020model} and Section~\ref{sec:benchmark_results}, where we benchmark different strategies using \sysname. Finally, Figure~\ref{fig:incremental_10_percent} shows the runtime for incrementally running \sysname on 10\% randomly changed samples, leading to significant speedups for larger datasets, where the GPUs are fully utilized.

\subsection{Scalability of \sysname}

\begin{figure}[t!]
    \centering
    \begin{subfigure}[b]{0.225\textwidth}
        \centering
        \includegraphics[width=\textwidth]{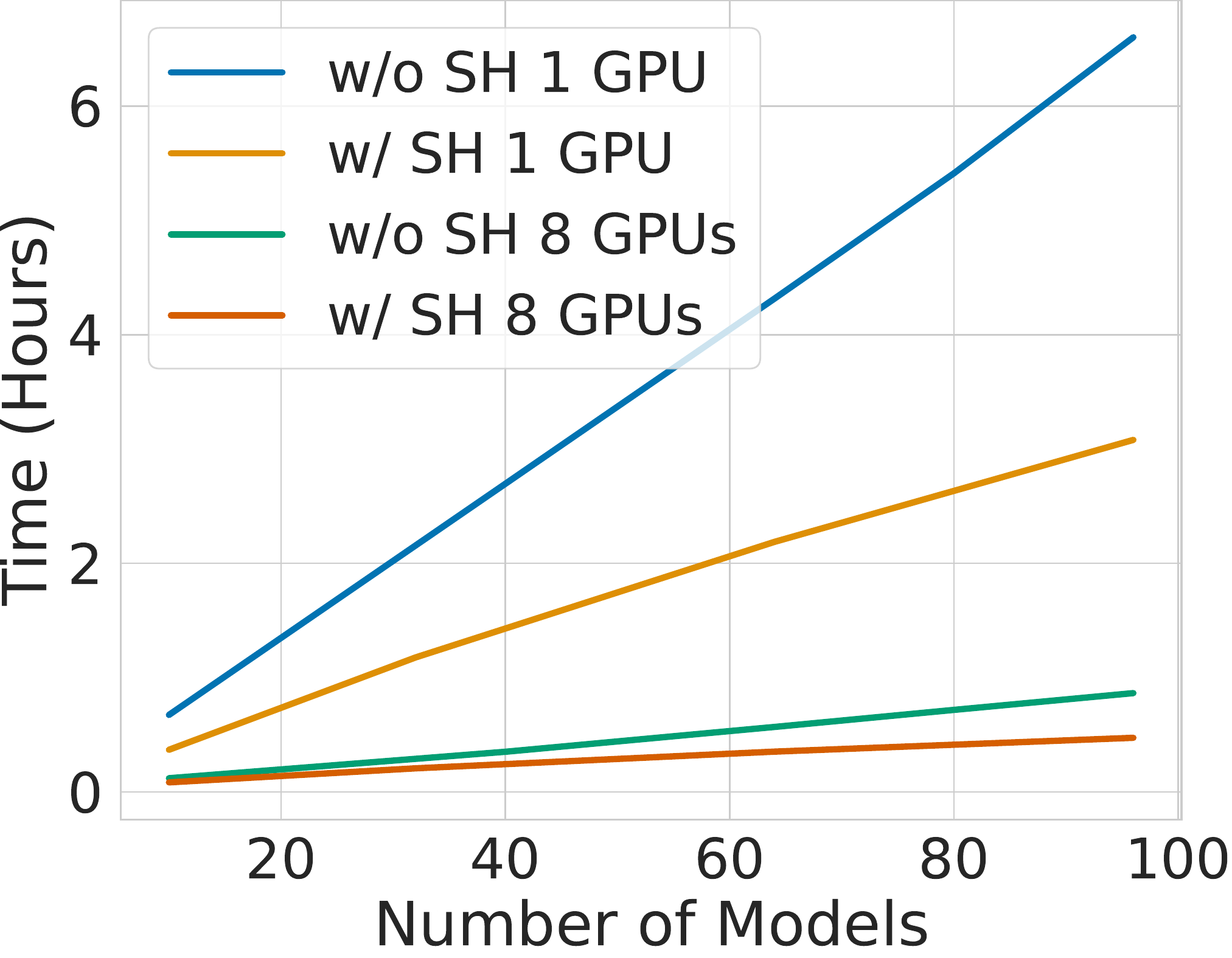}
        \caption{Execution times of Q2 for CIFAR.}
        \label{fig:scaling_models_shift}
    \end{subfigure}
    \hfill
    \begin{subfigure}[b]{0.245\textwidth}
        \centering
        \includegraphics[width=\textwidth]{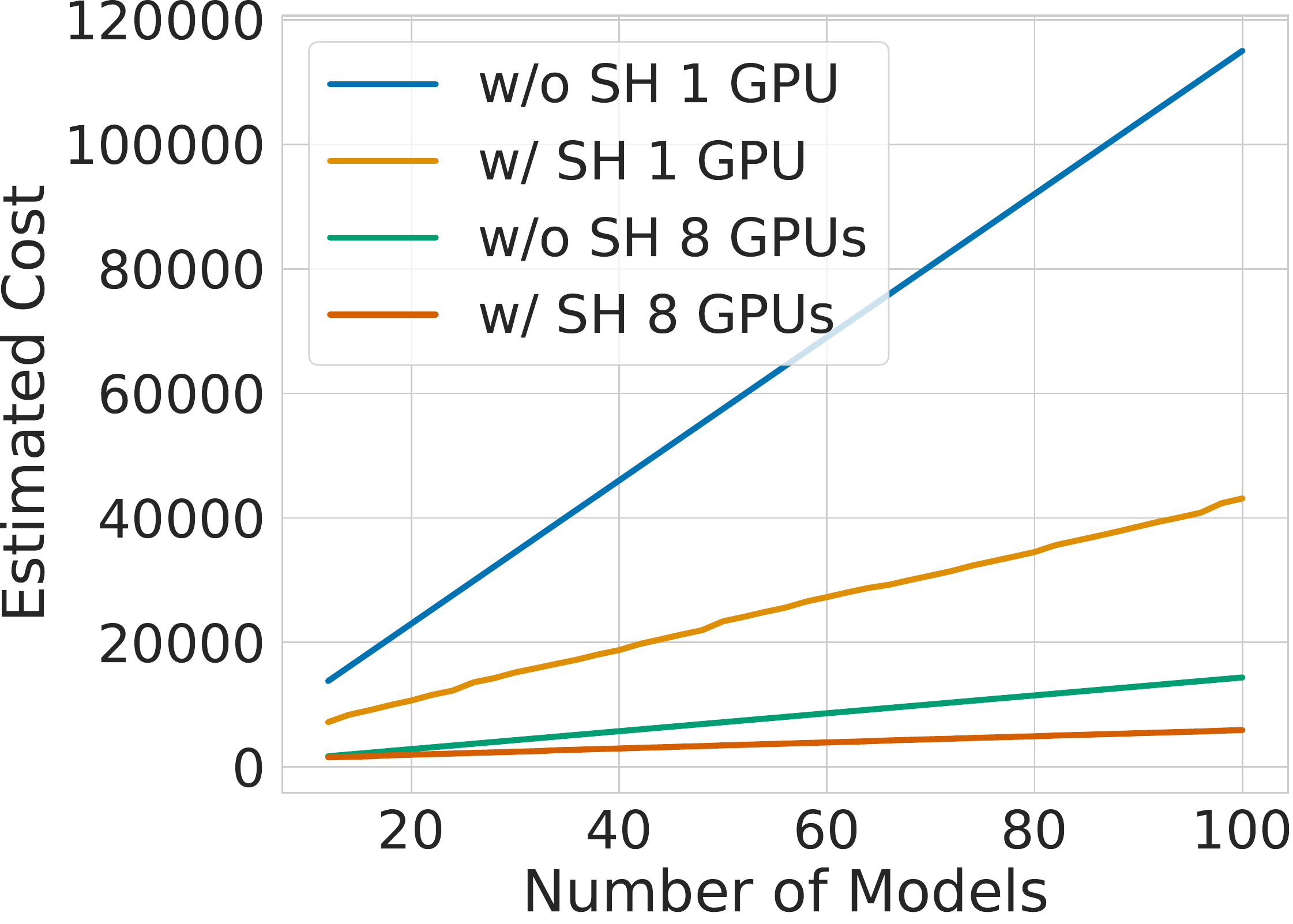}
        \caption{Cost model for Q2 on 50K training samples.}
        \label{fig:scaling_models_costmodel}
    \end{subfigure}
    \vspace{-2em}
    \caption{Increasing number of (homogeneous) models. The ResNet-101 V2 model is replicated for the experiments and the cost model.}
    \label{fig:scaling_models}
    \vspace{-1em}
\end{figure}

With the experimental setting described, we implicitly analyze the scaling behavior of \sysname for an increasing number of GPUs (one and eight), and (training) samples (1K, 50K, and 62K). Increasing the number of models is analyzed in Figure~\ref{fig:scaling_models} (left), where we deliberately chose a homogeneous setting of replicating the same model architecture multiple times.

\subsection{Cost Model: SH Trade-offs}

\begin{figure}[t!]
    \centering
    \begin{subfigure}[b]{0.48\textwidth}
        \centering
        \includegraphics[width=\textwidth]{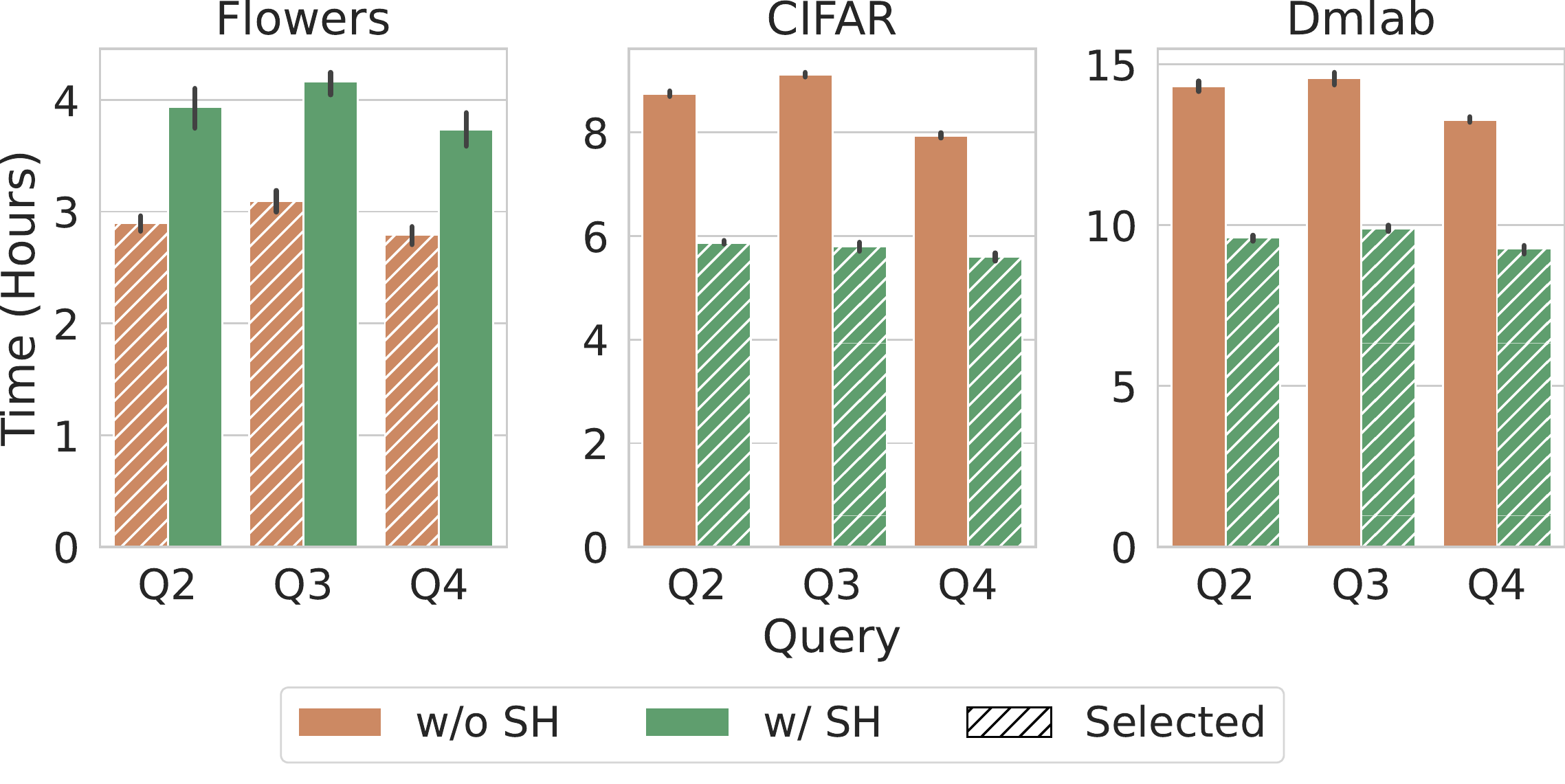}
        \vspace{-1em}
        \caption{1 GPU}
        \label{fig:cost_model_1gpu}
    \end{subfigure}
    \hfill
    \begin{subfigure}[b]{0.48\textwidth}
        \centering
        \includegraphics[width=\textwidth]{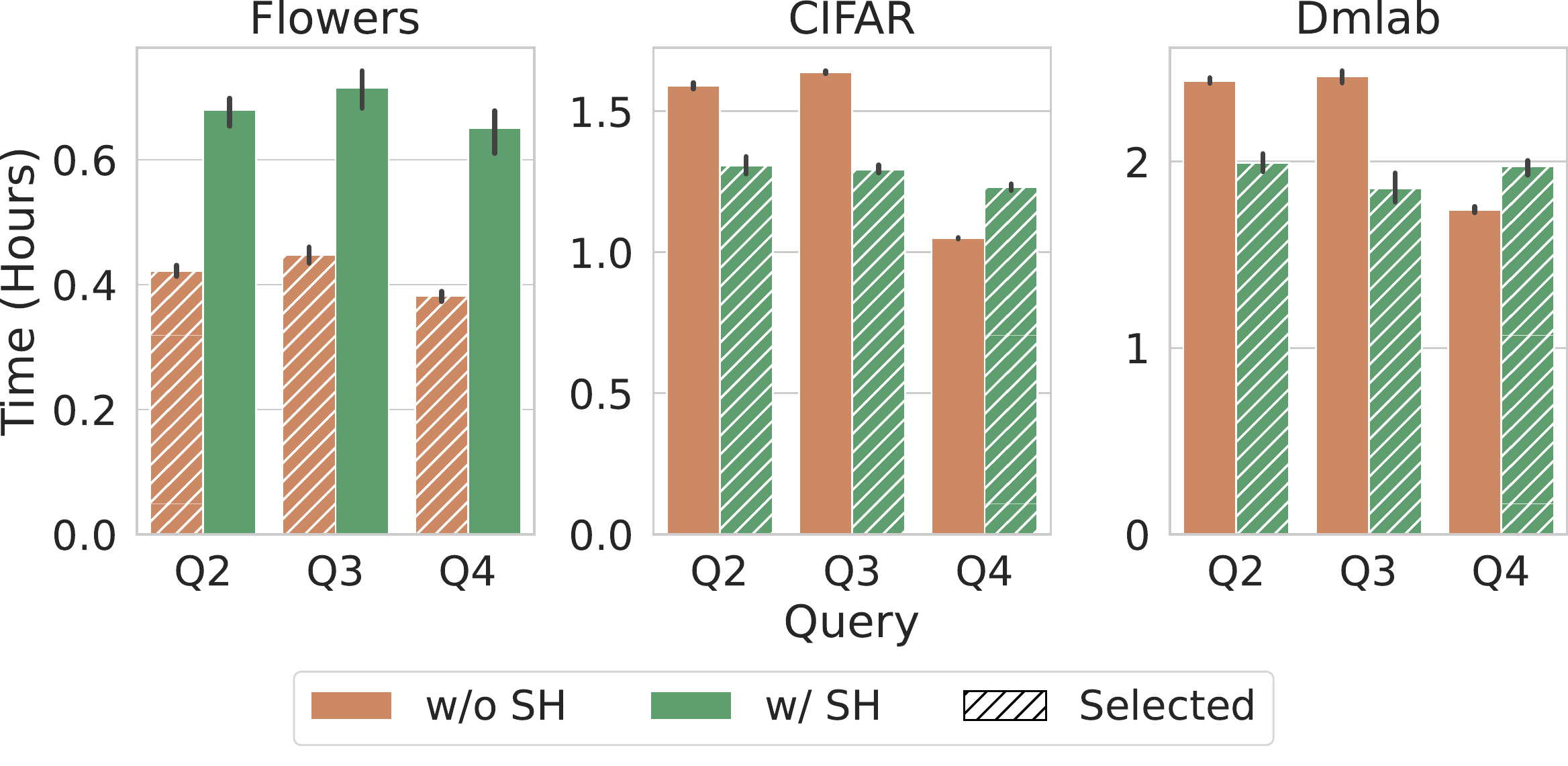}
        \vspace{-1em}
        \caption{8 GPUs}
        \label{fig:cost_model_8gpu}
    \end{subfigure}
    \vspace{-2em}
    \caption{Execution times for all settings. The variance in black illustrates the min and max over 4 independent runs, showing little fluctuation. The hatched bar indicates the selected plan based on the cost model.}
    \label{fig:cost_model}
    \vspace{-1em}
\end{figure}

Figure~\ref{fig:scaling_models} (right) validates the relative performance of our cost model for a set of homogeneous models. In Figure~\ref{fig:cost_model} we show the different runtimes with and without SH for 1 and 8 GPU and all queries along with the configuration picked by \sysname on the 100 diverse models. On the Flowers dataset, our cost model accurately predicts the relative improvements to be expected when not using SH over using SH for a single (1.68x)
and multiple GPUs (1.95x).
 On the larger datasets, CIFAR and Dmlab, the ratio for using 1 GPU (both 1.8x)
is validated by our experiments. For multiple GPUs and the larger datasets, \sysname predicts that SH should outperform non-SH by 
~1.2x, e.g. on CIFAR for all three queries, while the performance only matches Q2 and Q3, as visible in Figure~\ref{fig:cost_model}. The hybrid query Q4 removes a very large model, however, our cost model overestimates the benefits of SH in such a case. The reason lies in the heterogeneity of the models (e.g., the largest model takes up almost 10\% of the overall inference time) and the order of execution currently neglected in the cost model for both with and without SH on multiple GPUs. This explains the gaps visible in Figure~\ref{fig:cost_model_8gpu}. Fusing these aspects into the cost model requires runtime-specifc variables (e.g., information about other queries executed in parallel), making it much more complex and potentially introducing extra latency when executing a query. It is therefore left as future work.

\subsection{Incremental Execution}\label{sec:experiments:incremental}

\begin{figure}[t!]
    \centering
    \begin{subfigure}[c]{0.225\textwidth}
        \centering
        \includegraphics[width=\textwidth]{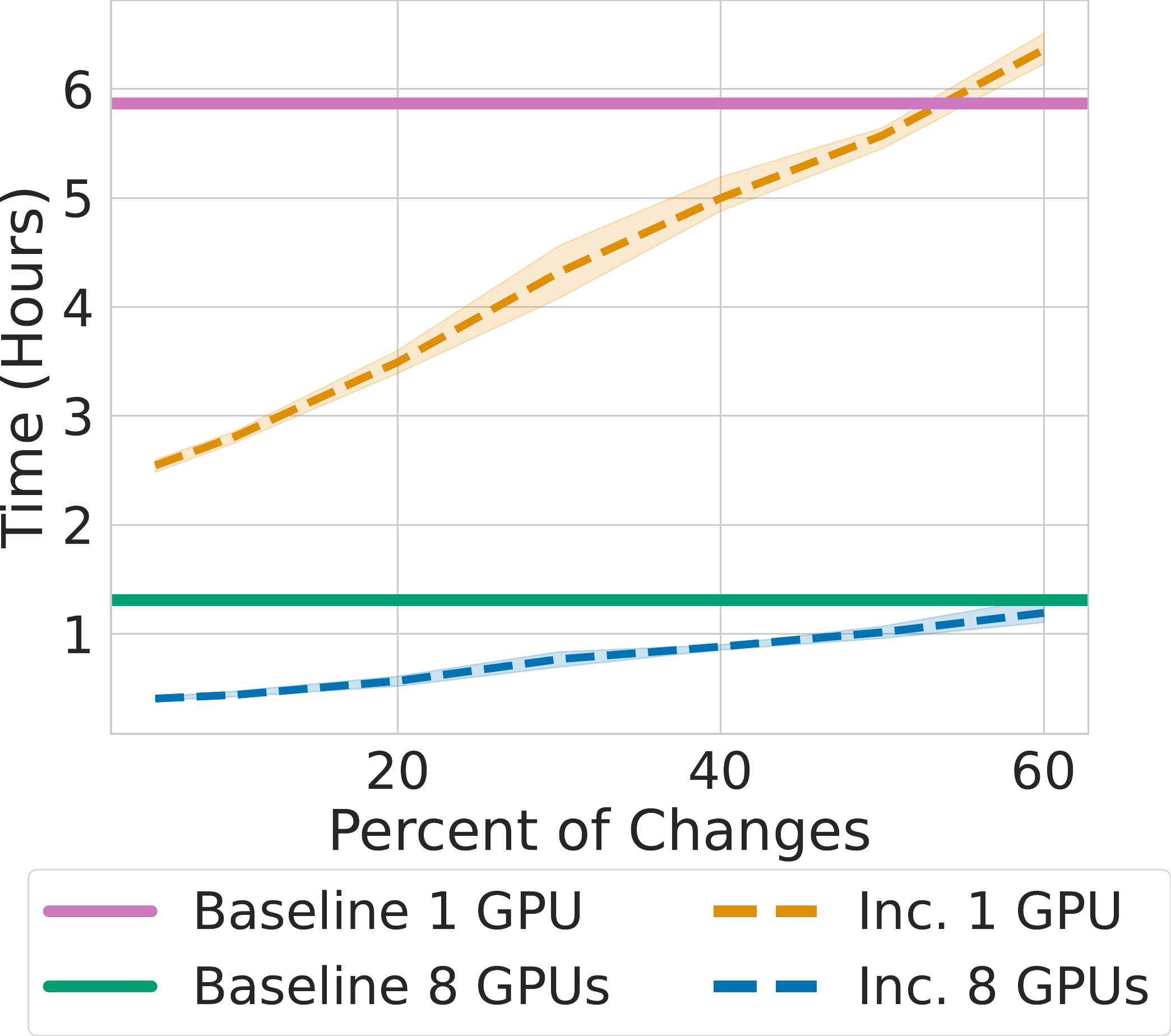}
        \vspace{1em}
        \caption{Execution time with respect to different number of feature changes for Q2 on CIFAR.}
    \label{fig:exp_change}
    \end{subfigure}
    \hfill
    \begin{subfigure}[c]{0.225\textwidth}
        \centering
        \includegraphics[width=\textwidth]{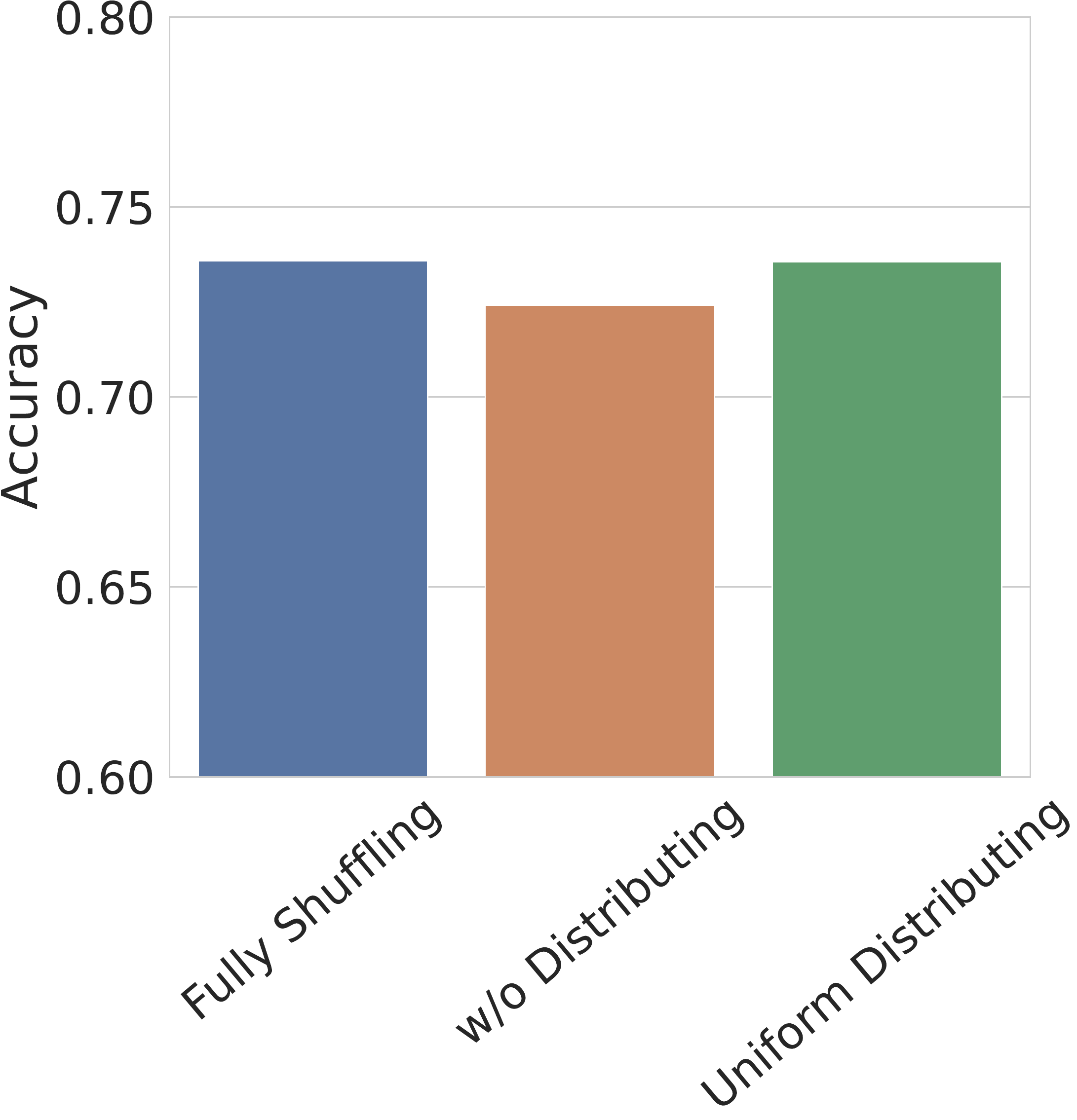}
        \caption{Distribution of new training data (a full class) for Q2 on Retinopathy.}
    \label{fig:incremental_uniform_distrib}
    \end{subfigure}
    \vspace{-1em}
    \caption{Incremental Execution.}
    \label{fig:inc_execution}
    \vspace{-1em}
\end{figure}

In Figure~\ref{fig:exp_change}, we randomly change a fixed percentage of the samples (i.e., manipulating the features) and plot the time required to perform an incremental execution of Q2 on CIFAR. Unsurprisingly, after a significant fraction of changes (e.g., $>50$\%), users might want to enforce a re-execution from scratch (i.e., by building a new initial reader instead of using a change-reader). However, when a small fraction of the samples are changed and the query is run incrementally, \sysname offers a significant speedup over the baseline. The computational performance of adding data follows the same trend as changing data. Note that the accuracy of any incremental query executions is heavily dataset- and distribution-dependent. The compute time for adding models to the query corresponds to the time needed to run a second independent query on these new models, due to the independence between the computation on the new models and the old ones. Changing any number of labels in \sysname is significantly cheaper compared to a re-execution. The reason lies in the large computational overhead of running inference compared to the cheap proxy computation. \sysname only requires to re-run the latter for all models. Finally, to further show the importance of distributing new samples, we construct a data reader for the Retinopathy dataset consisting of only four out of five classes. We then append the fifth class via an add-reader and compare the post fine-tune accuracy using different shuffling variants in Figure~\ref{fig:incremental_uniform_distrib}. This example illustrates a scenario in which the distributional difference between both readers, the initial one and the add-reader, leads to a good overall model being eliminated before seeing the new data if the samples are not distributed.

\subsection{Using The Benchmark Module}\label{sec:benchmark_results}

We empirically validate the usefulness of the benchmark module of \sysname be fine-tuning all VTAB-1K datasets~\citep{zhai2019visual} on a large set of more than 250 Huggingface transformer modules.\footnote{The fine-tune accuracies and model details are available under \url{https://github.com/DS3Lab/SHiFT}} Due to space limitations, we give the detailed time vs. post fine-tune accuracy comparison for all 19 datasets and a wide range of search queries, including randomly selecting one or two models, in Appendix~\ref{app:benchmark_module_results}. The results confirm the usefulness of our guidelines, as well as the results given by previous work~\citep{renggli2020model, kornblith2019better}.

\section{Other Related Work}

\paragraph{ML Specific Data Management} The data-management has been working on improving the usability of ML in a flurry of work over the last decade, by focusing on different components of the ML development process. A few examples include data acquisition with \textit{weak supervision} (e.g., Snorkel~\cite{ratner2017snorkel}, ZeroER~\cite{wu2020zeroer}), \textit{debugging and validation} (e.g., TFX~\cite{baylor2017tfx, polyzotis2019data}, ``Query 2.0''~\cite{wu2020complaint}, Krypton~\cite{nakandala2019incremental}), \textit{Model deployment} (e.g., MLFlow~\cite{zaharia2018accelerating}),
\textit{knowledge integration} (e.g., DeepDive~\cite{zhang2017deepdive}),
\textit{data cleaning}
(e.g., HoloClean~\cite{rekatsinas2017holoclean}, ActiveClean~\cite{krishnan2016activeclean}),
and \textit{interaction} (e.g., NorthStar~\cite{kraska2018northstar}). All these systems facilitate the ML development process, yet none of these focuses on transfer learning specifically.

\vspace{-0.5em}
\paragraph{Model Management} The data management community has also seen an intriguing line of work around  \textit{model management}. Systems like Cerebro~\cite{nakandala2020cerebro} or ModelDB~\cite{vartak2016modeldb}, and follow-up works~\citep{li2021towards, kumar2017data, li2021intermittent, schelter2018challenges}, are part of the main motivation to work on this new, transfer learning specific model management system. We hope that by lying the initial conceptual foundation of a model management system specifically for transfer learning, and by open sourcing \sysname, we are able to trigger and facilitate future research in this area.

\vspace{-0.5em}
\paragraph{Dataset Search}

Neural Data Server~\citep{neural-data-server, scalable-neural-data-server} takes the approach of searching for \textit{datasets} instead of pre-trained models to improve transfer learning. Other works such as Data2Vec~\citep{baevski2022data2vec} follow a similar goal by embedding a dataset and searching for similarity. Both approaches are orthogonal to our work, since we do not require access to the upstream datasets used to pre-train the models registered in \sysname. Furthermore, it is unclear how these techniques can be used to distinguish models trained on the \textit{same} upstream dataset.

\vspace{-0.5em}
\paragraph{Other Search Strategies}

There are other search strategies omitted in Section~\ref{sec:background} operating on semantical level (i.e., via a learned taxonomy)~\citep{zamir2018taskonomy}. These methods are not well suited for searching in a pre-trained model hub, mainly given the fact that they assume the input domain to remain fix, and datasets only to be different in their labels.

\section{Conclusion}

We presented \sysname, the first downstream task-aware search engine for transfer learning. Using our custom query language \queryname, users can generically define different model search strategies. Based on a cost-model, we automatically optimize prominent search queries and show significant speedups. Furthermore, by caching intermediate results, we allow our users to efficiently execute similar queries incrementally.
In the future, we hope that \sysname, together with our benchmark module, will enable researchers to easily implement and evaluate newer search strategies.

\bibliographystyle{ACM-Reference-Format}
\bibliography{references}

\clearpage
\newpage

\onecolumn

\appendix

\section{Overall Architecture of \sysname}\label{app:System}

\begin{figure*}[htp]
    \centering
    \includegraphics[width=\textwidth]{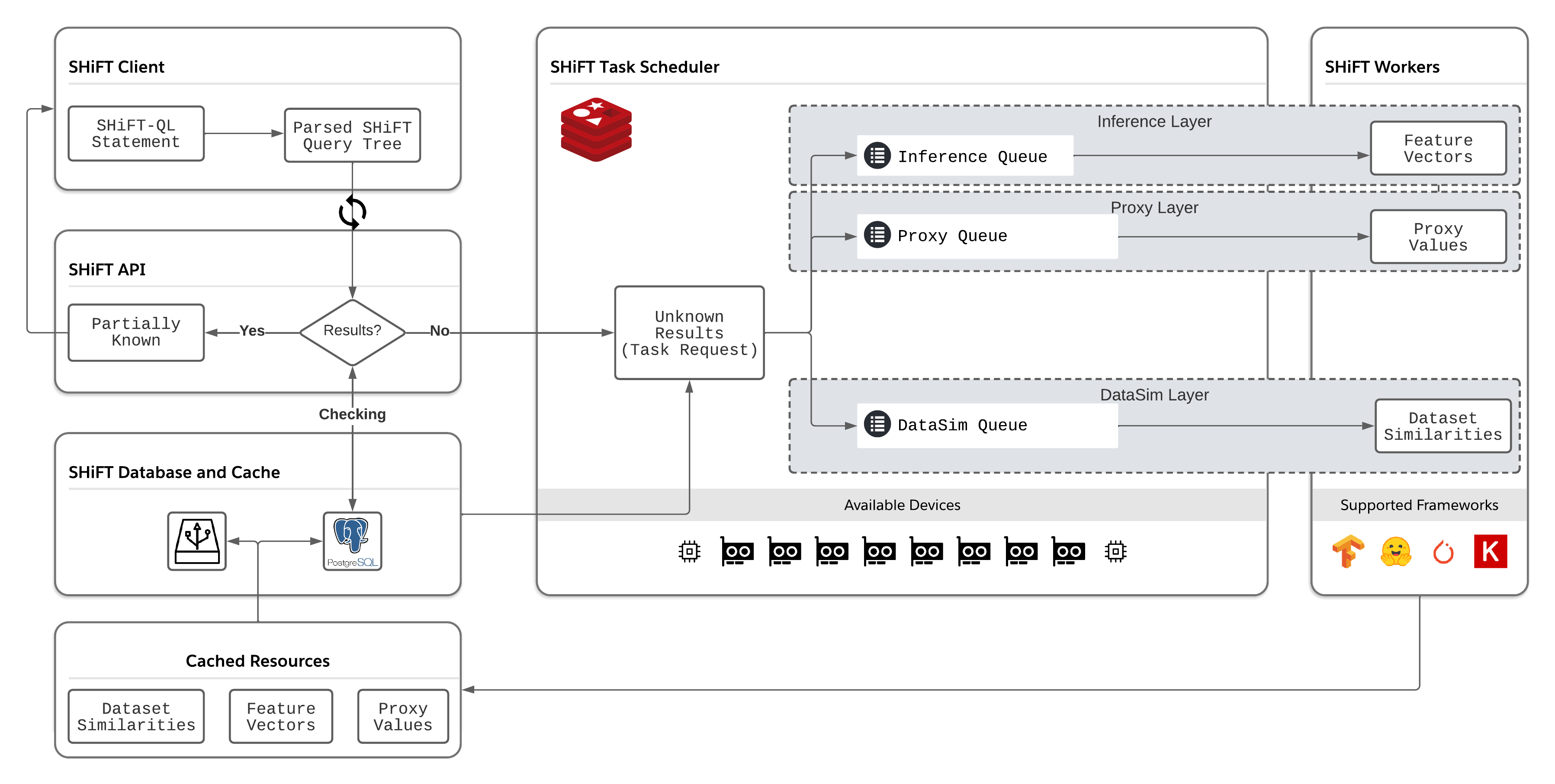}
    \caption{Overall system architecture of \sysname.}
    \label{fig:arch}
\end{figure*}

\FloatBarrier

\clearpage
\newpage

\section{Example \queryname Queries}\label{app:example_queries}

Given the flexible logical abstraction of \sysname,
we are able to express a diverse range of search strategies,
while allowing a user to conduct her own filtering and 
selections operations using standard SQL queries.
As an example, the five popular search 
strategies in Table~\ref{tbl:queries} can be expressed in \sysname as follows:

{

\begin{verbatim}
    Q1 := SELECT ModelId FROM Models
          WHERE Input = 'Vision'
          ORDER BY UpstreamAccuracy DESC LIMIT 1
\end{verbatim}

\begin{verbatim}
    Q2 := SELECT ModelId FROM Models
          WHERE Input = 'Vision'
          ORDER BY CosineNN ASC LIMIT 1
          TESTED ON TestReader TRAINED ON TrainReader
\end{verbatim}

\begin{verbatim}
    Q3 := SELECT ModelId FROM Models
          WHERE Input = 'Vision'
          ORDER BY Linear(lr=0.1) ASC LIMIT 1
          TESTED ON TestReader TRAINED ON TrainReader
\end{verbatim}

\begin{verbatim}
    Q4 := Q1
          UNION
          SELECT ModelId FROM Models
          WHERE Input = 'Vision' AND ModelId NOT IN Q1
          ORDER BY Linear(lr=0.1) ASC LIMIT 1
          TESTED ON TestReader TRAINED ON TrainReader
\end{verbatim}

\begin{verbatim}
    Q5 := SELECT ModelId FROM Models
          WHERE Input = 'Vision' AND
                DataReaders.ReaderId IN
                (
                    SELECT DataReaderId FROM DataReaders
                    ORDER BY Task2Vec LIMIT 1
                    TESTED ON TestReader
                ) Q6
          ORDER BY FineTune LIMIT 1
\end{verbatim}
}

\vspace{-0.5em}
\paragraph{More Complex Nested Query} 
Assuming users know they have a \textit{structured} computer vision dataset, it might be useful to prob only the best models for all structured datasets in the list of benchmark datasets. Noting that fine-tuning that many models can still be rather expensive, picking only the best task-aware model (e.g., via a linear proxy) out of those can further speedup the search process. This concrete example translates to the following \queryname query Q7, where Q8 represent the meta-learned task-agnostic part.

{
\begin{verbatim}
    Q7 := SELECT ModelId FROM 
          (
            SELECT ModelId FROM BenchmarkResults
            NATURAL JOIN DataReaders
            WHERE Model.Input = 'Vision' 
              AND DataReader.Type == 'Structured'
            RETRIEVE 2 GRP DataReaderId 
              ORD Accuracy DESC
          ) as Q8
          ORDER BY Linear(lr=0.1) ASC LIMIT 1
          TESTED ON TestReader TRAINED ON TrainReader
\end{verbatim}
}

\noindent
Notice that \sysname uses a syntactical sugar \verb!RETRIEVE!
to enable users of \queryname to retrieve the top 2 elements grouped by the attribute \verb!DataReaderId! ordered by the attribute \verb!Accuracy!. Internally, this keyword is automatically translated to a standard SQL query using the statements \verb!PARTITION BY! and \verb!RANK()!\footnote{The implementation is PostgreSQL specific. The function and statement names might be different for other distributions.}.

\clearpage
\newpage

\section{Successive-Halving Algorithm}\label{app:sh_algorithm}

\begin{algorithm}[h!]
    \caption{Successive-Halving~\citep{jamieson2016non}}
    \label{alg:SH}
    \begin{algorithmic}
        \STATE {\bfseries Input:} Budget $B$, chunk size $C$, $M$ candidate models  with $l_{i,k}$ denoting the loss of the $i$th model trained on the data samples in $[0, k]$\\
        \STATE {\bfseries Initialize:} $S_0 = [M]$
        \FOR{$k=0,1,\ldots, \lceil \log_2 (|M|)\rceil - 1$}
        \STATE Let $L=|S_k|$;
        \STATE Evaluate each model $i=1,\ldots, L$ with $r_k\times C$ more training data samples where $r_k=\lfloor \frac{B}{L\cdot \lceil \log_2 (|M|) \rceil} \rfloor$
        \STATE Set $R_k = C\times \sum_{j=0}^{k} r_j$
        \STATE Let $\sigma_k$ be a permutation on $S_k$ s.t. $l_{\sigma_k(1), R_k} \leq \ldots \leq l_{\sigma_k(|S_k|), R_k}$
        \STATE Let $S_{k+1} = \{ \sigma_k(1), \ldots, \sigma_k(\lceil L/2 \rceil)\}$.
        \IF{No data to perform more arm pulls}
        \STATE {\bfseries Output:} $\sigma_k(1)$
        \ENDIF
        \ENDFOR
        \STATE {\bfseries Output:} Singleton element of $S_{\lceil \log_2 (|M|) \rceil}$ \\
    \end{algorithmic}
\end{algorithm}

\clearpage
\newpage

\section{Model Details}
\label{sec:app:model_details}

\begin{center}
\small
    \begin{table}[!ht]
        \centering
        \caption{All vision models are available with ``https://tfhub.dev/'' as prefix (part 1/2).}
        \vspace{-1em}
        \begin{tabular}{lllll}
            \toprule
            Model & Inference Cost (ms) \\
            \midrule
            google/cropnet/feature\_vector/cassava\_disease\_V1/1 & 11  \\ \hline
        google/cropnet/feature\_vector/cassava\_disease\_V1/1 & 11  \\ \hline
        google/cropnet/feature\_vector/concat/1 & 12  \\ \hline
        google/cropnet/feature\_vector/imagenet/1 & 12  \\ \hline
        google/imagenet/efficientnet\_v2\_imagenet1k\_b0/feature\_vector/2 & 11  \\ \hline
        google/imagenet/efficientnet\_v2\_imagenet1k\_b1/feature\_vector/2 & 14  \\ \hline
        google/imagenet/efficientnet\_v2\_imagenet1k\_b2/feature\_vector/2 & 14  \\ \hline
        google/imagenet/efficientnet\_v2\_imagenet1k\_b3/feature\_vector/2 & 17  \\ \hline
        google/imagenet/efficientnet\_v2\_imagenet1k\_l/feature\_vector/2 & 85  \\ \hline
        google/imagenet/efficientnet\_v2\_imagenet1k\_m/feature\_vector/2 & 55  \\ \hline
        google/imagenet/efficientnet\_v2\_imagenet1k\_s/feature\_vector/2 & 23  \\ \hline
        google/imagenet/efficientnet\_v2\_imagenet21k\_b0/feature\_vector/2 & 11  \\ \hline
        google/imagenet/efficientnet\_v2\_imagenet21k\_b1/feature\_vector/2 & 11  \\ \hline
        google/imagenet/efficientnet\_v2\_imagenet21k\_b2/feature\_vector/2 & 12  \\ \hline
        google/imagenet/efficientnet\_v2\_imagenet21k\_b3/feature\_vector/2 & 14  \\ \hline
        google/imagenet/efficientnet\_v2\_imagenet21k\_ft1k\_b0/feature\_vector/2 & 11  \\ \hline
        google/imagenet/efficientnet\_v2\_imagenet21k\_ft1k\_b1/feature\_vector/2 & 12  \\ \hline
        google/imagenet/efficientnet\_v2\_imagenet21k\_ft1k\_b2/feature\_vector/2 & 13  \\ \hline
        google/imagenet/efficientnet\_v2\_imagenet21k\_ft1k\_b3/feature\_vector/2 & 14  \\ \hline
        google/imagenet/efficientnet\_v2\_imagenet21k\_ft1k\_l/feature\_vector/2 & 73  \\ \hline
        google/imagenet/efficientnet\_v2\_imagenet21k\_ft1k\_m/feature\_vector/2 & 33  \\ \hline
        google/imagenet/efficientnet\_v2\_imagenet21k\_ft1k\_xl/feature\_vector/2 & 74  \\ \hline
        google/imagenet/efficientnet\_v2\_imagenet21k\_l/feature\_vector/2 & 63  \\ \hline
        google/imagenet/efficientnet\_v2\_imagenet21k\_m/feature\_vector/2 & 33  \\ \hline
        google/imagenet/efficientnet\_v2\_imagenet21k\_s/feature\_vector/2 & 21  \\ \hline
        google/imagenet/inception\_resnet\_v2/feature\_vector/4 & 26  \\ \hline
        google/imagenet/inception\_v1/feature\_vector/4 & 13  \\ \hline
        google/imagenet/inception\_v2/feature\_vector/4 & 11  \\ \hline
        google/imagenet/inception\_v3/feature\_vector/4 & 14  \\ \hline
        google/imagenet/inception\_v3/feature\_vector/5 & 18  \\ \hline
        google/imagenet/mobilenet\_v1\_025\_128/feature\_vector/5 & 6  \\ \hline
        google/imagenet/mobilenet\_v1\_025\_160/feature\_vector/5 & 6  \\ \hline
        google/imagenet/mobilenet\_v1\_025\_192/feature\_vector/5 & 6  \\ \hline
        google/imagenet/mobilenet\_v1\_025\_224/feature\_vector/5 & 6  \\ \hline
        google/imagenet/mobilenet\_v1\_050\_128/feature\_vector/5 & 6  \\ \hline
        google/imagenet/mobilenet\_v1\_050\_160/feature\_vector/5 & 8  \\ \hline
        google/imagenet/mobilenet\_v1\_050\_192/feature\_vector/5 & 6  \\ \hline
        google/imagenet/mobilenet\_v1\_050\_224/feature\_vector/5 & 6  \\ \hline
        google/imagenet/mobilenet\_v1\_075\_128/feature\_vector/5 & 7  \\ \hline
        google/imagenet/mobilenet\_v1\_075\_160/feature\_vector/5 & 6  \\ \hline
        google/imagenet/mobilenet\_v1\_075\_192/feature\_vector/5 & 6  \\ \hline
        google/imagenet/mobilenet\_v1\_075\_224/feature\_vector/5 & 6  \\ \hline
        google/imagenet/mobilenet\_v1\_100\_128/feature\_vector/5 & 6  \\ \hline
        google/imagenet/mobilenet\_v1\_100\_160/feature\_vector/5 & 7  \\ \hline
        google/imagenet/mobilenet\_v1\_100\_192/feature\_vector/5 & 6  \\ \hline
        google/imagenet/mobilenet\_v1\_100\_224/feature\_vector/4 & 7  \\ \hline
        google/imagenet/mobilenet\_v2\_035\_128/feature\_vector/5 & 8  \\ \hline
        google/imagenet/mobilenet\_v2\_035\_160/feature\_vector/5 & 8  \\ \hline
        google/imagenet/mobilenet\_v2\_035\_192/feature\_vector/5 & 9  \\ \hline
        google/imagenet/mobilenet\_v2\_035\_224/feature\_vector/5 & 8  \\ \hline
        google/imagenet/mobilenet\_v2\_035\_96/feature\_vector/5 & 9 \\
        \bottomrule
        \end{tabular}
    \end{table}
\end{center}
\begin{center}
\small
    \begin{table}
        \centering
        \caption{All vision models are available with ``https://tfhub.dev/'' as prefix (part 2/2).}
        \vspace{-1em}
        \begin{tabular}{lllll}
            \toprule
            Model & Inference Cost (ms)  \\
            \midrule
        google/imagenet/mobilenet\_v2\_050\_128/feature\_vector/5 & 9  \\ \hline
        google/imagenet/mobilenet\_v2\_050\_160/feature\_vector/5 & 8  \\ \hline
        google/imagenet/mobilenet\_v2\_050\_192/feature\_vector/5 & 8 \\ \hline
        google/imagenet/mobilenet\_v2\_050\_224/feature\_vector/5 & 9  \\ \hline
        google/imagenet/mobilenet\_v2\_050\_96/feature\_vector/5 & 8  \\ \hline
        google/imagenet/mobilenet\_v2\_075\_128/feature\_vector/5 & 9  \\ \hline
        google/imagenet/mobilenet\_v2\_075\_160/feature\_vector/5 & 11  \\ \hline
        google/imagenet/mobilenet\_v2\_075\_192/feature\_vector/5 & 11  \\ \hline
        google/imagenet/mobilenet\_v2\_075\_224/feature\_vector/5 & 10  \\ \hline
        google/imagenet/mobilenet\_v2\_075\_96/feature\_vector/5 & 11  \\ \hline
        google/imagenet/mobilenet\_v2\_100\_128/feature\_vector/5 & 9  \\ \hline
        google/imagenet/mobilenet\_v2\_100\_160/feature\_vector/5 & 9  \\ \hline
        google/imagenet/mobilenet\_v2\_100\_192/feature\_vector/5 & 11  \\ \hline
        google/imagenet/mobilenet\_v2\_100\_224/feature\_vector/4 & 9  \\ \hline
        google/imagenet/mobilenet\_v2\_100\_96/feature\_vector/5 & 8 \\ \hline
        google/imagenet/mobilenet\_v2\_130\_224/feature\_vector/5 & 9  \\ \hline
        google/imagenet/mobilenet\_v2\_140\_224/feature\_vector/5 & 11  \\ \hline
        google/imagenet/mobilenet\_v3\_large\_075\_224/feature\_vector/5 & 11 \\ \hline
        google/imagenet/mobilenet\_v3\_large\_100\_224/feature\_vector/5 & 10  \\ \hline
        google/imagenet/mobilenet\_v3\_small\_075\_224/feature\_vector/5 & 9  \\ \hline
        google/imagenet/mobilenet\_v3\_small\_100\_224/feature\_vector/5 & 9  \\ \hline
        google/imagenet/nasnet\_mobile/feature\_vector/4 & 19  \\ \hline
        google/imagenet/resnet\_v1\_101/feature\_vector/4 & 16  \\ \hline
        google/imagenet/resnet\_v1\_152/feature\_vector/4 & 22  \\ \hline
        google/imagenet/resnet\_v1\_50/feature\_vector/4 & 13  \\ \hline
        google/imagenet/resnet\_v2\_101/feature\_vector/4 & 16  \\ \hline
        google/imagenet/resnet\_v2\_152/feature\_vector/4 & 22  \\ \hline
        google/imagenet/resnet\_v2\_50/feature\_vector/4 & 13  \\ \hline
        tensorflow/efficientnet/b0/feature-vector/1 & 12  \\ \hline
        tensorflow/efficientnet/b1/feature-vector/1 & 15  \\ \hline
        tensorflow/efficientnet/b2/feature-vector/1 & 17  \\ \hline
        tensorflow/efficientnet/b3/feature-vector/1 & 21  \\ \hline
        tensorflow/efficientnet/b4/feature-vector/1 & 32  \\ \hline
        tensorflow/efficientnet/b5/feature-vector/1 & 44  \\ \hline
        tensorflow/efficientnet/b6/feature-vector/1 & 79  \\ \hline
        tensorflow/efficientnet/b7/feature-vector/1 & 151  \\ \hline
        vtab/exemplar/1 & 13  \\ \hline
        vtab/jigsaw/1 & 25  \\ \hline
        vtab/relative-patch-location/1 & 21  \\ \hline
        vtab/rotation/1 & 12  \\ \hline
        vtab/semi-exemplar-10/1 & 12  \\ \hline
        vtab/semi-rotation-10/1 & 12  \\ \hline
        vtab/sup-100/1 & 12  \\ \hline
        vtab/sup-exemplar-100/1 & 11  \\ \hline
        vtab/sup-rotation-100/1 & 12  \\ \hline
        vtab/uncond-biggan/1 & 17  \\ \hline
        vtab/vae/1 & 10  \\ \hline
        vtab/wae-gan/1 & 10  \\ \hline
        vtab/wae-mmd/1 & 10  \\ \hline
        vtab/wae-ukl/1 & 10  \\
        \bottomrule
        \end{tabular}
    \end{table}
\end{center}
\begin{center}
    \small
        \begin{table}
            \centering
            \caption{All NLP models are available as HuggingFace transformers (part 1/2).}
            \vspace{-1em}
            \begin{tabular}{lllll}
                \toprule
                Model & Inference Cost (ms)  \\
                \midrule
                18811449050/bert\_finetuning\_test & 18 \\ \hline
        aditeyabaral/finetuned-sail2017-xlm-roberta-base & 17 \\ \hline
        aliosm/sha3bor-metre-detector-arabertv2-base & 21 \\ \hline
        Alireza1044/albert-base-v2-qnli & 22 \\ \hline
        anferico/bert-for-patents & 59 \\ \hline
        anirudh21/bert-base-uncased-finetuned-qnli & 18 \\ \hline
        ASCCCCCCCC/distilbert-base-chinese-amazon\_zh\_20000 & 21 \\ \hline
        aviator-neural/bert-base-uncased-sst2 & 19 \\ \hline
        aychang/bert-base-cased-trec-coarse & 21 \\ \hline
        bert-base-cased & 19 \\ \hline
        bert-base-uncased & 19 \\ \hline
        bert-large-uncased& 55 \\ \hline
        bondi/bert-semaphore-prediction-w4 & 21 \\ \hline
        CAMeL-Lab/bert-base-arabic-camelbert-da-sentiment & 21 \\ \hline
        CAMeL-Lab/bert-base-arabic-camelbert-mix-did-nadi & 21 \\ \hline
        Capreolus/bert-base-msmarco & 17 \\ \hline
        chiragasarpota/scotus-bert & 6 \\ \hline
        classla/bcms-bertic-parlasent-bcs-ter & 21 \\ \hline
        connectivity/bert\_ft\_qqp-1 & 19 \\ \hline
        connectivity/bert\_ft\_qqp-17 & 19 \\ \hline
        connectivity/bert\_ft\_qqp-25 & 19 \\ \hline
        connectivity/bert\_ft\_qqp-7 & 21 \\ \hline
        connectivity/bert\_ft\_qqp-94  & 21\\ \hline
        connectivity/bert\_ft\_qqp-96 & 19 \\ \hline
        connectivity/feather\_berts\_28 & 17 \\ \hline
        dhimskyy/wiki-bert & 13 \\ \hline
        DoyyingFace/bert-asian-hate-tweets-asian-unclean-freeze-4 & 19 \\ \hline
        emrecan/bert-base-multilingual-cased-snli\_tr & 21 \\ \hline
        gchhablani/bert-base-cased-finetuned-rte & 19 \\ \hline
        gchhablani/bert-base-cased-finetuned-wnli & 18 \\ \hline
        Guscode/DKbert-hatespeech-detection & 20 \\ \hline
        ishan/bert-base-uncased-mnli & 17 \\ \hline
        jb2k/bert-base-multilingual-cased-language-detection & 22 \\ \hline
        Jeevesh8/512seq\_len\_6ep\_bert\_ft\_cola-91 & 17 \\ \hline
        Jeevesh8/6ep\_bert\_ft\_cola-12 & 17 \\ \hline
        Jeevesh8/6ep\_bert\_ft\_cola-29 & 18 \\ \hline
        Jeevesh8/6ep\_bert\_ft\_cola-47 & 19 \\
            \bottomrule
            \end{tabular}
        \end{table}
    \end{center}

\begin{center}
    \small
        \begin{table}
            \centering
            \caption{All NLP models are available as HuggingFace transformers (part 2/2).}
            \vspace{-1em}
            \begin{tabular}{lllll}
                \toprule
                Model & Inference Cost (ms)  \\
                \midrule
                Jeevesh8/bert\_ft\_cola-60 & 19\\ \hline
        Jeevesh8/bert\_ft\_cola-88 & 19 \\ \hline
        Jeevesh8/bert\_ft\_qqp-39 & 20 \\ \hline
        Jeevesh8/bert\_ft\_qqp-40 & 19 \\ \hline
        Jeevesh8/bert\_ft\_qqp-55 & 20 \\ \hline
        Jeevesh8/bert\_ft\_qqp-68 & 21 \\ \hline
        Jeevesh8/bert\_ft\_qqp-88 & 19\\ \hline
        Jeevesh8/bert\_ft\_qqp-9  & 19\\ \hline
        Jeevesh8/feather\_berts\_46 & 17 \\ \hline
        Jeevesh8/feather\_berts\_96 & 17\\ \hline
        Jeevesh8/init\_bert\_ft\_qqp-24 & 21\\ \hline
        Jeevesh8/init\_bert\_ft\_qqp-28 & 21\\ \hline
        Jeevesh8/init\_bert\_ft\_qqp-33  &19 \\ \hline
        Jeevesh8/init\_bert\_ft\_qqp-49 & 21\\ \hline
        Jeevesh8/lecun\_feather\_berts-3 & 17\\ \hline
        Jeevesh8/lecun\_feather\_berts-51 & 17 \\ \hline
        manueltonneau/bert-twitter-en-is-hired & 19\\ \hline
        Monsia/camembert-fr-covid-tweet-classification & 21 \\ \hline
        moshew/bert-mini-sst2-distilled & 2\\ \hline
        mujeensung/bert-base-cased\_mnli\_bc & 17 \\ \hline
        navsad/navid\_test\_bert & 17 \\ \hline
        oferweintraub/bert-base-finance-sentiment-noisy-search & 18 \\ \hline
        Recognai/bert-base-spanish-wwm-cased-xnli & 21 \\ \hline
        socialmediaie/TRAC2020\_IBEN\_B\_bert-base-multilingual-uncased & 21 \\ \hline
        Splend1dchan/bert-base-uncased-slue-goldtrascription-e3-lr1e-4 & 17 \\ \hline
        w11wo/sundanese-bert-base-emotion-classifier & 21 \\ \hline
        waboucay/camembert-base-finetuned-xnli\_fr-finetuned-nli-rua\_wl & 21 \\ \hline
        XSY/albert-base-v2-imdb-calssification & 19 \\

            \bottomrule
            \end{tabular}
        \end{table}
    \end{center}

\clearpage
\newpage

\section{NLP Results}
\label{sec:app:nlp_results}

\begin{figure}[hb!]
    \centering
    \includegraphics[width=0.48\textwidth]{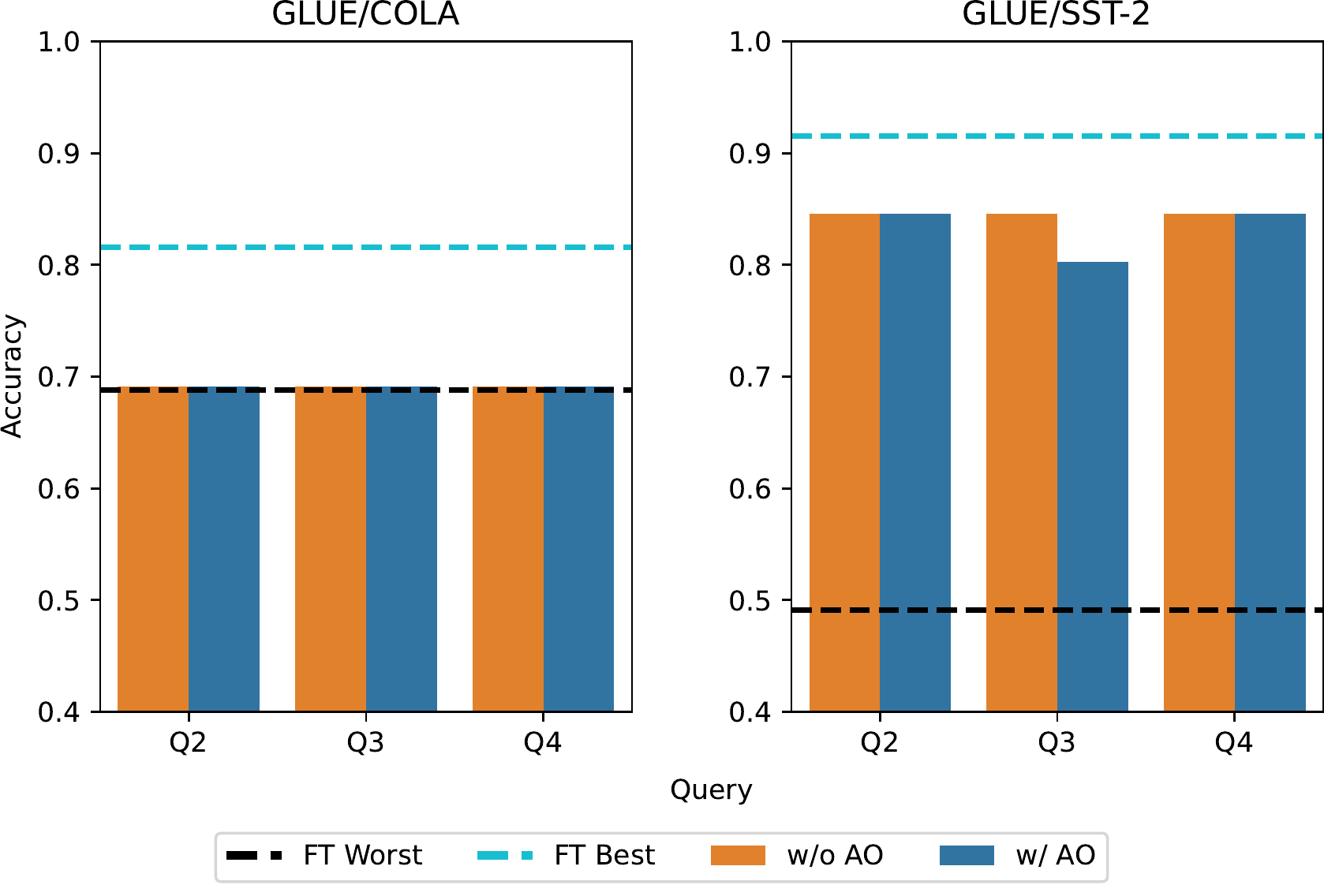}
    \caption{Fine-tune (FT) accuracy of returned NLP model for all settings.}
    \label{fig:finetune_vs_shift_accuracy_nlp}
\end{figure}

We conduct the same end-to-end experiments from the main paper on the two NLP datasets and the models listed in Appendix~\ref{sec:app:model_details}. Table~\ref{tbl:finetune_vs_shift_runtime_nlp} shows the relative speedup of running Q2-Q4 with and without automatic optimization compared to enumerate all models by fine-tuning them. We realize that in all cases, \sysname with automatic optimization significantly outperforms the FT enumeration baseline.
When comparing the post fine-tune accuracies in Figure~\ref{fig:finetune_vs_shift_accuracy_nlp}, we see that there is little variance between the different methods. The linear proxy using SH (i.e., Q3 with automatic optimization) for SST2 is slightly inferior to the other queries. The reason is assumed to lie in the variance induced from the very small test dataset compared to the large training set for SST2. The same fact also yields a speedup of almost 10x when using SH via automatic optimization compared to not using SH. For COLA, we see that all search queries are picking a model on par with the worst model. When inspecting the distribution of the fine-tune accuracies, we see that all of them, except one, result in 0.7. The search queries fail to select this specific, better model.

\begin{table}[hb!]
 \vspace{1em}
 \caption{Execution time for fine-tuning (FT) all the models via enumeration compared to running Q2-Q4 using \sysname with and without automatic optimization (AO).}
 \vspace{1em}
 \label{tbl:finetune_vs_shift_runtime_nlp}
 \small
 \begin{tabular}{rrrllll}
  \hline
&& & \multicolumn{2}{c}{1 GPU}  & \multicolumn{2}{c}{8 GPU}  \\
&& & \multicolumn{1}{r}{\begin{tabular}[c]{@{}r@{}}Runtime\\ (Hours)\end{tabular}} & \multicolumn{1}{r}{\begin{tabular}[c]{@{}r@{}}Speedup\\ (vs. FT)\end{tabular}} & \multicolumn{1}{r}{\begin{tabular}[c]{@{}r@{}}Runtime\\ (Hours)\end{tabular}} & \multicolumn{1}{r}{\begin{tabular}[c]{@{}r@{}}Speedup\\ (vs. FT)\end{tabular}} \\ \hline
  \multirow{7}{*}{GLUE/COLA}& FT&  & 151.2 &  & 18.9 &  \\
& \multirow{2}{*}{Q2} & w/o AO & 3.6& 41.8x& 0.5 & 37.3x\\
&& w/ AO & 2.7& \textbf{56.0x } & 0.3 & \textbf{67.4x} \\
& \multirow{2}{*}{Q3} & w/o AO & 4.2& 36.4x& 0.6& 32.0x\\
&& w/ AO &1.7& \textbf{88.3x } & 0.3& \textbf{64.1x  }\\
& \multirow{2}{*}{Q4} & w/o AO & 3.6& 42.5x& 0.5& 39.5x  \\
&& w/ AO & 1.4& \textbf{106.4x  }& 0.2& \textbf{75.8x}\\ \hline
  \multirow{7}{*}{GLUE/SST-2}& FT&  & 1045.9 &  & 130.7 &  \\
& \multirow{2}{*}{Q2} & w/o AO & 22.3 & 47.0x& 3.0& 43.0x\\
&& w/ AO & 3.0& \textbf{348.2x  }&0.7 & \textbf{181.6x  }\\

& \multirow{2}{*}{Q3} & w/o AO & 22.5& 46.5x& 2.9& 44.5x\\
&& w/ AO & 2.7& \textbf{389.0x } & 0.6& \textbf{209.0x  }\\
& \multirow{2}{*}{Q4} & w/o AO & 21.7& 48.2x& 2.9& 45.7x \\
&& w/ AO & 2.3& \textbf{452.0x}& 0.5& \textbf{259.6x}\\ \hline
 \end{tabular}
\end{table}

\clearpage
\newpage

\section{Benchmark Module Results}\label{app:benchmark_module_results}

We next show how the benchmark module can be used to easily position a new search strategy against existing ones.

\subsection{Protocol}

\paragraph{Datasets} We use the 19 VTAB-1K~\citep{zhai2019visual} datasets. The datasets are chosen such that they cover a large range of possible classification tasks in the visual domain. Furthermore, the search space for fine-tuning any model on these datasets is well understood.

\paragraph{Fine-tune protocol and computation time} We follow the fine-tune protocol outline in the main paper following \citet{zhai2019visual}. The computation time consists of two parts: (a) the time to run the search query, if any, and (b) fine-tuning all resulting models. The max fine-tune accuracy of these models is then plotted against the compute time. Note that for meta-learned approaches, the time to compute the cross product of fine-tune accuracies between benchmark datasets and all models is not included into the computation time.

\paragraph{Models} We select a large set of 250 publicly available HuggingFace Transformers models. The list and all fine-tune accuracies are available in our GitHub repository (\url{https://github.com/DS3Lab/shift}).

\paragraph{System state simulation} Having access to all $19 \times 250$ fine-tune accuracies, we remove a single dataset including the corresponding fine-tune results from the list of benchmark datasets and fine-tune accuracies. We then use this dataset as a target dataset and the other 18 as benchmark datasets for meta-learned queries. The fine-tune results of the models returned by a search strategy are known and can be used to plot the post fine-tune accuracy of a search strategy. We execute the search strategies on a single GPU.

\subsection{Strategies}

We compare multiple strategies from the paper (Q4, Q5 and Q7) as well as new ones. For Q7, we replace the filter ``structured'' with the corresponding target dataset type as described by \citet{zhai2019visual} (e.g., Natural, Specialized, and Structured).

\paragraph{Random Sampling} A non-deterministic search strategy might consist of random sampling (with a uniform distribution) one or multiple models (without replacement) out of the list of available models. Clearly this method will suffer from a large variance despite being free of search costs. We sample uniformly for 50 times and show the maximum, minimum and mean in the plots.

\paragraph{LEEP} There are many other purely task-aware search strategies, similar to Q2 and Q3. We implement LEEP by \citep{nguyen2020leep} and benchmark it against other methods next.

\subsection{Evaluation}

\begin{figure}[t!]
    \centering
    \begin{subfigure}[ht]{0.33\textwidth}
        \centering
        \includegraphics[width=\textwidth]{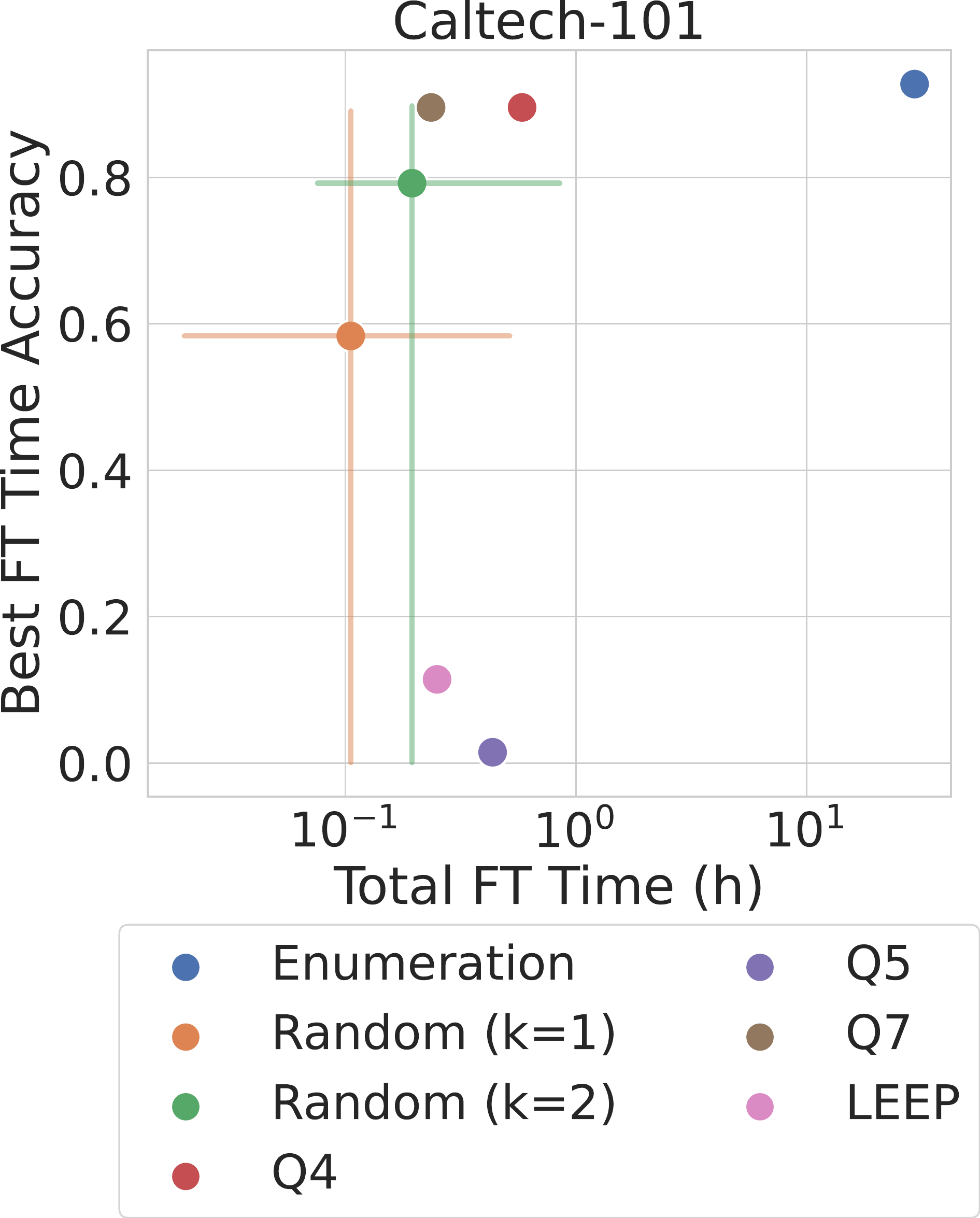}
    \end{subfigure}
    \hfill
    \begin{subfigure}[ht]{0.33\textwidth}
        \centering
        \includegraphics[width=\textwidth]{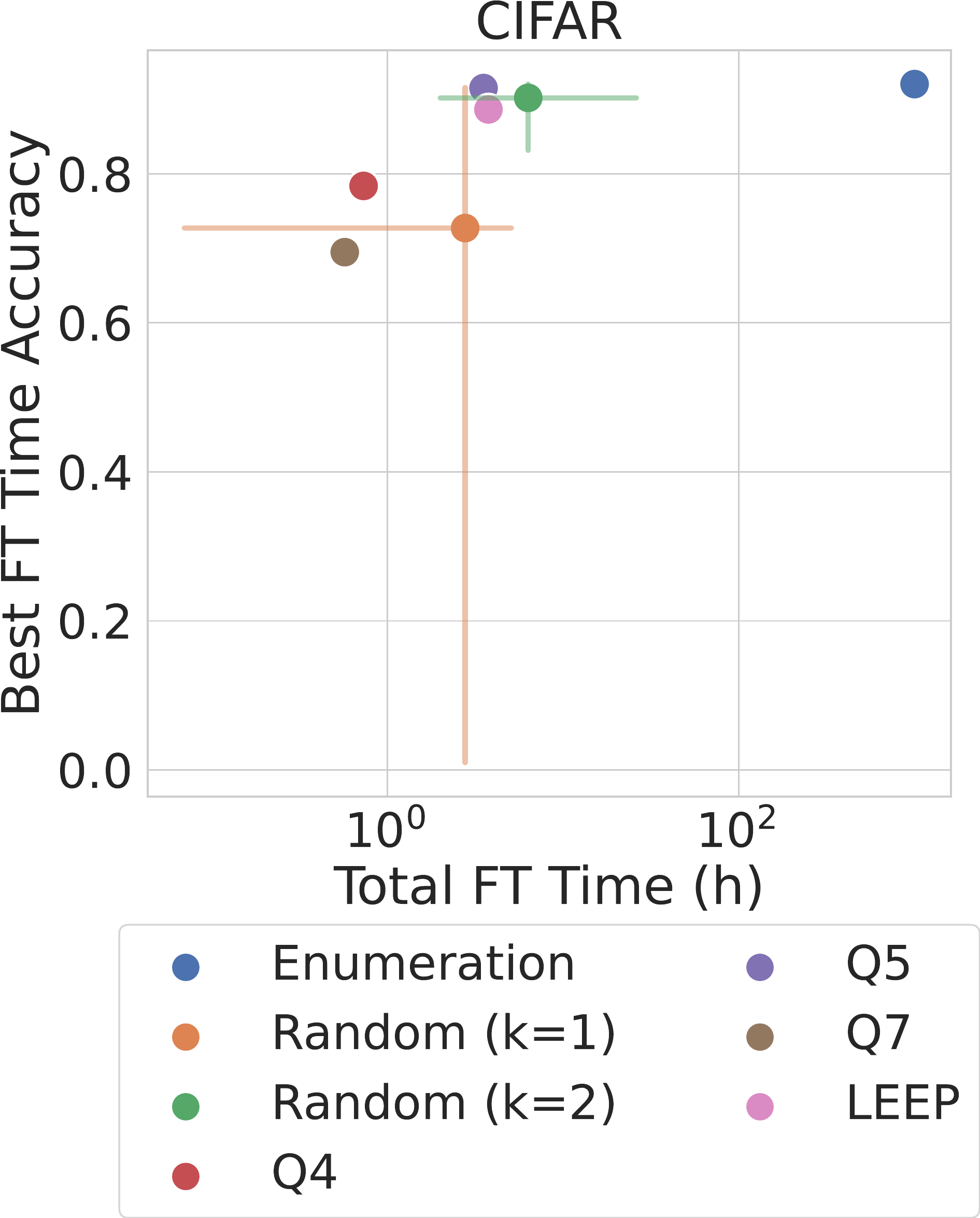}
    \end{subfigure}
    \hfill
    \begin{subfigure}[ht]{0.33\textwidth}
        \centering
        \includegraphics[width=\textwidth]{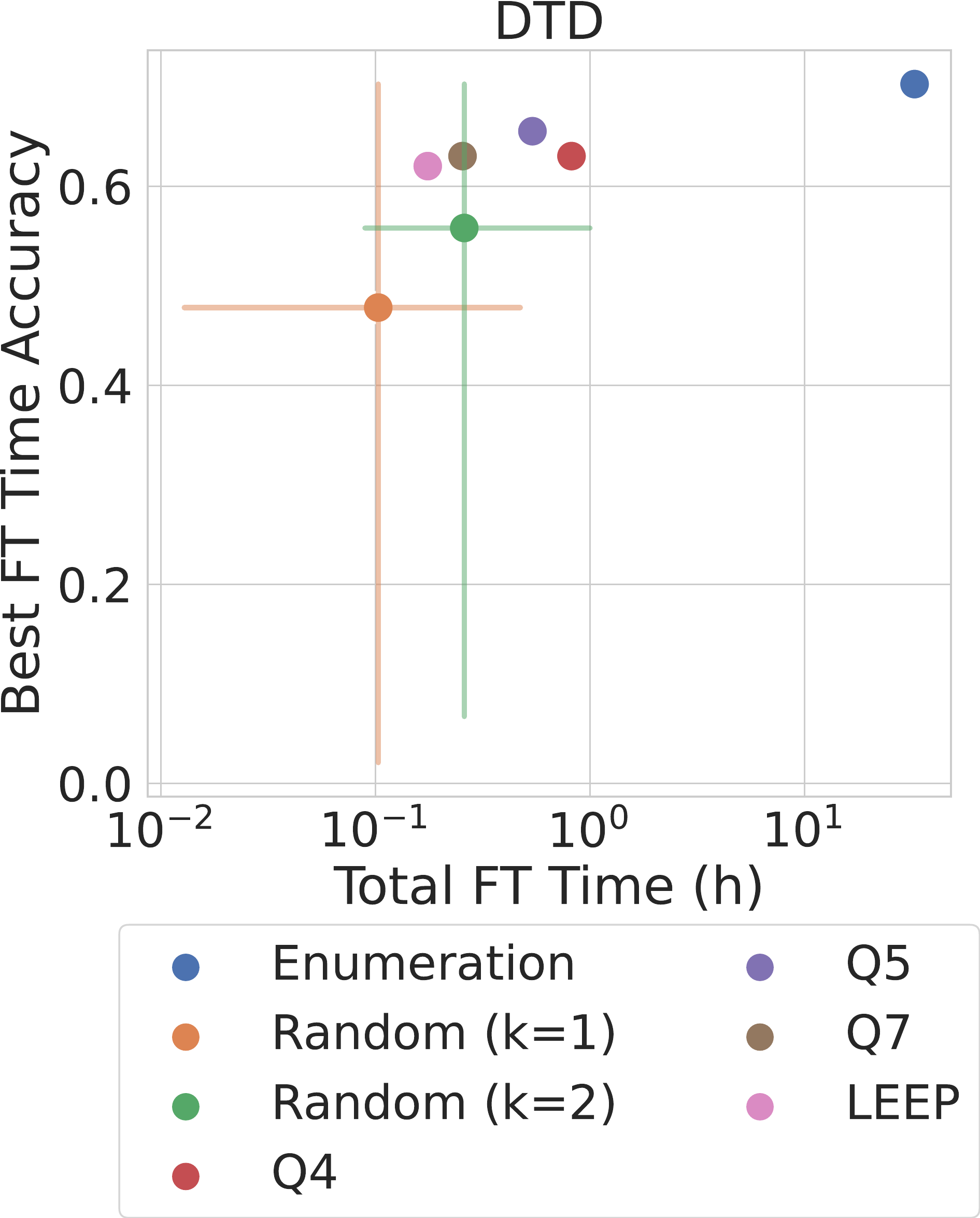}
    \end{subfigure}
    \caption{Benchmark module results 1/6.}
    \label{fig:benchmark_resutls_1}
\end{figure}

\begin{figure}[t!]
    \centering
    \begin{subfigure}[ht]{0.33\textwidth}
        \centering
        \includegraphics[width=\textwidth]{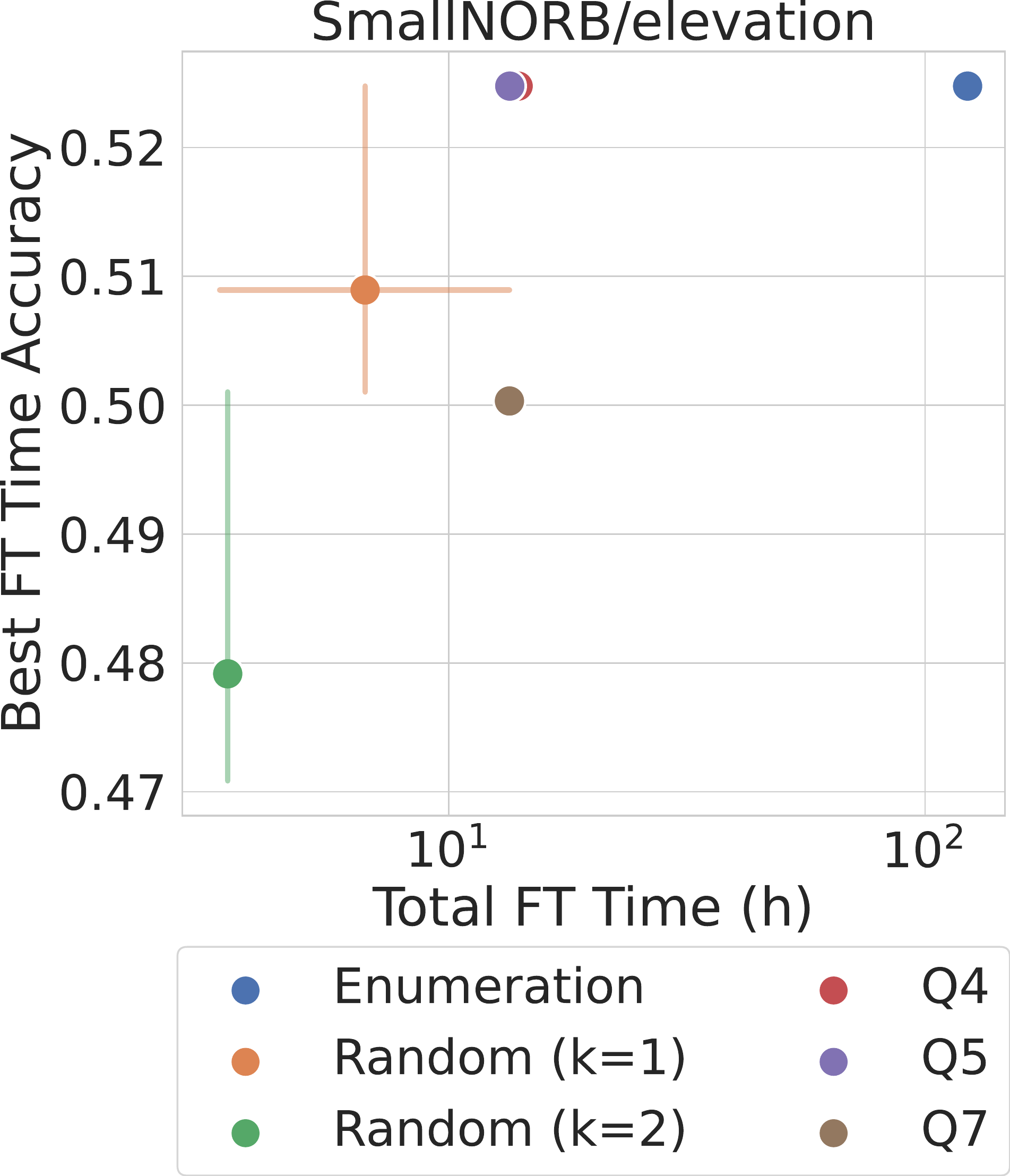}
    \end{subfigure}
    \hfill
    \begin{subfigure}[ht]{0.33\textwidth}
        \centering
        \includegraphics[width=\textwidth]{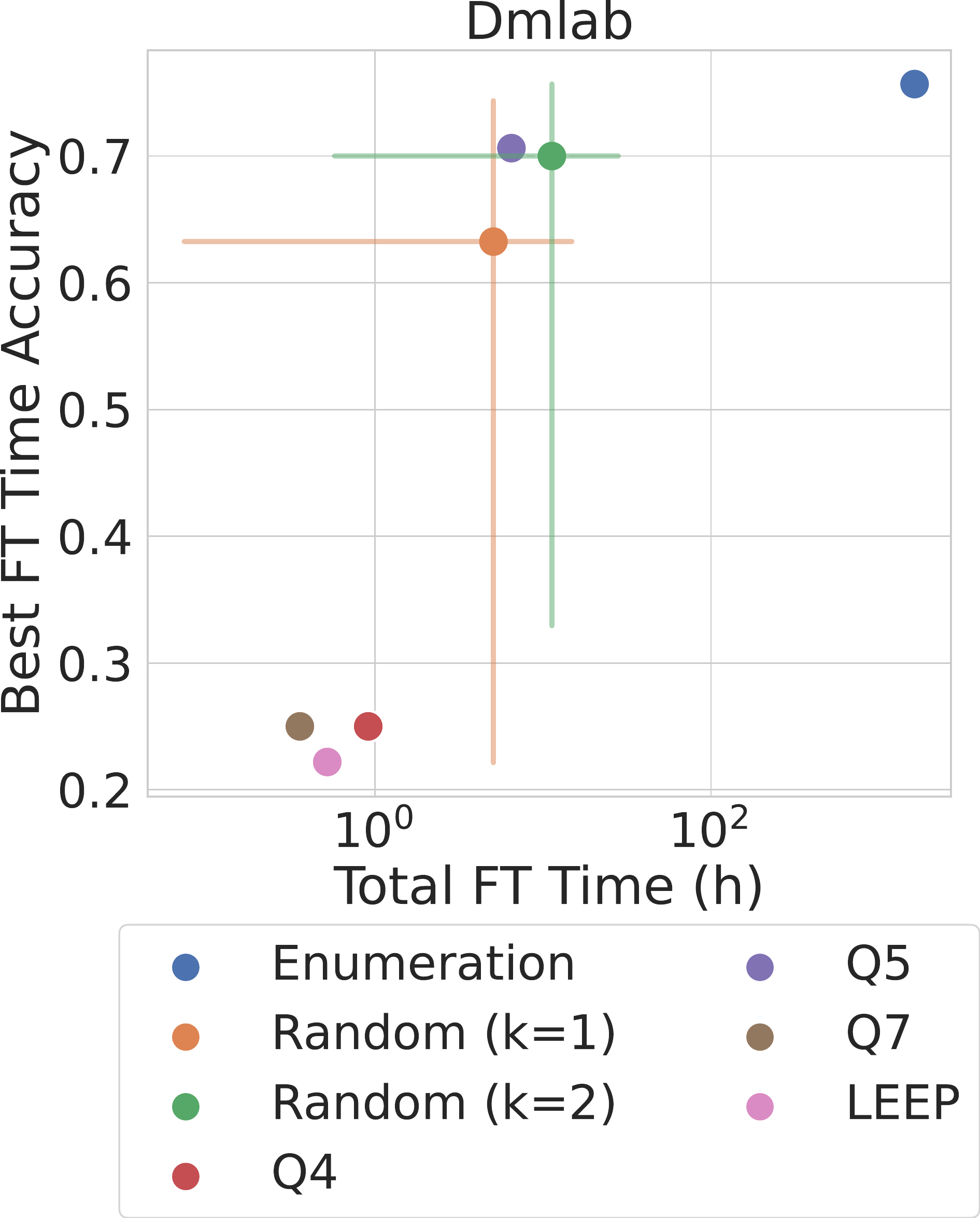}
    \end{subfigure}
    \hfill
    \begin{subfigure}[ht]{0.33\textwidth}
        \centering
        \includegraphics[width=\textwidth]{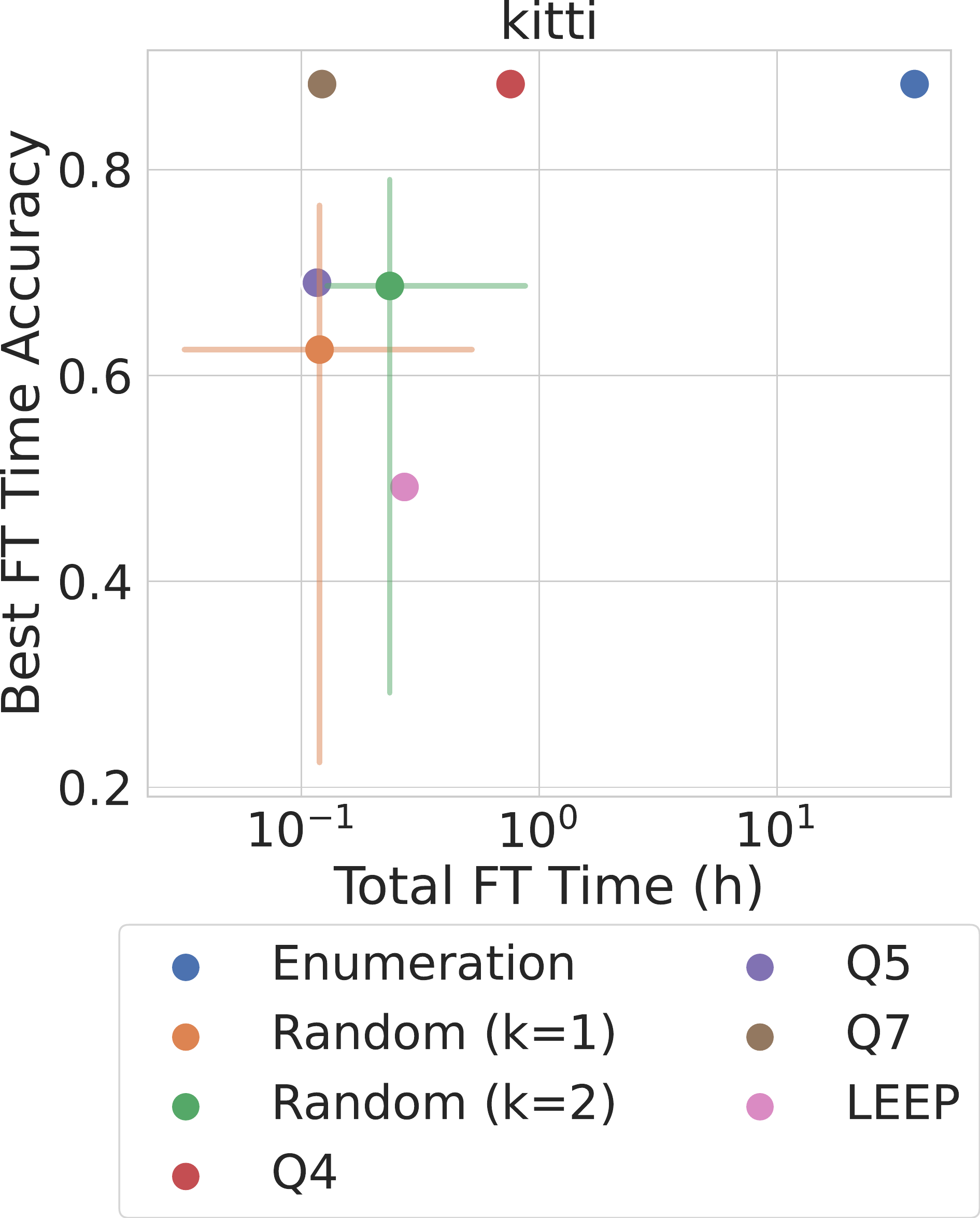}
    \end{subfigure}
    \caption{Benchmark module results 6/6.}
    \label{fig:benchmark_resutls_6}
\end{figure}

We provide the benchmark module results for all 19 VTAB-1K datasets in Figures~\ref{fig:benchmark_resutls_1} - \ref{fig:benchmark_resutls_6}. All search strategies except then random sampling one are deterministic. The enumeration baseline represents the best reachable accuracy. Ideally, we would want a search strategy which is cheaper than this baseline (i.e., on its left) and does not suffer from large regret (i.e., at the same height).

\paragraph{Random Sampling}
The mean performance is not representative for such a sampling-based search strategy. The performance of a random baseline is rather implicitly linked to the variance and concentration around the mean of post fine-tune accuracies for a fixed dataset and model pool. In an extreme case, where all models perform similarly or only has outliers performing worse than the majority (i.e., mean near the maximum), the random baseline will perform well (e.g., for CIFAR). On the other hand, if there are outliers performing much better than the mean, the probability of selecting this model is low, and users will likely end up with a sub-optimal model (e.g., for SVHN).

\paragraph{Q4 vs Q7}

When comparing Q4 and Q7 we see through most graphs, that the hybrid strategy mostly outperforms the meta-learned complex one in terms of accuracy. The latter is faster though for two reasons: (a) based on the meta-learned part (i.e., Q8), the search strategy only has to run the proxy computation over a small set of at most 18 models, as opposed to 250 for Q4. Then, running Q8 will only return a single model to fine-tune, whereas Q4 suggests two models, both of them having to be fine-tuned.

\paragraph{LEEP} When comparing LEEP against Q4, we see that the query is, as expected, often slightly cheaper compared to Q4. It is nonetheless often inferior in terms of fine-tune accuracy, and sometimes even significantly less (e.g., for Flowers).

\clearpage
\newpage

\begin{figure}[t!]
    \centering
    \begin{subfigure}[ht]{0.33\textwidth}
        \centering
        \includegraphics[width=\textwidth]{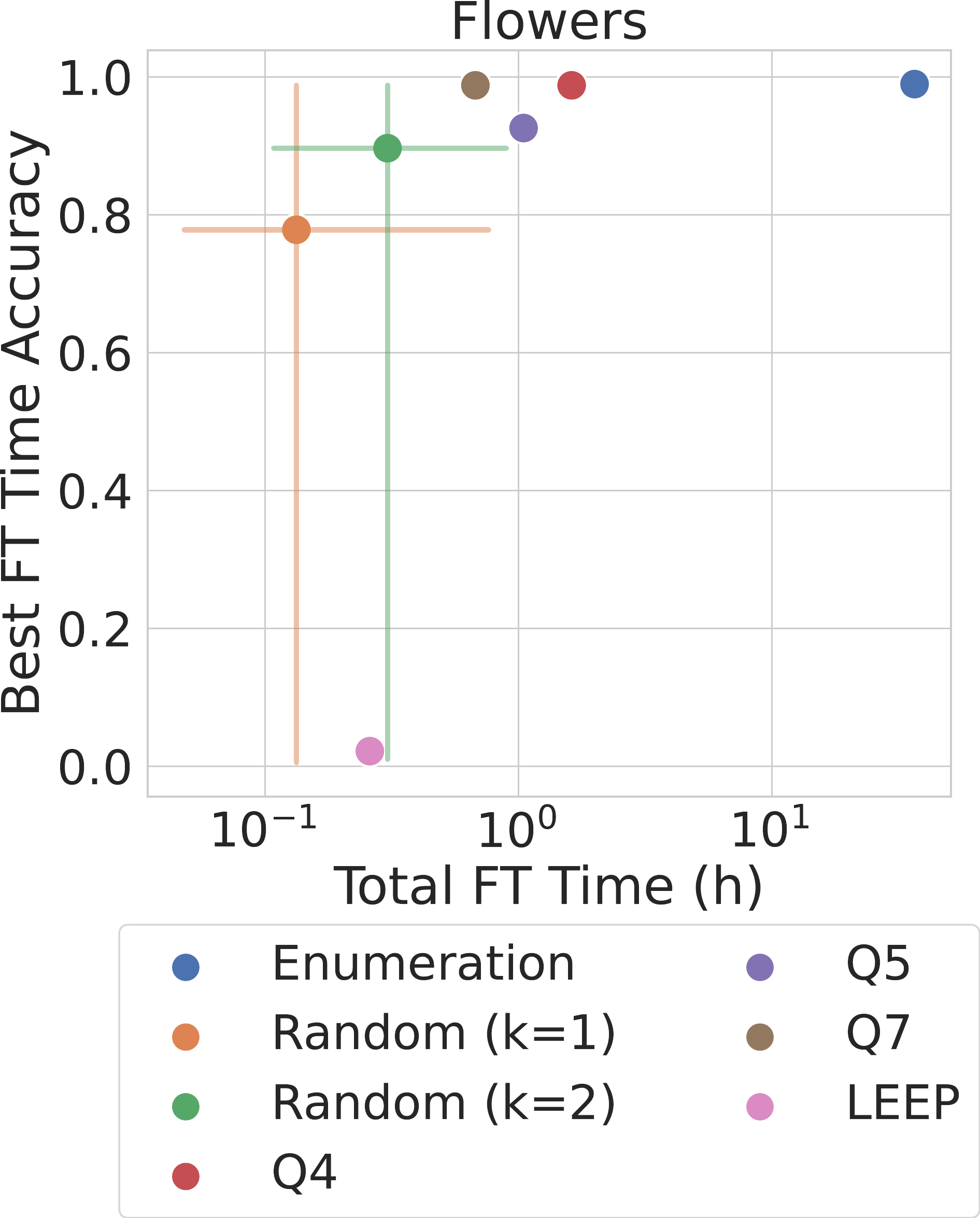}
    \end{subfigure}
    \hfill
    \begin{subfigure}[ht]{0.33\textwidth}
        \centering
        \includegraphics[width=\textwidth]{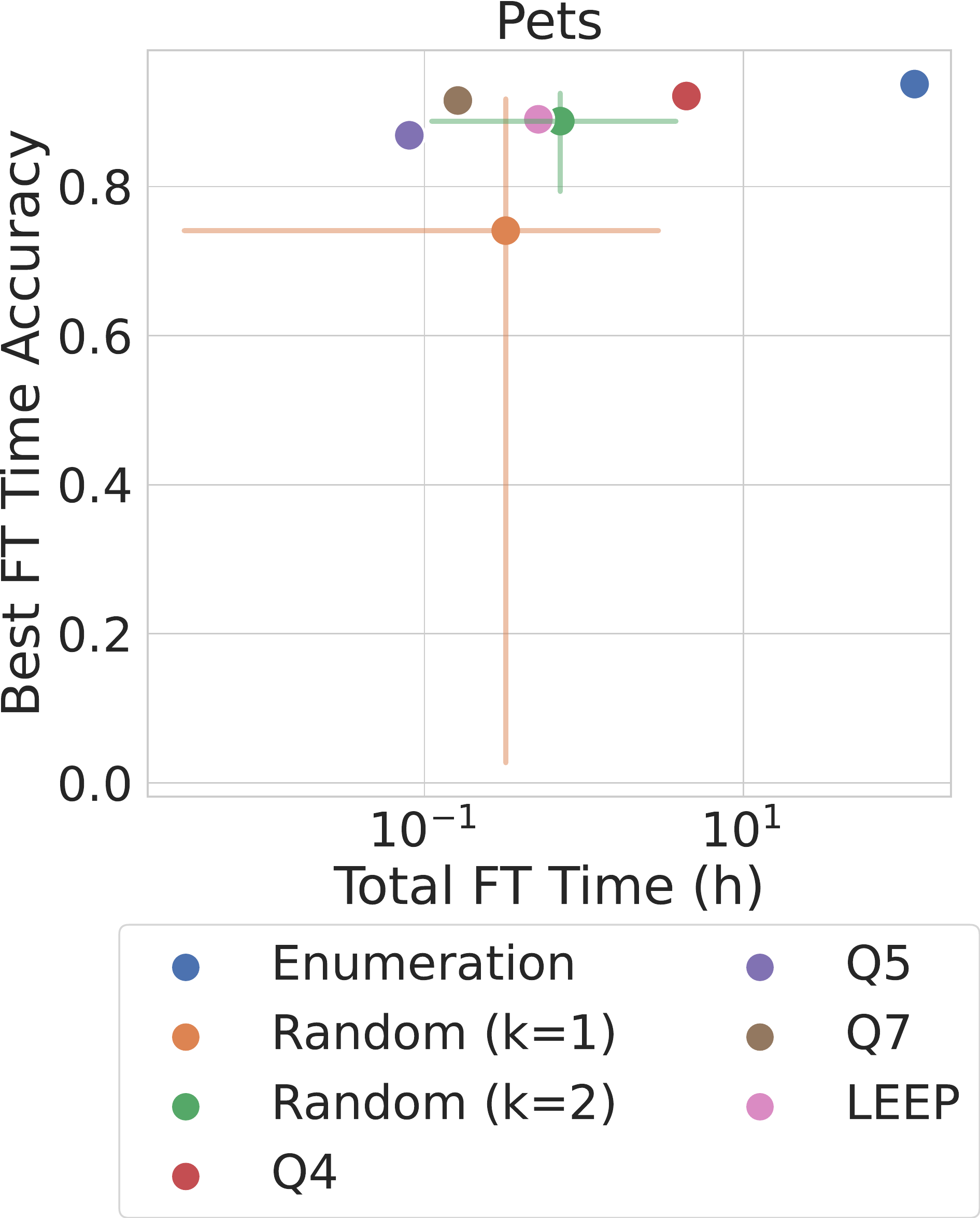}
    \end{subfigure}
    \hfill
    \begin{subfigure}[ht]{0.33\textwidth}
        \centering
        \includegraphics[width=\textwidth]{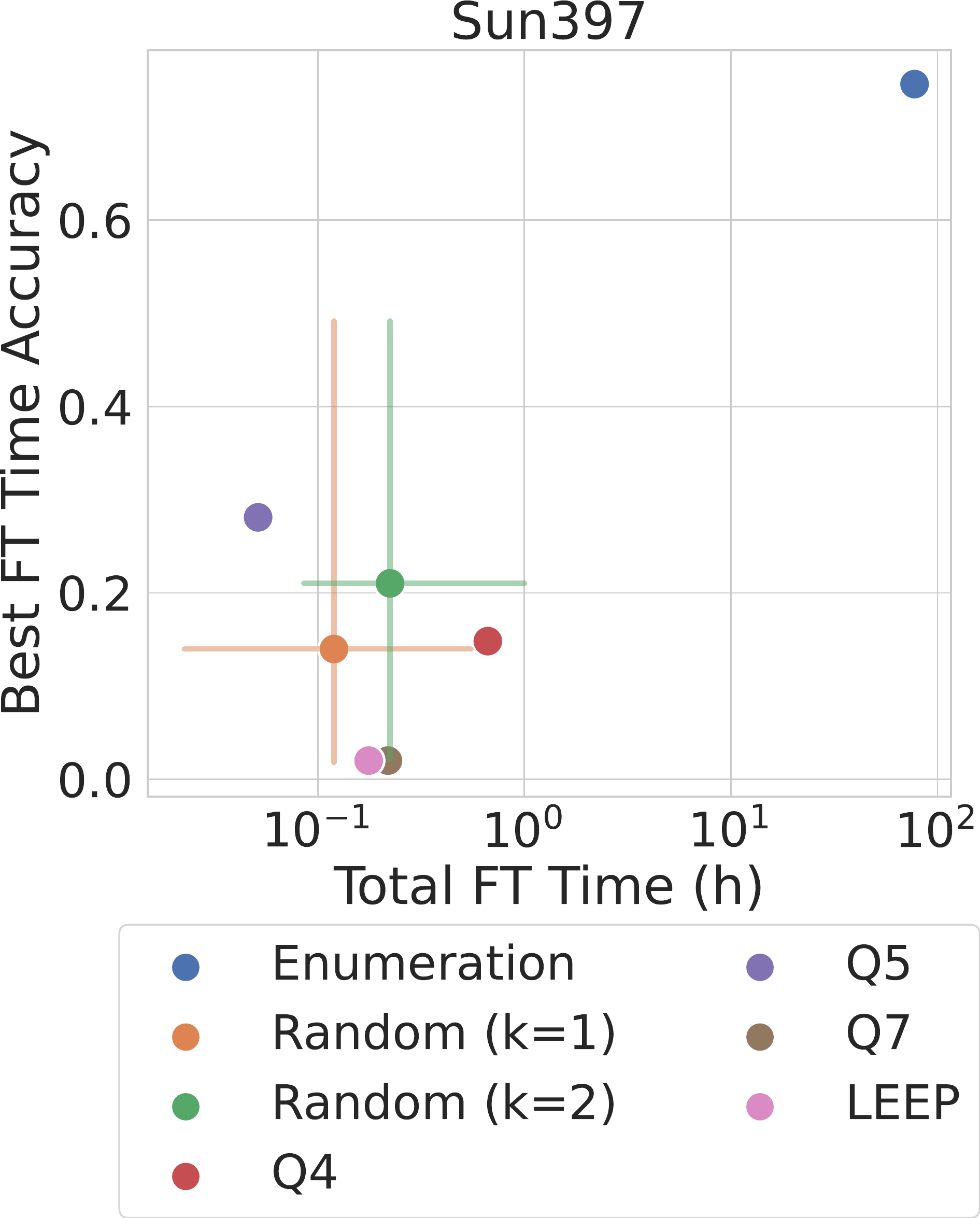}
    \end{subfigure}
    \caption{Benchmark module results 2/6.}
\end{figure}

\begin{figure}[t!]
    \centering
    \begin{subfigure}[ht]{0.33\textwidth}
        \centering
        \includegraphics[width=\textwidth]{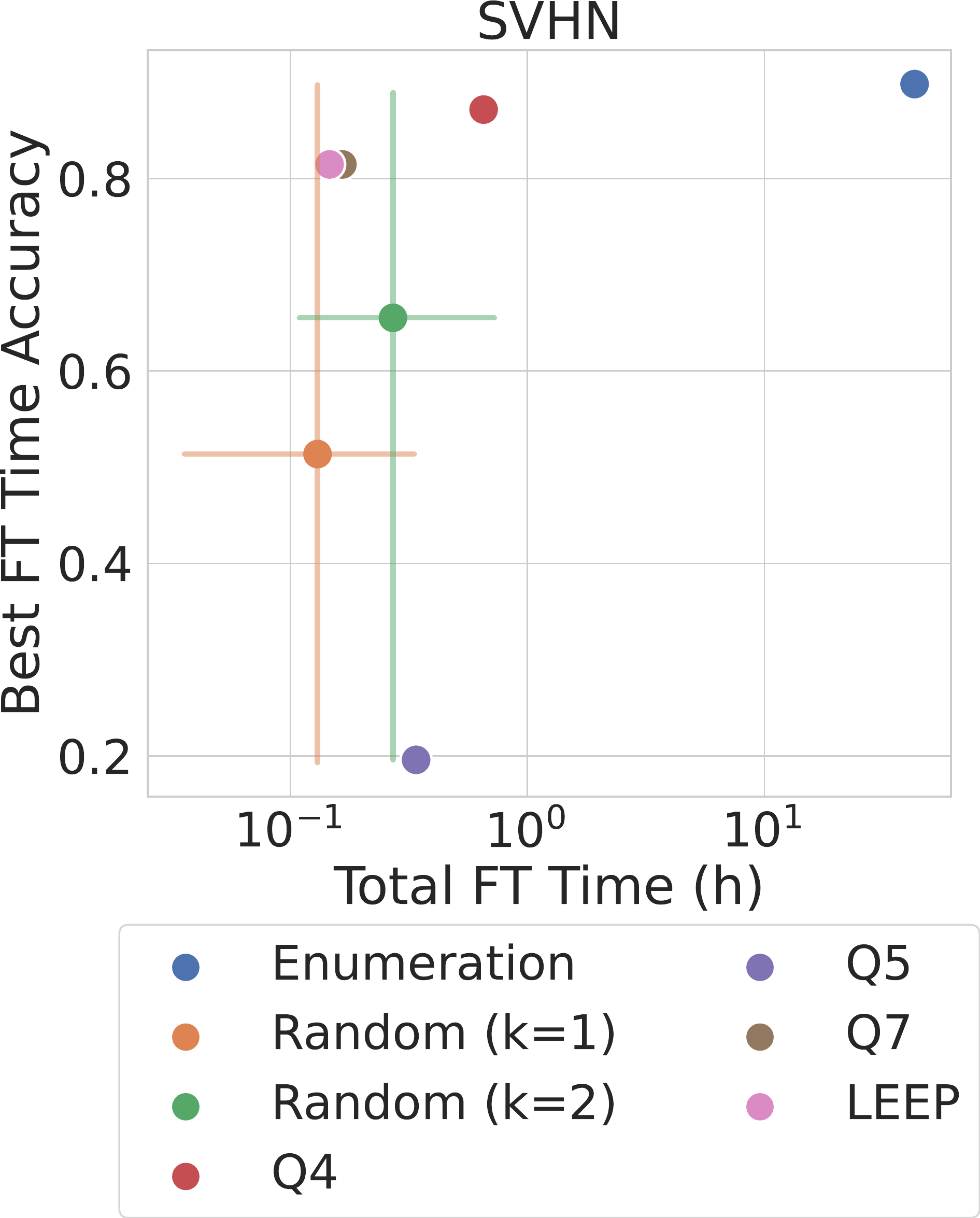}
    \end{subfigure}
    \hfill
    \begin{subfigure}[ht]{0.33\textwidth}
        \centering
        \includegraphics[width=\textwidth]{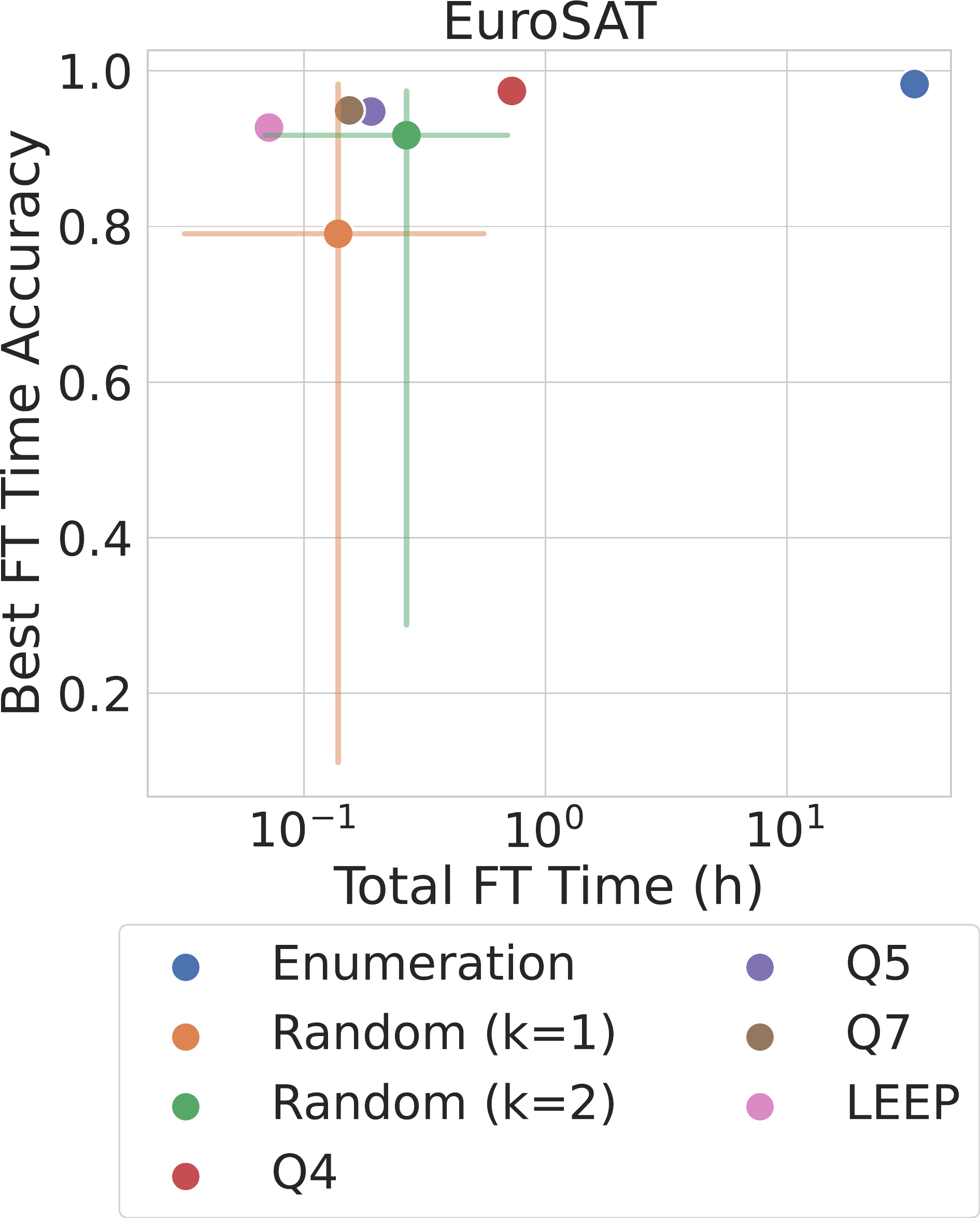}
    \end{subfigure}
    \hfill
    \begin{subfigure}[ht]{0.33\textwidth}
        \centering
        \includegraphics[width=\textwidth]{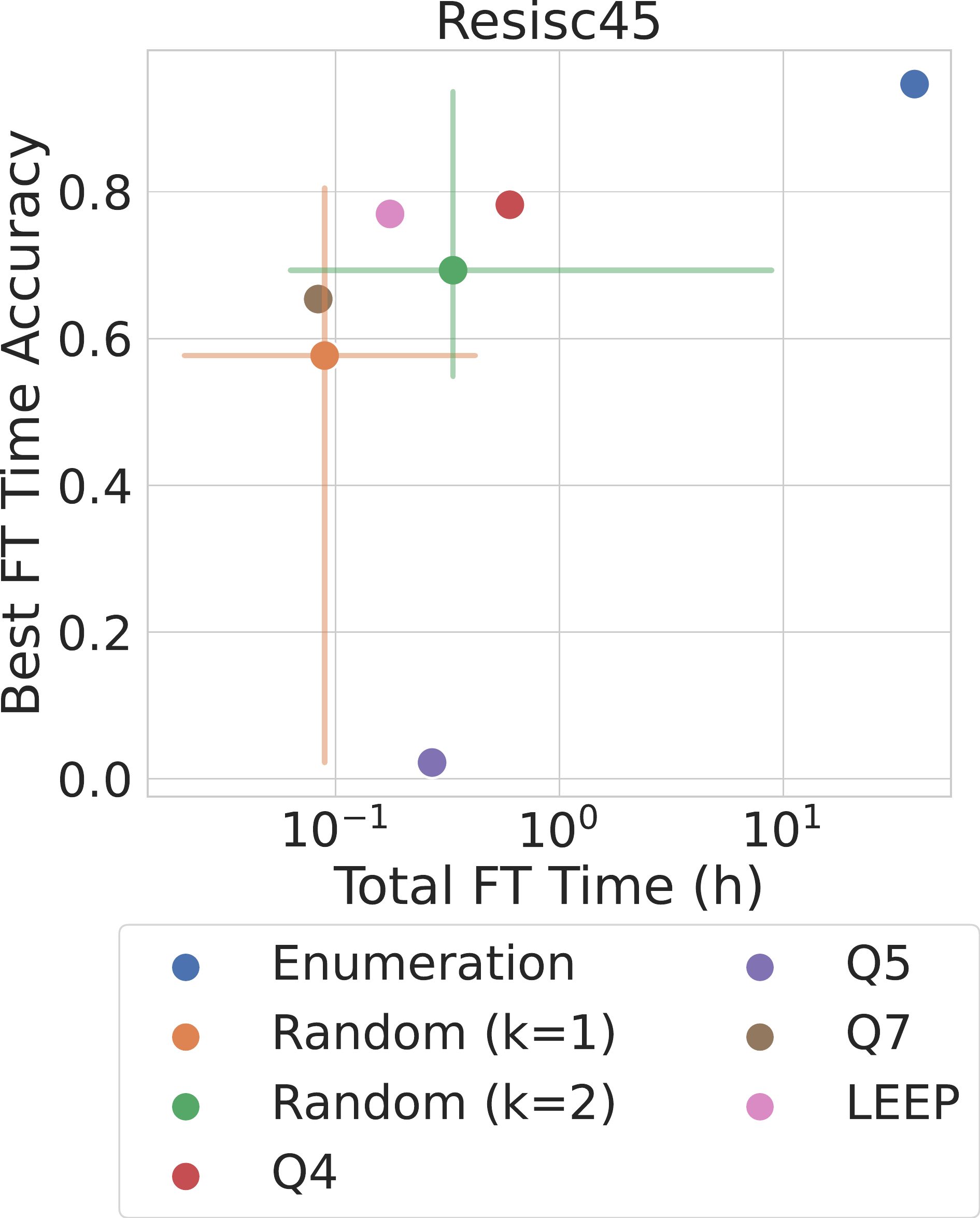}
    \end{subfigure}
    \caption{Benchmark module results 3/6.}
\end{figure}

\begin{figure}[t!]
    \centering
    \begin{subfigure}[ht]{0.33\textwidth}
        \centering
        \includegraphics[width=\textwidth]{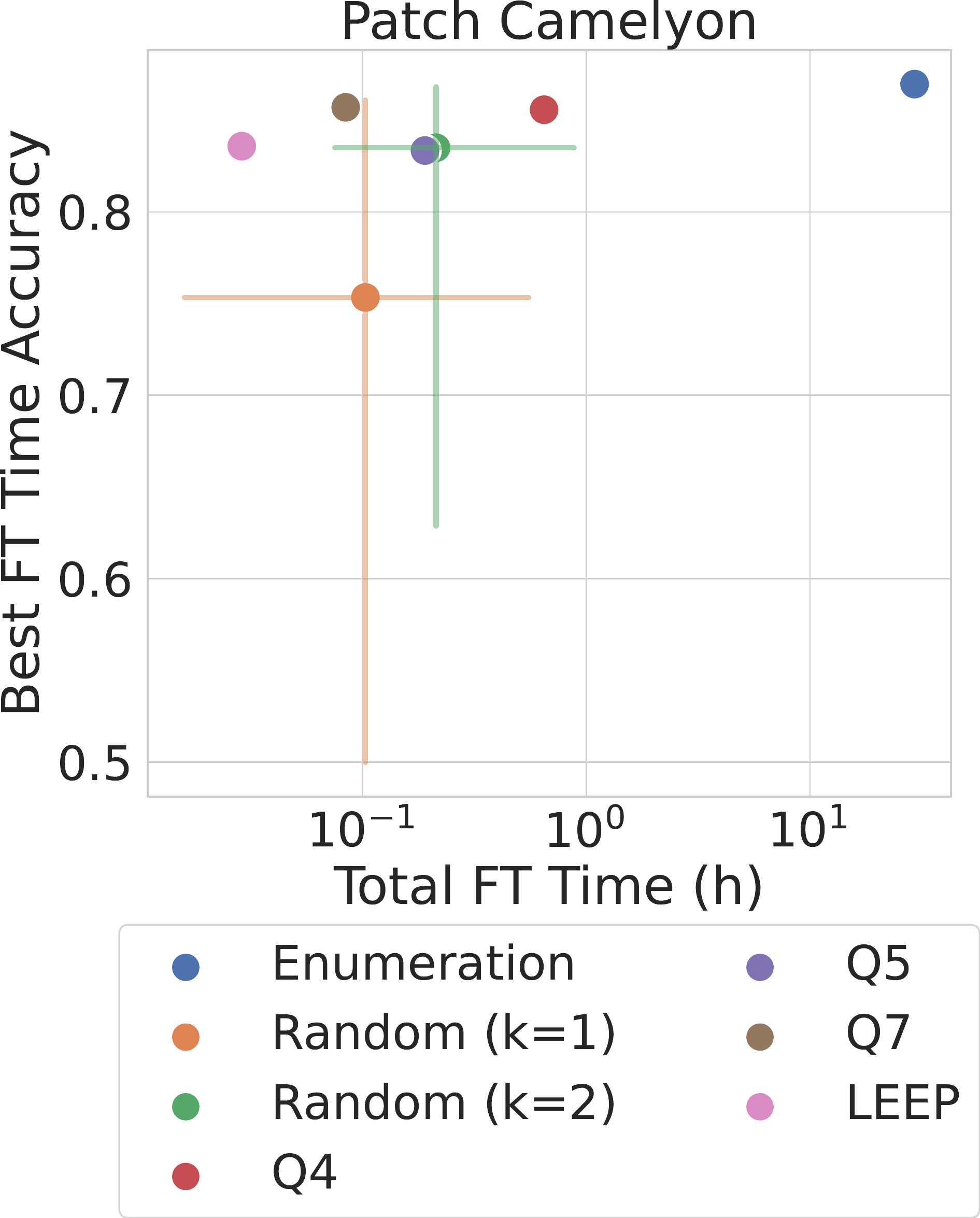}
    \end{subfigure}
    \hfill
    \begin{subfigure}[ht]{0.33\textwidth}
        \centering
        \includegraphics[width=\textwidth]{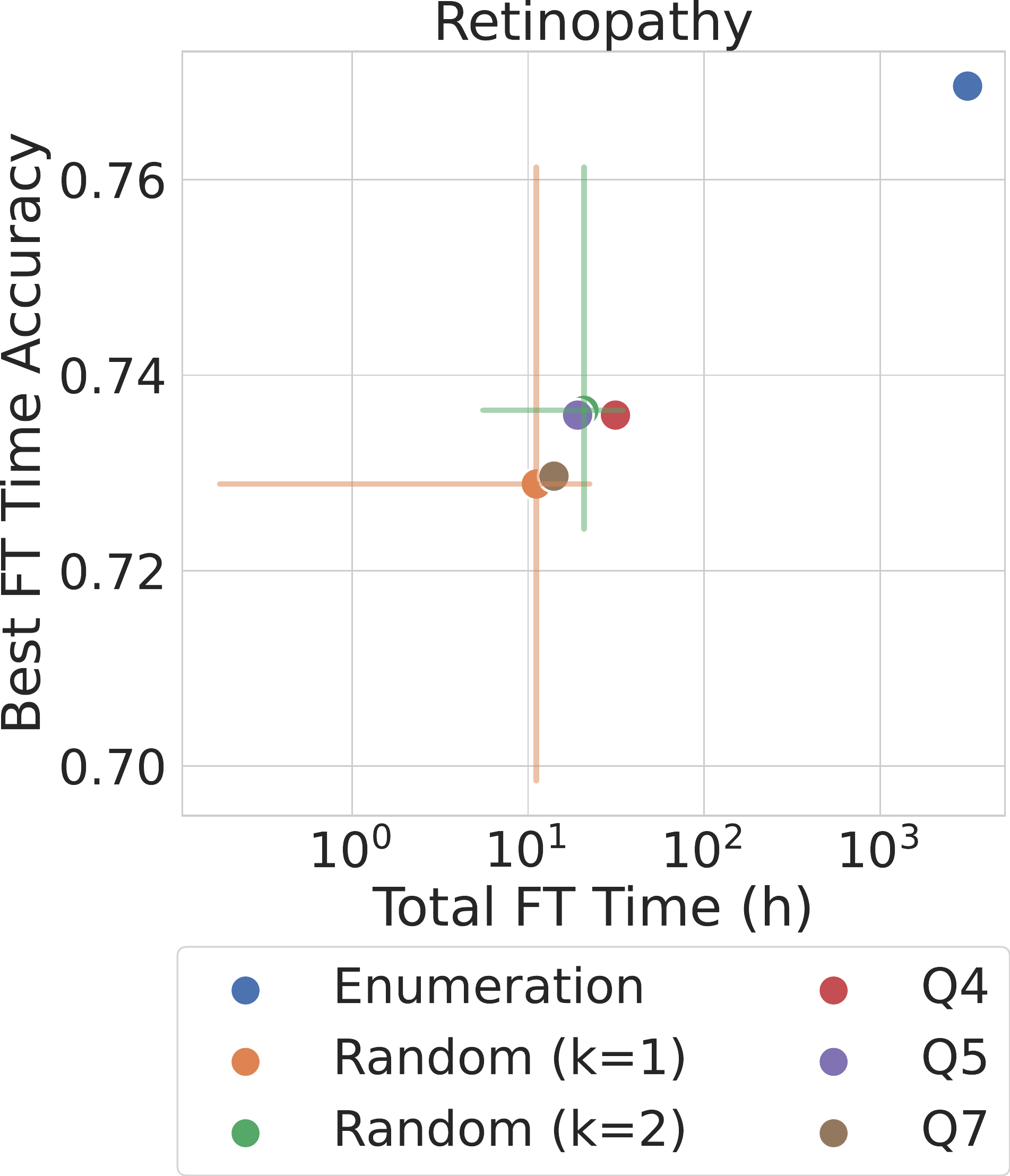}
    \end{subfigure}
    \hfill
    \begin{subfigure}[ht]{0.33\textwidth}
        \centering
        \includegraphics[width=\textwidth]{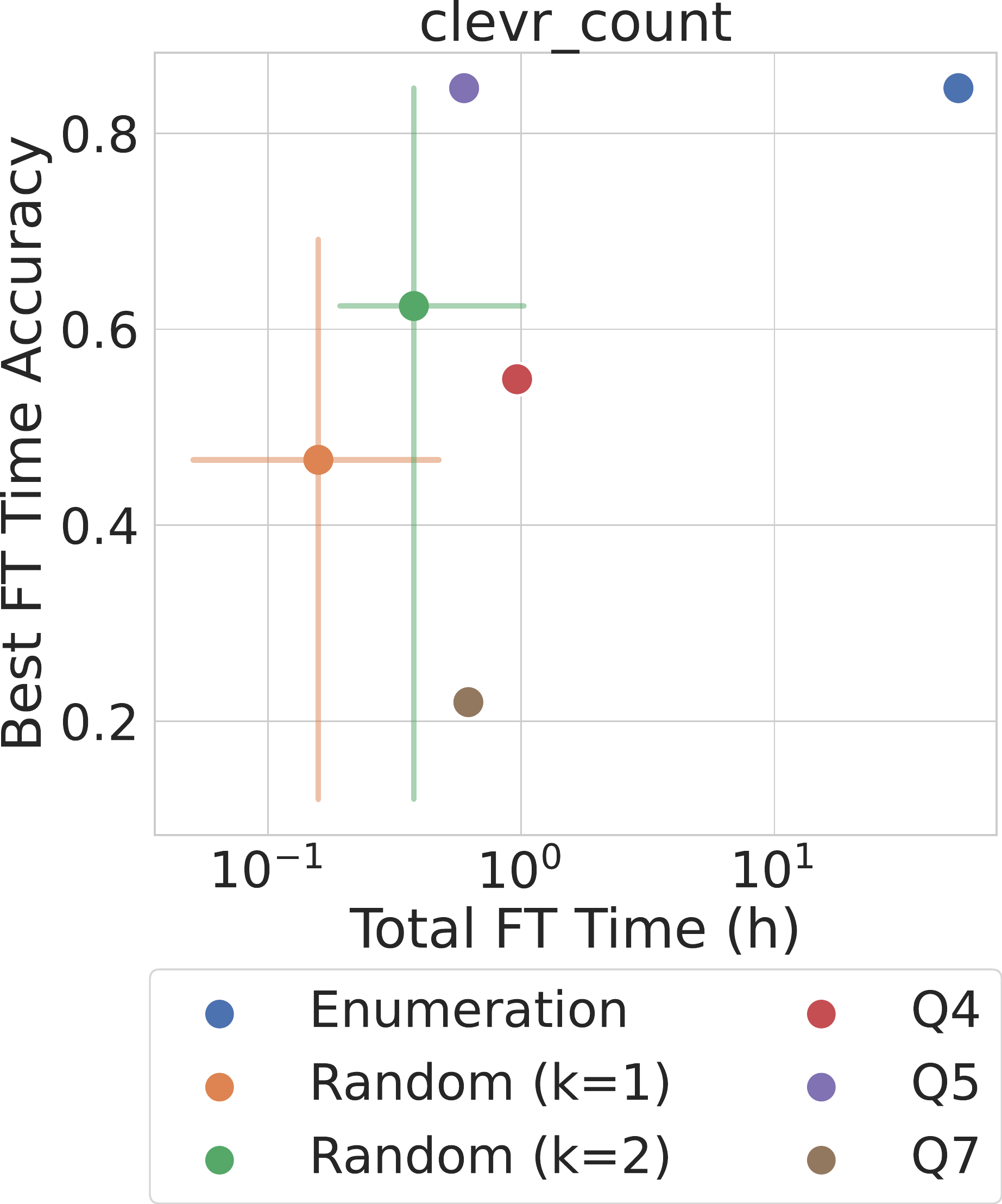}
    \end{subfigure}
    \caption{Benchmark module results 4/6.}
\end{figure}

\begin{figure}[t!]
    \centering
    \begin{subfigure}[ht]{0.33\textwidth}
        \centering
        \includegraphics[width=\textwidth]{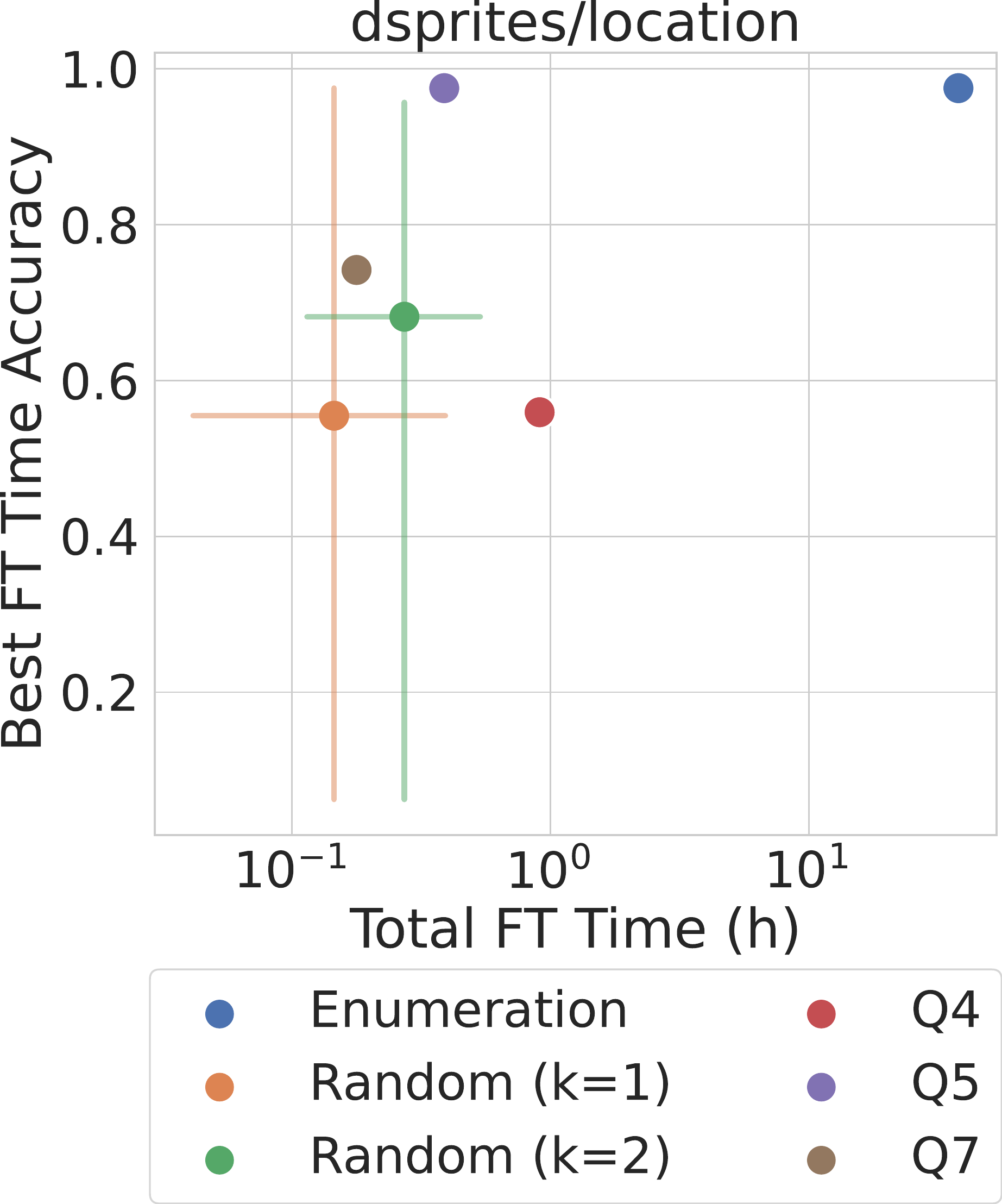}
    \end{subfigure}
    \hfill
    \begin{subfigure}[ht]{0.33\textwidth}
        \centering
        \includegraphics[width=\textwidth]{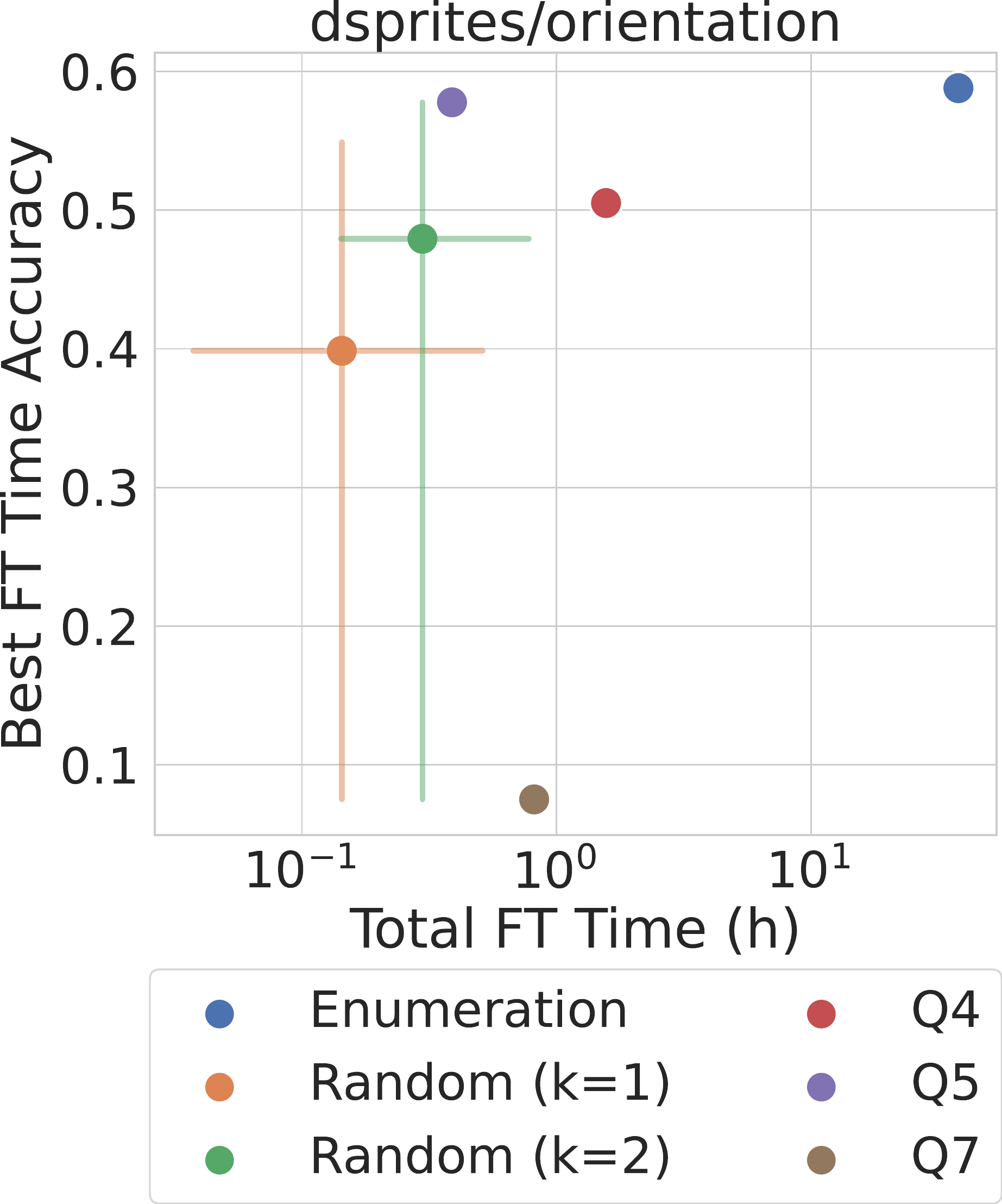}
    \end{subfigure}
    \hfill
    \begin{subfigure}[ht]{0.33\textwidth}
        \centering
        \includegraphics[width=\textwidth]{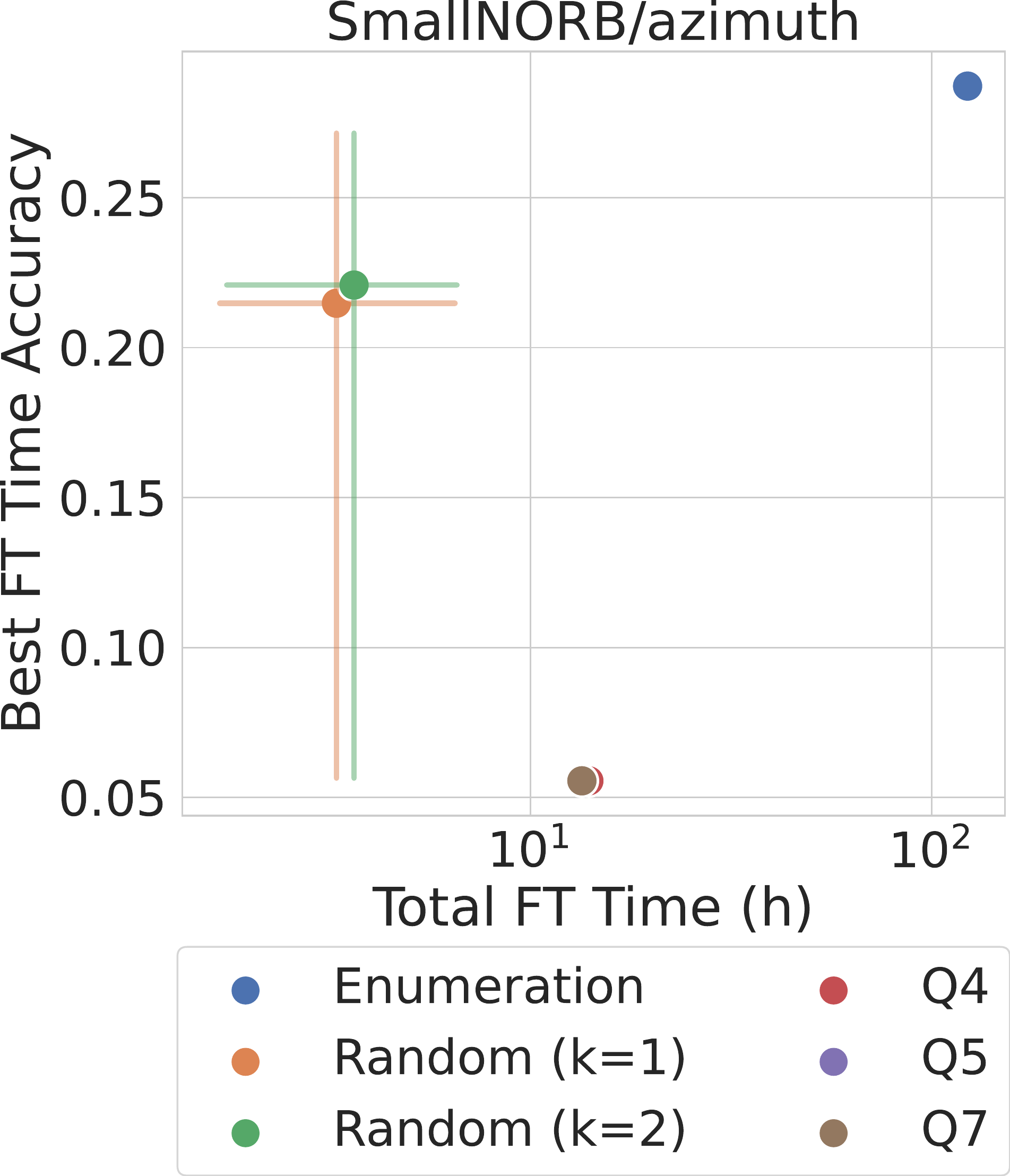}
    \end{subfigure}
    \caption{Benchmark module results 5/6.}
\end{figure}

\end{document}